%% file: main.tex
\documentclass[11pt]{article}

\usepackage[utf8]{inputenc}
\usepackage[T1]{fontenc}
\usepackage{graphicx}
\graphicspath{{plots/}{./}}
\usepackage{amsmath}
\usepackage{amssymb}
\usepackage{booktabs}
\usepackage{float}
\usepackage{multirow}
\usepackage[colorlinks=true, allcolors=blue]{hyperref}
\usepackage{natbib}
\usepackage{geometry}
\usepackage{caption}
\usepackage{subcaption}
\usepackage{xcolor}
\usepackage{pdflscape}
\usepackage[most,listings]{tcolorbox}
\usepackage{listingsutf8}
\newcommand{\authorbreak}{\end{tabular}\\[1ex]\begin{tabular}[t]{c}}

\newtcolorbox{questionbox}[1]{
    colback=gray!5!white,
    colframe=gray!75!black,
    fonttitle=\bfseries,
    title={Question #1}
}

\newtcblisting{promptbox}{
    colback=gray!5!white,
    colframe=gray!75!black,
    fonttitle=\bfseries,
    title={Prompt Template},
    breakable,
    listing only,
    listing options={
        basicstyle=\small\ttfamily,
        breaklines=true,
        breakatwhitespace=true,
        columns=fullflexible,
        keepspaces=true,
        xleftmargin=0pt,
        aboveskip=0pt,
        belowskip=0pt
    }
}

\newtcblisting{answerbox}{
    colback=blue!5!white,
    colframe=blue!50!black,
    fonttitle=\bfseries,
    title={Gemma 3 27B Response},
    breakable,
    listing only,
    listing options={
        basicstyle=\small\ttfamily,
        breaklines=true,
        breakatwhitespace=true,
        columns=fullflexible,
        keepspaces=true,
        xleftmargin=0pt,
        aboveskip=0pt,
        belowskip=0pt
    }
}

\newtcblisting{qwenanswer}{
    colback=blue!5!white,
    colframe=blue!50!black,
    fonttitle=\bfseries,
    title={Qwen3-VL-235B-A22B-Thinking Response},
    breakable,
    listing only,
    listing options={
        basicstyle=\small\ttfamily,
        breaklines=true,
        breakatwhitespace=true,
        columns=fullflexible,
        keepspaces=true,
        xleftmargin=0pt,
        aboveskip=0pt,
        belowskip=0pt
    }
}

\title{A Comparative Study in Surgical AI: \\ Potential and Limitations of Data, Compute, and Scaling}

\author{
    Kirill Skobelev\textsuperscript{1}\thanks{Lead Author: \texttt{Kirill.Skobelev@chicagobooth.edu}} \and
    Eric Fithian\textsuperscript{1} \and
    Yegor Baranovski\textsuperscript{1} \authorbreak
    Jack Cook\textsuperscript{2} \and
    Sandeep Angara\textsuperscript{2} \and
    Shauna Otto\textsuperscript{2} \and
    Zhuang-Fang Yi\textsuperscript{2} \and
    John Zhu\textsuperscript{2} \and
    Neeraj Mainkar\textsuperscript{2}\thanks{Corresponding Authors: \texttt{\{dan,neeraj.mainkar,margaux\}@surgicalvideo.io},  \texttt{XY.Han@chicagobooth.edu}} \and
    Margaux Masson-Forsythe\textsuperscript{2}\footnotemark[2] \authorbreak
    Daniel A. Donoho\textsuperscript{2,3}\footnotemark[2] \qquad X.Y. Han\textsuperscript{1,4}\footnotemark[2]
    \\[1em]
    \textsuperscript{1}Center for Applied AI, Chicago Booth, Chicago, IL, USA \\
    \textsuperscript{2}Surgical Data Science Collective, Washington D.C., USA
    \\
    \textsuperscript{3}Children's National Hospital, Washington D.C., USA
    \\
    \textsuperscript{4}Operations Management \& Tolan Center for Healthcare, Chicago Booth, Chicago, IL, USA
}

\date{\today}

\begin{document}

\sloppy
\emergencystretch=2em

\maketitle
\vspace{-0.3in}
\input{sections/0_abstract}
\pagebreak
{\small \sloppy \paragraph*{\small Results Summary.} We present findings from seven experiments. \textbf{(1)} We evaluate zero-shot surgical tool detection performance across 20 open-weight Vision Language Models (VLMs) from 2023 to 2026 on SDSC-EEA, a video dataset consisting of endoscopic endonasal approach (EEA) neurosurgical procedures. Despite dramatic increases in model scale and benchmark scores, only one model marginally exceeds the 13.4\% majority class baseline on the validation set. \textbf{(2)} We fine-tune Gemma 3 27B with LoRA adapters to generate structured JSON predictions. The model achieves 47.63\% exact match accuracy, surpassing the validation set baseline of 13.41\%. \textbf{(3)} We replace off-the-shelf JSON generation with a specialized classification head. This approach achieves 51.08\% exact match accuracy. \textbf{(4)} To assess the potential of increasing computational resources, we gradually increase the effective number of trainable parameters (by increasing LoRA rank) by nearly three orders of magnitude. While training accuracy reaches 98.6\%, validation accuracy remains below 40\%, showing that scaling alone cannot overcome distribution shift. \textbf{(5)} We compare zero-shot and fine-tuned VLM performance against YOLOv12-m, a specialized 26M-parameter object detection model. YOLOv12-m achieves 54.73\% exact match accuracy, outperforming all VLM-based methods while using 1,000$\times$ fewer parameters. \textbf{(6)} We demonstrate these findings generalize to three independent and public datasets---CholecT50, PitVis-2023, and SurgVU---with additional comparisons on eleven frontier VLMs. On CholecT50, a dataset of laparoscopic cholecystectomy procedures, the fine-tuned open-weight model and YOLOv12-m outperform all zero-shot VLM methods including zero-shot methods using proprietary frontier VLMs. On PitVis-2023, a public endoscopic pituitary neurosurgery benchmark with 18 instrument classes, the fine-tuned open-weight model again leads (84.77\% exact match accuracy) followed by YOLOv12-m (82.78\%); the best closed-weight frontier model, Claude Fable 5.1, reaches 58.70\%. On SurgVU, a public benchmark of robotic-assisted surgery training sessions on porcine tissue with 17 released instrument classes, zero-shot Gemma~3 27B achieves only 2.90\% exact match accuracy, well below the 16.94\% majority class baseline; only five of the eleven frontier models clearly exceed that baseline (Claude Sonnet 4.6 at 23.05\%, Gemini 3.1 Pro Preview at 22.46\%, Claude Fable 5.1 at 21.00\%, Claude Fable 5 at 20.09\%, Gemini 3.8 Flash at 19.80\%), while the remaining six sit at or below it. LoRA fine-tuning of Gemma~3 27B reaches 50.61\% and YOLOv12-m reaches 51.75\%, both at least 28~percentage points above every frontier model. As on SDSC-EEA, the train-validation gap on CholecT50, PitVis-2023, and SurgVU widens with LoRA rank, confirming the same pattern across four surgical domains. \textbf{(7)} To test whether these findings are an artifact of the detection task, we evaluate four bounding-box-free annotation tasks on public datasets---surgical workflow (phase and step) recognition on PitVQA, fine-grained suturing-gesture recognition on SAR-RARP50, multi-label surgical action recognition on CholecT50, and anatomy presence on DSAD---comparing zero-shot open-weight VLMs, ten frontier VLMs, a surgical foundation model with a linear probe (LemonFM), LoRA fine-tuning of Gemma~3 27B, and supervised ResNet-50 specialists. Small supervised specialists lead three of the four datasets (77.1\% micro-F1 on PitVQA, 53.0\% on SAR-RARP50, 71.6\% on DSAD); on CholecT50 action recognition the LoRA classification head leads (78.8\%). All ten frontier VLMs trail the top two models on every dataset, and only Claude Fable 5.1 clears the majority class baseline on gesture recognition.}

\null \vfill


\input{sections/1_introduction}

\input{sections/2_methods}

\input{sections/3_results}

\input{sections/4_discussion}

\input{sections/5_limitations}

\input{sections/6_conclusion}

\newpage

\paragraph{Funding and Support.}
This project is jointly funded by the Booth School of Business at UChicago, the Center for Applied AI at Chicago Booth, and the Surgical Data Science Collective (SDSC). Collaborative data sharing between Chicago Booth and the SDSC was facilitated by the Tolan Center for Healthcare at Chicago Booth and the SDSC Engineering Team. Computational experiments were conducted on the Pythia Supercomputer Cluster at Chicago Booth.

\begingroup
\raggedright
\sloppy
\bibliographystyle{plainnat}
\bibliography{references}
\endgroup

\newpage

\appendix
\input{sections/7_appendix}

\end{document}

%% file: sections/0_abstract.tex
\begin{abstract}
\sloppy

    Recent Artificial Intelligence (AI) models have matched or exceeded human experts in several benchmarks of biomedical task performance, but multi-modal benchmarks involving surgery in particular are often missing from prominent medical benchmark suites (specifically, those requiring visual recognition beyond just text question-answering). Since surgery requires coordinating disparate tasks---including multimodal data integration, human interaction, and physical effects---generally-capable AI models could be particularly attractive as collaborative tools if performance could be improved. On the one hand, the canonical approach of scaling architecture size and training data is attractive, especially since there are millions of hours of surgical video data generated per year. On the other hand, preparing surgical data for AI training requires significantly higher levels of professional expertise, and training on that data requires expensive computational resources. These trade-offs paint an uncertain picture of whether and to-what-extent modern AI could aid surgical practice. In this paper, we explore this question through a case study of surgical tool detection using state-of-the-art AI methods available in 2026. We demonstrate that even with multi-billion parameter models and extensive training, current Vision Language Models fall short in the seemingly simple task of tool detection in neurosurgery. Additionally, we show scaling experiments indicating that increasing model size and training time only leads to diminishing improvements in relevant performance metrics. To test whether these conclusions are an artifact of the detection task, we further evaluate four bounding-box-free annotation tasks on public datasets---surgical workflow recognition, fine-grained gesture recognition, surgical action recognition, and anatomy presence---comparing zero-shot and frontier Vision Language Models against a surgical foundation model with a linear probe, LoRA fine-tuning, and small supervised specialists; the same specialist-over-generalist hierarchy replicates on every task. Thus, our experiments suggest that current models could still face significant obstacles in surgical use cases. Moreover, some obstacles cannot simply be ``scaled away'' with additional compute and persist across diverse model architectures, raising the question of whether data and label availability are the only limiting factors. We discuss the main contributors to these constraints and advance potential solutions. 
    

\end{abstract}

%% file: sections/1_introduction.tex
\section{Introduction}
\label{sec:introduction}

The scaling hypothesis has become the dominant paradigm in AI research. \cite{kaplan2020scalinglawsneurallanguage} documented that cross-entropy loss scales with model size, data, and compute as a power law. \cite{wei2022emergentabilitieslargelanguage} argued that certain capabilities emerge beyond critical model scales, while \cite{chowdhery2022palmscalinglanguagemodeling} demonstrated broad few-shot performance gains and emergent abilities in a 540B-parameter language model. These observations have led to increasingly bold claims: \cite{bubeck2023sparksartificialgeneralintelligence} interpret GPT-4's behavior as indicative of emerging AGI, and \cite{aschenbrenner2024situational} explicitly argues that continued scaling alone is sufficient to reach AGI.

In medicine, similar optimism has taken hold. \cite{saab2024capabilitiesgeminimodelsmedicine} present Med-Gemini, a family of models achieving 91.1\% on MedQA and large gains over GPT-4V on multimodal benchmarks, as evidence that large multimodal foundation models can deliver strong generalist capabilities across medical specialties. Such benchmark results have fueled speculation about the feasibility of a ``Medical Artificial General Intelligence'' (Med-AGI) through scaling. Yet, when tested in realistic clinical settings, the picture is less optimistic. For example, \cite{hager2024evaluationmitigationlimitations} find that state-of-the-art LLMs perform significantly worse than physicians across pathologies, often failing to follow instructions. \cite{wu2025generalistfoundationradiology} further demonstrate that ``generalist'' radiology capability depends on large-scale in-domain pretraining and radiology-specific instruction tuning, suggesting progress toward Med-AGI may be bottlenecked by domain data coverage as much as by parameter count. Conversely, \citet{vishwanath_general-purpose_2026} find that general-purpose frontier LLMs outperform specialized clinical AI tools on several text-based medical-knowledge benchmarks, indicating that the relative performance of generalist and specialized models is dataset-dependent and, in particular, sensitive to a model's exposure to the task during training. For intraoperative perception, this question remains untested.

In surgery specifically, recent work has begun to apply vision–language models to surgical data across a range of tasks. Surgical-VQA \citep{seenivasan2022surgicalvqavisualquestionanswering} introduces visual question answering over laparoscopic scenes, while GP-VLS \citep{schmidgall2024gpvlsgeneralpurposevisionlanguage} demonstrates that large foundation models can be adapted to multiple surgical tasks, including instrument recognition, through extensive in-domain supervision. Related efforts fine-tune vision–language models for tool-related tasks such as keypoint estimation using low-rank adaptation, often relying on synthetic datasets to augment limited real annotations \citep{duangprom2025estimating2dkeypointssurgical}. This literature establishes VLMs as a viable modeling paradigm for surgical understanding and motivates their evaluation on fine-grained surgical perception tasks using real operative video. (See Section \ref{sec:discussion} for additional discussion on related works in medical and surgical AI evaluation.)

Despite progress on medical visual tasks, whether these models would lead to Med-AGI is an open question. However, through the lens of surgery in particular, locating and classifying surgical instruments is the earliest (necessary, not sufficient) task for Med-AGI to achieve surgical competence. Non-expert humans excel at this task: annotators in our study learned to label these tools with near-perfect accuracy after minimal training. Yet, evaluations of the competency of generally-capable state-of-the-art foundation methods for surgery are notably underrepresented in the literature: For example, prominent works such as MedGemma \citep{sellergren2025medgemmatechnicalreport}, MedMarks \citep{warner2025medmarks}, MedFrameQA \citep{yu2026medframeqamultiimagemedicalvqa}, MedXpertQA-MM \citep{zuo2025medxpertqabenchmarkingexpertlevelmedical}, MultiMedEval \citep{pmlr-v250-royer24a}, OmniMedVQA \citep{hu2024omnimedvqanewlargescalecomprehensive}, PMC-VQA \citep{zhang2024pmcvqavisualinstructiontuning} all present extensive benchmarks of medical AI capabilities, but none contain any evaluations that involve visual surgical recognition.
In this paper, we evaluate state-of-the-art AI models for tool detection on SDSC-EEA, a unique dataset of 67,634 annotated frames from neurosurgical videos provided by the \cite{sdsc2026}. 

Because object detection is only one of the major annotation tasks performed on surgical video---and because its bounding-box structure could favor detector architectures---we further extend the evaluation to four bounding-box-free annotation tasks on public datasets: frame-level surgical workflow (phase and step) recognition on PitVQA \citep{he2024pitvqa}, fine-grained suturing-gesture recognition on SAR-RARP50 \citep{psychogyios2023sarrarp50}, multi-label surgical action recognition on CholecT50 \citep{cholec} (the verb component of its action-triplet labels), and anatomical-structure presence on DSAD \citep{carstens2023dresden}. On these tasks we compare zero-shot open-weight VLMs, eight frontier VLMs from the model generation current at the time of writing, a surgical vision foundation model probed with a linear head (LemonFM; \citealp{che2026lemonlargeendoscopicmonocular}), LoRA fine-tuning of Gemma~3 27B, and small supervised specialists, bringing the study to 35 distinct vision--language models (21 open-weight, 14 frontier closed-weight)---plus a surgical foundation model and two specialist architectures---across seven datasets and five annotation-task families.

The paper is organized as follows:

\begin{itemize}
\item Section~\ref{sec:methods} describes the datasets, models, and experimental methodology for eight evaluations spanning zero-shot inference, two flavors of LoRA fine-tuning (JSON generation and classification head), LoRA rank scaling, a specialized object-detection baseline (YOLOv12-m), and full replication of these protocols on three independent public datasets: CholecT50 (laparoscopic cholecystectomy), PitVis-2023 (endoscopic pituitary neurosurgery), and SurgVU (robotic-assisted training sessions on porcine tissue). It also describes the four bounding-box-free annotation tasks---workflow, gesture, action, and anatomy recognition on PitVQA, SAR-RARP50, CholecT50, and DSAD---and the five model classes compared on them.

\item Section~\ref{sec:results} presents seven findings, one per subsection:
\begin{enumerate}
\item Zero-shot open-weight VLMs do not surpass a trivial baseline (Section~\ref{subsec:zero_shot}). Across 20 models spanning 2B--235B parameters and nearly three years of development, validation accuracy on SDSC-EEA remains at or near the majority class baseline of 13.4\%.
\item LoRA fine-tuning with structured JSON generation improves exact match accuracy from 9.8\% to 47.6\%, but a substantial train--validation gap remains (Section~\ref{subsec:lora_formatting}).
\item Replacing JSON generation with a classification head pushes validation accuracy to 51.1\%, with training accuracy reaching 89.5\%---confirming that the model fits the training distribution well but generalizes poorly to held-out procedures (Section~\ref{subsec:lora_classification_head}).
\item Scaling LoRA adapter rank by nearly three orders of magnitude does not resolve the gap: training accuracy reaches 98.6\% while validation stays below 40\% (Section~\ref{subsec:lora_rank_sweep}).
\item A 26M-parameter specialized object detector outperforms every VLM-based approach: YOLOv12-m reaches 54.7\% exact match with 1{,}000$\times$ fewer parameters than the best VLM (Section~\ref{subsec:yolo_performance}).
\item The same patterns replicate on three additional independent and public surgical  datasets: CholecT50, PitVis-2023, and SurgVU. We furthermore include additional comparisons using eleven frontier VLMs (GPT-5.4, GPT-5.6 Sol, Claude Opus 4.6, Claude Sonnet 4.6, Claude Fable 5, Claude Fable 5.1, Gemini 3 Flash Preview, Gemini 3.1 Pro Preview, Gemini 3.8 Flash, Kimi K3, Qwen3.8 Max). On CholecT50, every frontier model improves over zero-shot Gemma but remains well below the fine-tuned open-weight Gemma and YOLOv12-m. On PitVis-2023, fine-tuned Gemma~3 27B leads (84.8\%), followed by YOLOv12-m (82.8\%), with the best frontier model (Claude Fable 5.1$^*$) at 58.7\%. On SurgVU, the same fourteen-model evaluation widens the gap: only five of the eleven frontier models clearly exceed the trivial 16.94\% baseline, while fine-tuned Gemma~3 27B (50.6\%) and YOLOv12-m (51.8\%) lead by 28--29~percentage points (Section~\ref{subsec:robustness}).
\item The specialist-over-generalist hierarchy is not an artifact of the detection task. On four bounding-box-free annotation tasks---workflow recognition (PitVQA), gesture recognition (SAR-RARP50), action recognition (CholecT50), and anatomy presence (DSAD)---small supervised specialists or a task-adapted open-weight VLM lead every dataset, a frozen surgical foundation model with a linear probe follows, and all ten frontier VLMs trail the top two models on every dataset, with only one clearing the majority class baseline on gesture recognition (Section~\ref{subsec:newtasks}).
\end{enumerate}

\item Section~\ref{sec:discussion} argues that the bottleneck to surgical AI is specialized data, not model scale, and proposes hierarchical architectures where generalist VLMs delegate to specialized perception modules. We also discuss closely related works on surgical AI evaluation and how their discoveries complement our results.

\item Section~\ref{sec:limitations} discusses limitations.

\item Section~\ref{sec:conclusion} concludes the paper.
\end{itemize}

%% file: sections/2_methods.tex
\section{Methods}
\label{sec:methods}

This section describes the dataset and experimental methodology. Section~\ref{subsec:dataset} introduces the SDSC-EEA dataset. Section~\ref{subsec:zero_shot_methods} describes zero-shot VLM evaluation. Section~\ref{subsec:lora_methods} describes LoRA fine-tuning of a VLM. Section~\ref{subsec:yolo_methods} describes a specialized object baseline. Sections~\ref{subsec:cholect50_methods}, \ref{subsec:pitvis_methods}, and~\ref{subsec:surgvu_methods} describe validation on three external public datasets, CholecT50, PitVis-2023, and SurgVU. Section~\ref{subsec:newtasks_methods} describes the extension to four bounding-box-free annotation tasks. Section~\ref{subsec:metrics} defines the evaluation metrics used throughout. Corresponding results for each experiment are reported in Section~\ref{sec:results}.

\subsection{SDSC-EEA Dataset}
\label{subsec:dataset}

We evaluate surgical tool detection using a dataset of endoscopic endonasal approach (EEA) neurosurgical procedures. EEA is a minimally invasive technique used to assess and treat lesions at the skull base through the nasal passages. The dataset is provided by the \citeauthor{sdsc2026} and comprises of 67,634 annotated frames extracted from 66 unique surgical procedures. Figure \ref{fig:sample_frames} exhibits frames from some videos sampled from this dataset. We refer to it as SDSC-EEA in this paper. 

The dataset was constructed from video recordings of surgical procedures donated to the SDSC by 10 surgeons across 7 institutions in the United States, France, and Spain. No exclusion criteria were applied. Ground truth annotations were produced by three annotators from a contracted labeling company, none of whom had clinical experience; annotators were provided with tool descriptions and representative example images prior to labeling. Labels were first reviewed by a senior annotator at the contracting company and subsequently by members of the SDSC. Fewer than 10\% of frames required correction.

Each frame is annotated with multi-label ground truth indicating the presence or absence of 31 distinct surgical instrument classes. Annotations are provided in YOLO format with bounding box coordinates. The average number of tools per frame is 1.72 (median: 2), with the distribution showing 7.6\% of frames containing no tools, 34.4\% containing one tool, 38.2\% containing two tools, and 19.8\% containing three or more tools.

The tool class distribution exhibits significant imbalance. Suction is the most prevalent instrument, appearing in 63.3\% of all frames. Cotton Patty (16.1\%), Grasper (10.6\%), Curette (8.6\%), and Rhoton Dissector (8.0\%) follow in frequency. 

For all fine-tuning experiments (Section~\ref{subsec:lora_methods}), we split the data by surgical procedure instances to prevent data leakage. Frames from the same surgical procedure appear exclusively in either the training or validation set, never both. This yields 47,618 training frames from 53 procedures and 20,016 validation frames from 13 procedures.

\subsection{Zero-Shot Evaluation of Vision-Language Models}
\label{subsec:zero_shot_methods}

\begin{figure}[th]
\centering
\newcommand{\exwidth}{0.18\textwidth}
\textbf{Correct Detections}\\[0.3em]
\includegraphics[width=\exwidth]{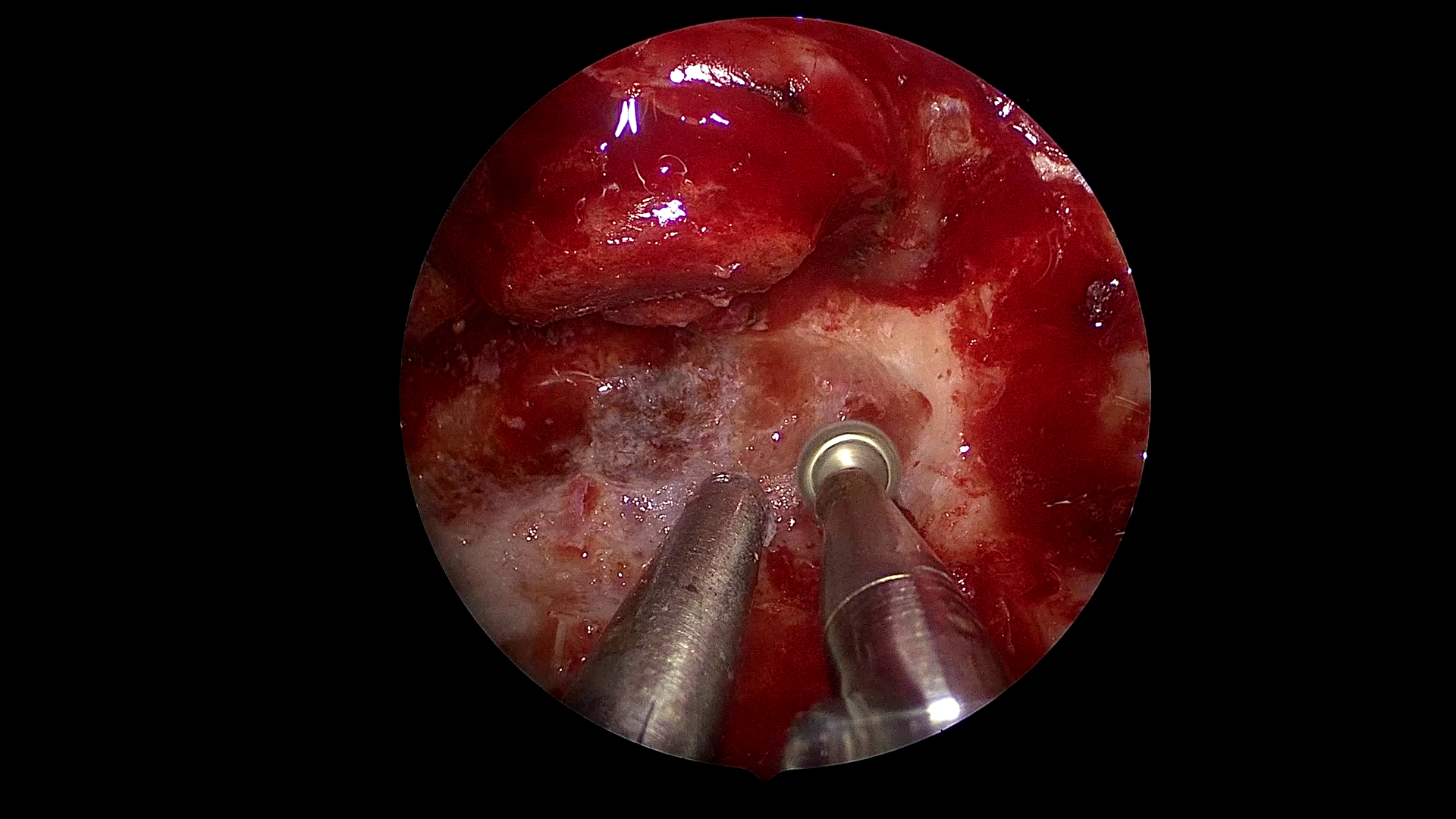}\hfill
\includegraphics[width=\exwidth]{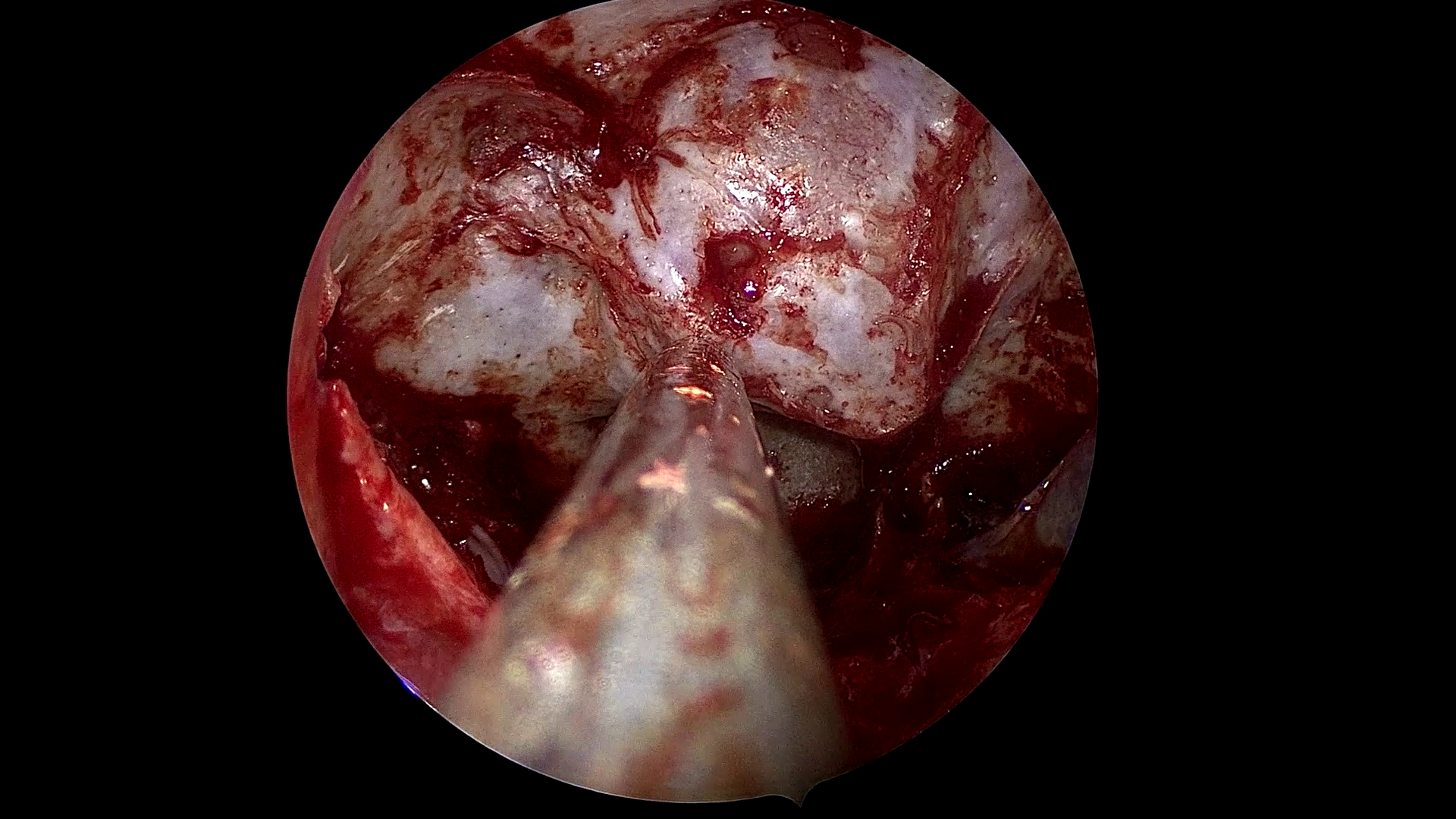}\hfill
\includegraphics[width=\exwidth]{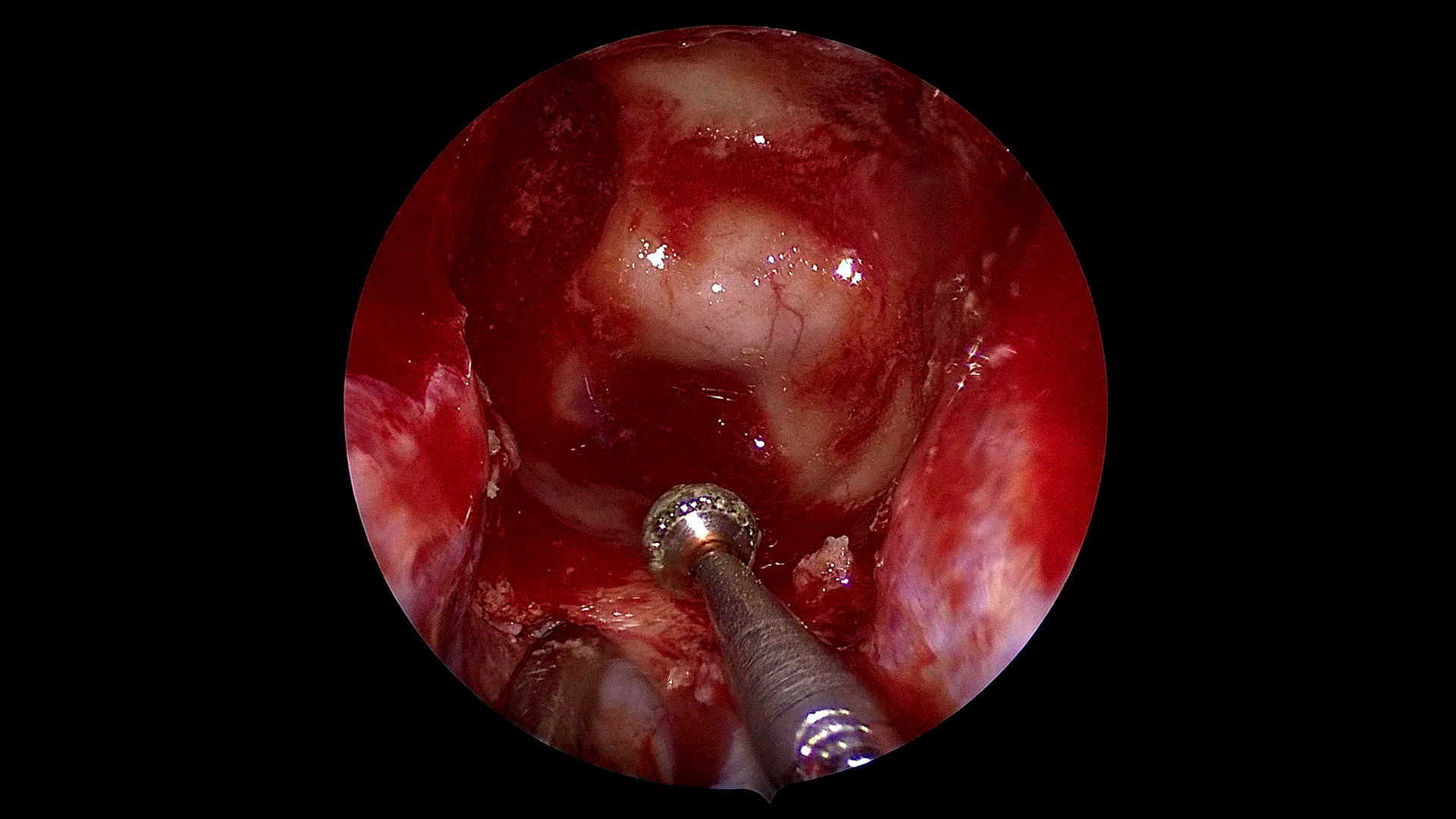}\hfill
\includegraphics[width=\exwidth]{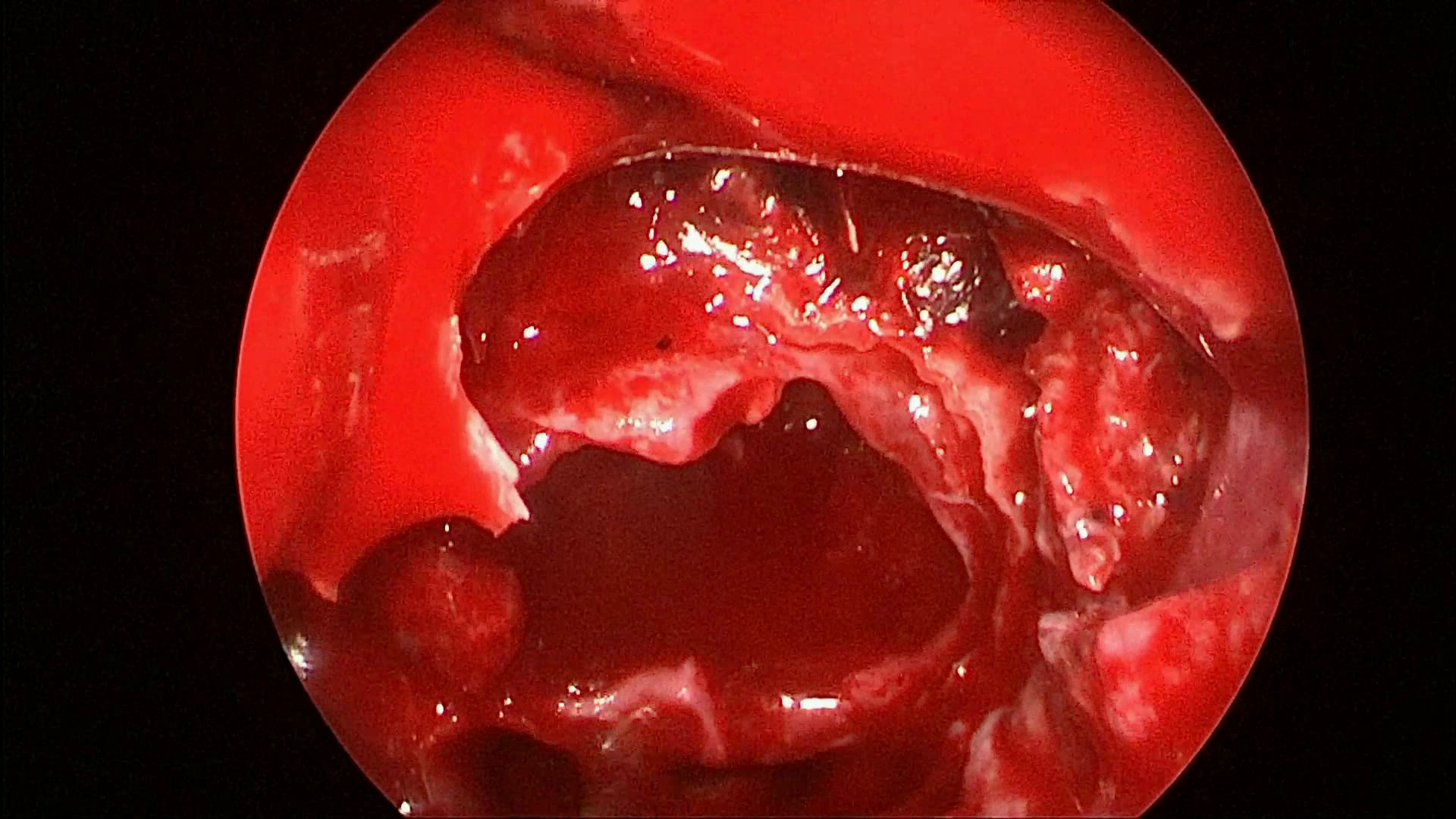}\hfill
\includegraphics[width=\exwidth]{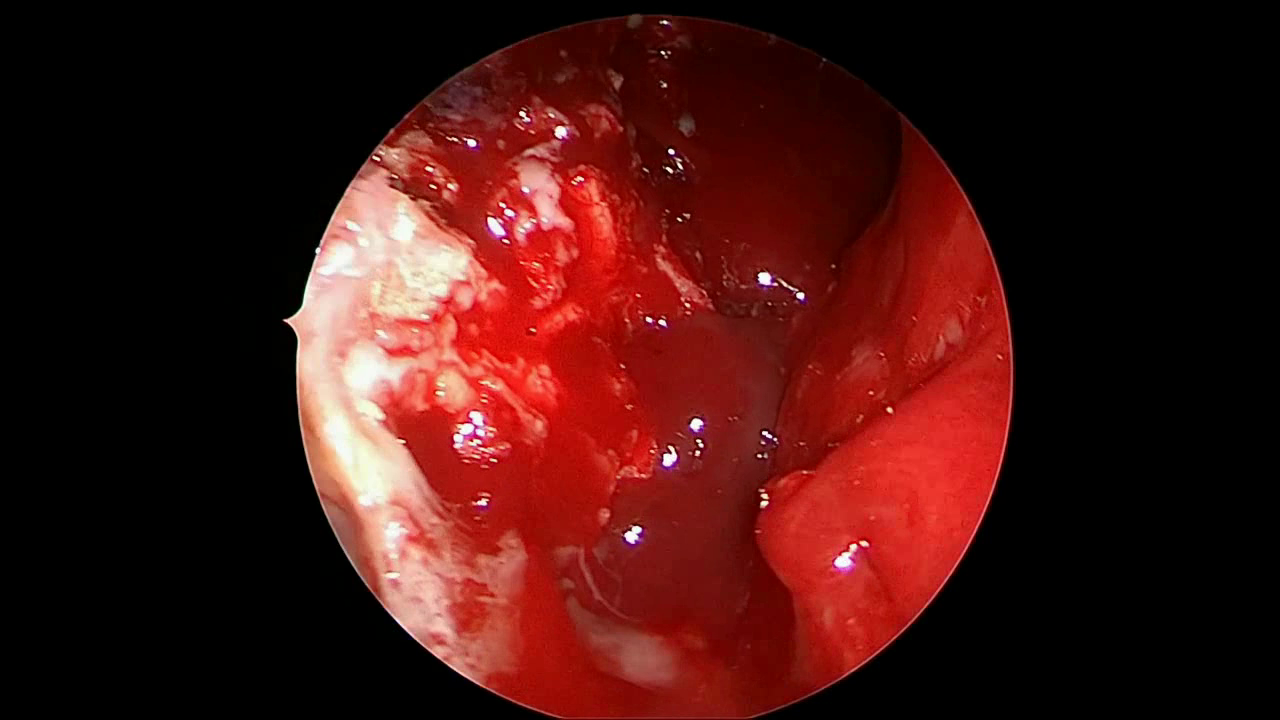}\\[0.5em]
\textbf{Incorrect Detections}\\[0.3em]
\includegraphics[width=\exwidth]{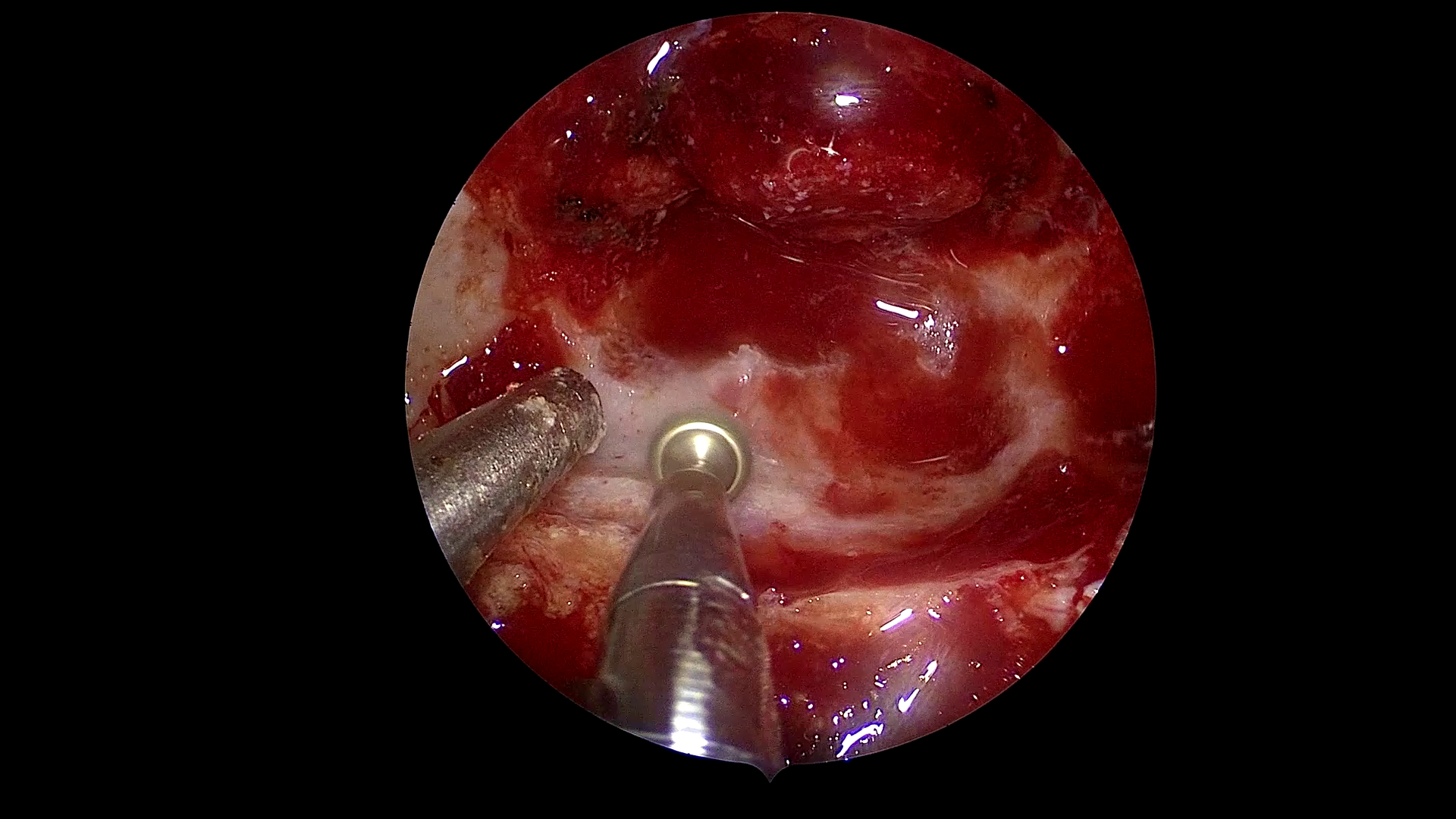}\hfill
\includegraphics[width=\exwidth]{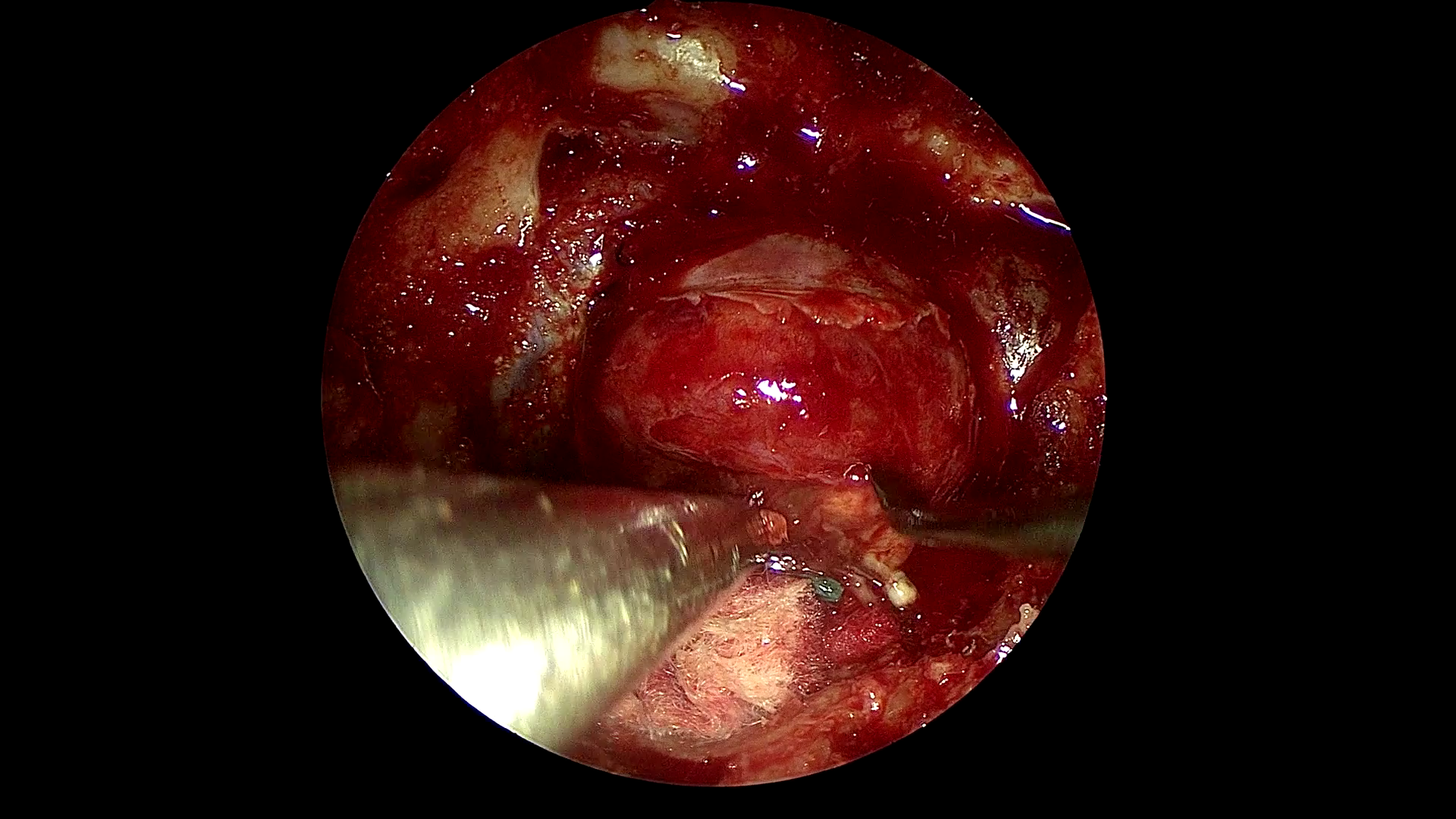}\hfill
\includegraphics[width=\exwidth]{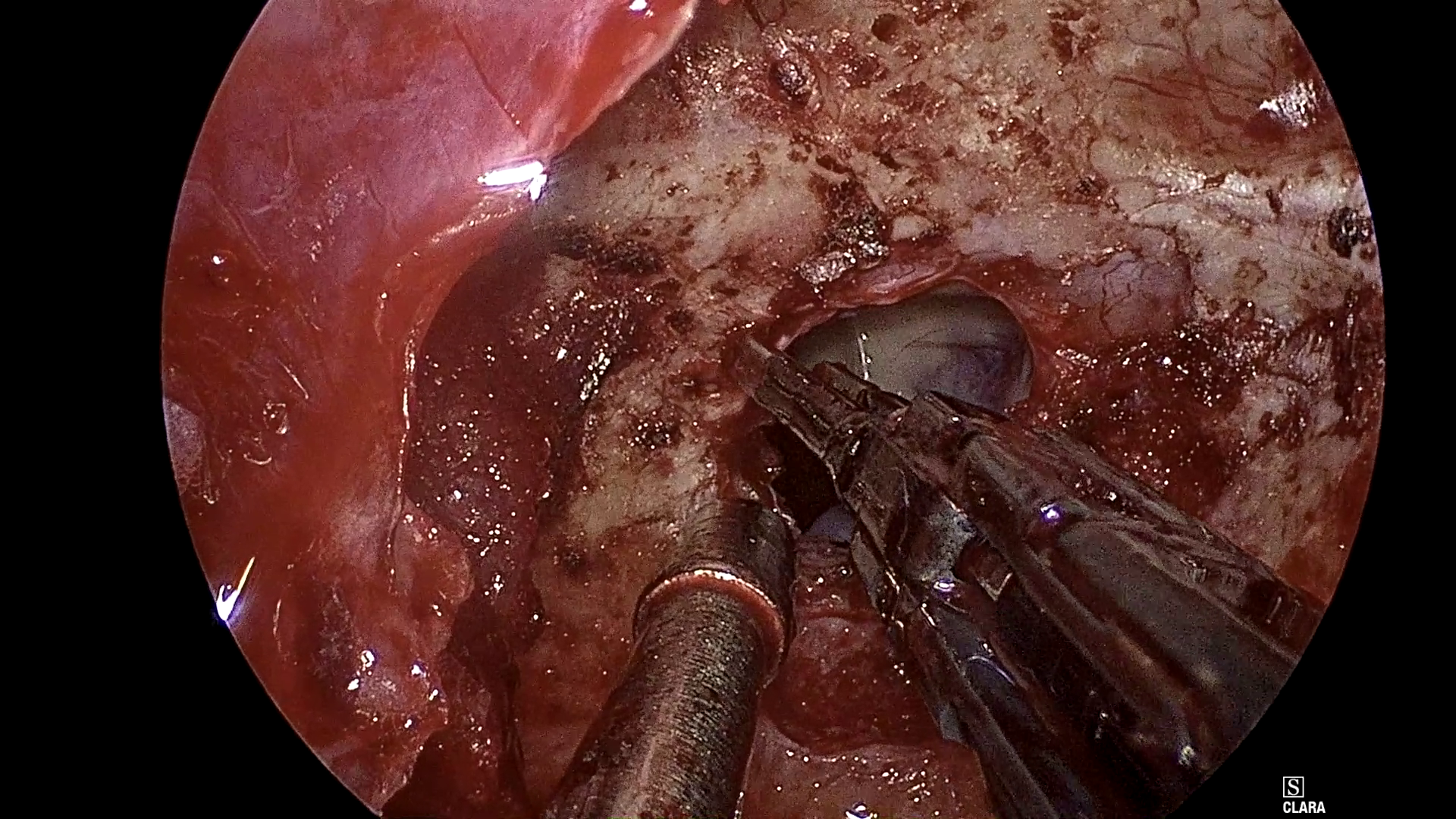}\hfill
\includegraphics[width=\exwidth]{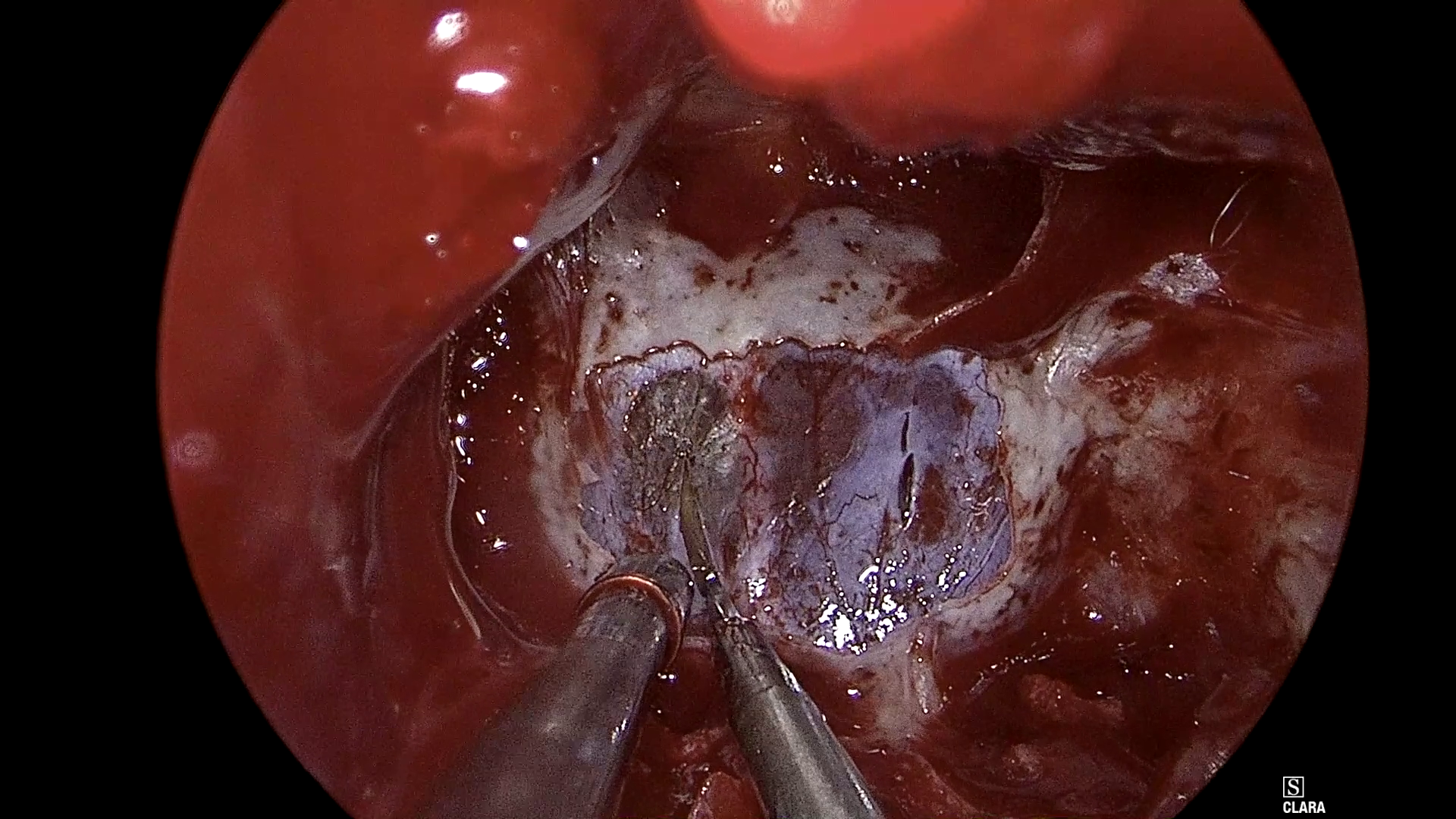}\hfill
\includegraphics[width=\exwidth]{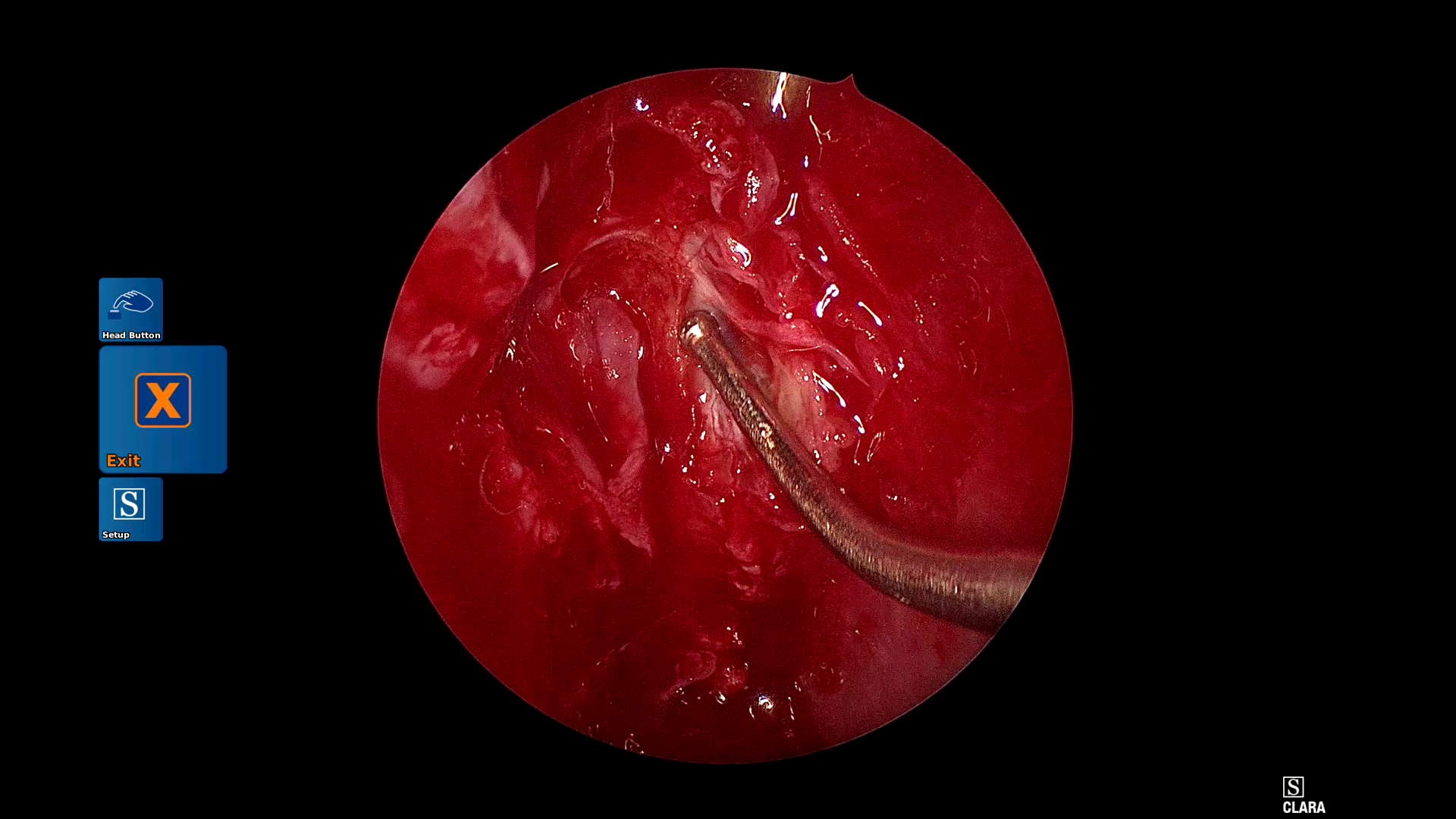}
\caption{Example frames from SDSC-EEA with zero-shot predictions from Gemma 3 27B. Top row: correct detections (left to right: Drill + Suction; Suction; Drill + Suction; no tools; no tools). Bottom row: incorrect detections, left to right: $y$\,=\,Drill, Suction; $\hat{y}$\,=\,Curette, Grasper, Irrigation, Monopolar Electrocautery, Suction; $y$\,=\,Cotton Patty, Rhoton Dissector, Suction; $\hat{y}$\,=\,Grasper, Monopolar Electrocautery, Suction; $y$\,=\,Bipolar Forceps, Suction; $\hat{y}$\,=\,Curette, Drill, Suction, Tissue shaver; $y$\,=\,Suction; $\hat{y}$\,=\,Grasper, Monopolar Electrocautery, Suction; $y$\,=\,Rhoton Dissector; $\hat{y}$\,=\,Monopolar Electrocautery, Suction.}
\label{fig:sample_frames}
\end{figure}

We evaluate zero-shot tool detection performance across 20 open-weight vision-language models spanning nearly three years of development (September 2023--April 2026). The complete list of models is shown in Table~\ref{tab:models_evaluated}.

\begin{table}[th]
\centering
\caption{Vision-language models evaluated for zero-shot surgical tool detection.}
\label{tab:models_evaluated}
\small
\begin{tabular}{lrcc}
\toprule
Model & Params~(B) & Release & MMBench \\
\midrule
Qwen3-VL-235B-A22B-Thinking~\citep{bai2025qwen3vltechnicalreport} & 235 & Sep~2025 & 90.6 \\
Qwen3-VL-32B-Instruct & 32 & Sep~2025 & 88.9 \\
Qwen3-VL-8B-Instruct & 8 & Sep~2025 & 85.0 \\
Qwen3-VL-4B-Instruct & 4 & Sep~2025 & 85.1 \\
Qwen3-VL-2B-Instruct & 2 & Sep~2025 & 77.8 \\
\midrule
Qwen2.5-VL-72B-Instruct~\citep{bai2025qwen25vltechnicalreport} & 72 & Mar~2025 & 87.8 \\
Qwen2.5-VL-32B-Instruct & 32 & Mar~2025 & 84.0 \\
Qwen2.5-VL-7B-Instruct & 7 & Mar~2025 & 82.2 \\
Qwen2.5-VL-3B-Instruct & 3 & Mar~2025 & 76.8 \\
\midrule
Qwen2-VL-72B-Instruct~\citep{wang2024qwen2vlenhancingvisionlanguagemodels} & 72 & Sep~2024 & 85.9 \\
Qwen2-VL-7B-Instruct & 7 & Sep~2024 & 81.0 \\
Qwen2-VL-2B-Instruct & 2 & Sep~2024 & 72.2 \\
\midrule
Gemma 3 27B-it~\citep{gemmateam2025gemma3technicalreport} & 27 & Mar~2025 & 78.9 \\
Gemma 3 12B-it & 12 & Mar~2025 & 74.6 \\
Gemma 3 4B-it & 4 & Mar~2025 & 66.4 \\
\midrule
MedGemma 3 27B-it \citep{sellergren2025medgemmatechnicalreport} & 27 & Jul~2025 & - \\
\midrule
Llama-3.2-90B-Vision~\citep{meta2024llama32vision} & 90 & Sep~2024 & 79.5 \\
Llama-3.2-11B-Vision & 11 & Sep~2024 & 67.5 \\
\midrule
LLaVA-1.5-13B~\citep{liu2024improvedbaselinesvisualinstruction} & 13 & Sep~2023 & 65.8 \\
\midrule
Gemma 4 31B-it~\citep{gemmateam2025gemma4technicalreport} & 31 & Apr~2026 & 90.9 \\
\bottomrule
\end{tabular}
\end{table}

Models span six families: Qwen (12 models across three generations), Gemma 3 (3 models), Gemma 4 (1 model), MedGemma 3 (1 model), Llama 3.2 Vision (2 models), and LLaVA 1.5 (1 model). Model sizes range from 2B to 235B parameters. MMBench~\citep{liu2024mmbenchmultimodalmodelallaround}, a holistic benchmark evaluating multimodal models across perception, reasoning, and knowledge, scores range from 65.8 (LLaVA 1.5) to 90.9 (Gemma 4 31B).

For each model, we prompt the model to identify all visible surgical tools from a list of 31 valid tool names and return predictions as a JSON object. The complete prompt template is provided in Appendix~\ref{sec:appendix_prompt}. Model outputs are validated against a strict schema; outputs that fail validation (malformed JSON, schema violations, or hallucinated tool names not in the ontology) are treated as empty predictions rather than silently excluded. The full output validation methodology is described in Appendix~\ref{sec:appendix_output_validation}.

Exact match accuracy is reported separately on the training set ($n=47{,}618$ frames from 53 procedures), validation set ($n=20{,}016$ frames from 13 procedures), and the full dataset (Table~\ref{tab:zero_shot}). Figure~\ref{fig:sample_frames} shows representative examples from our dataset, illustrating both successful and unsuccessful tool detection cases.

For the zero-shot results reported in Table~\ref{tab:zero_shot}, Figure~\ref{fig:accuracy_vs_params}, and Figure~\ref{fig:accuracy_vs_mmbench}, we use exact match accuracy and Jaccard similarity as primary metrics, with per-tool precision, recall, and F1 reported in Appendix~\ref{sec:appendix_zero_shot_metrics}. All evaluation metrics are defined in Section~\ref{subsec:metrics}. These results are analyzed in Section~\ref{subsec:zero_shot}.


\subsection{LoRA Fine-Tuning}
\label{subsec:lora_methods}
\label{subsec:lora_formatting_methods}
\label{subsec:classification_head_methods}
\label{subsec:rank_sweep_methods}

We fine-tune Gemma 3 27B using Low-Rank Adaptation (LoRA)~\citep{hu2021loralowrankadaptationlarge} with adapters applied to attention projection matrices in both the language model and vision encoder. We evaluate three configurations:

JSON generation (Figure~\ref{fig:s19_loss}): The model learns to produce structured JSON outputs in the format \texttt{\{"detected\_tools": ["Tool1", "Tool2"]\}} via supervised fine-tuning.

Classification head (Figure~\ref{fig:s16_loss}): We replace JSON generation with a single-layer linear classification head that maps mean-pooled hidden states to 31 output logits, trained with binary cross-entropy loss. At inference, predictions are obtained by thresholding sigmoid outputs at 0.5. This approach enables continuous prediction scores for ROC-AUC and AUPRC metrics and requires only a single forward pass rather than autoregressive generation.

Rank sweep (Figure~\ref{fig:accuracy_vs_rank}, Table~\ref{tab:rank_sweep}): To investigate whether increasing model capacity improves generalization, we sweep LoRA rank from $r=2$ to $r=1024$, varying trainable parameters by nearly three orders of magnitude (4.7M to 2.4B parameters).

All three configurations use the same procedure-level train/validation split described in Section~\ref{subsec:dataset}. Full configuration details (ranks, learning rates, batch sizes, and compute requirements) are provided in Appendix~\ref{sec:appendix_lora_config}.

\subsection{Specialized Supervised Model}
\label{subsec:yolo_methods}

As a supervised baseline, we train YOLOv12-m \citep{tian2025yolov12attentioncentricrealtimeobject}, a state-of-the-art object detection model with 26M parameters. Unlike VLMs, which perform set-based multi-label classification, YOLO directly predicts bounding boxes with associated class labels and confidence scores. On SDSC-EEA, we train YOLO on the per-frame bounding-box annotations provided by the SDSC annotation team (one box per visible tool, in the standard YOLO normalized $(x_c, y_c, w, h)$ format). We train using default YOLO hyperparameters; the full configuration is provided in Appendix~\ref{sec:appendix_yolo_config}.

\paragraph{Adapting YOLO to presence-only datasets.} The three external public datasets we use---CholecT50, PitVis-2023, and SurgVU---ship per-frame multi-label tool-presence annotations only; none release per-frame bounding boxes for the full set of training videos. Because the YOLO training pipeline requires a $(x_c, y_c, w, h)$ target for each tool instance, we follow the standard workaround for tool-presence-only datasets and write a synthetic full-frame box ($0.5\ 0.5\ 1.0\ 1.0$) for every present tool in each frame; absent tools contribute no box. Under this supervision, YOLO's localization head receives no spatial signal and the model is trained, in effect, as a multi-label image classifier built on the YOLOv12-m backbone (we keep the YOLO name to make the implementation unambiguous). At inference we still threshold YOLO's per-class confidence at $\geq 0.25$ and convert the resulting set of detected classes into a per-frame tool set, exactly as we do on SDSC-EEA. Section~\ref{subsec:cholect50_methods}, \ref{subsec:pitvis_methods}, and~\ref{subsec:surgvu_methods} reference this protocol when they describe YOLO training on each external dataset, and Appendix~\ref{sec:appendix_resnet} provides an additional ResNet-50 multi-label classifier baseline trained on the same set-level signal so the relative gain from the YOLO backbone (rather than from bounding-box supervision) can be assessed directly.

To enable direct comparison with VLMs, we convert YOLO's per-frame bounding-box predictions into tool sets: for each frame, we collect the unique set of tool classes with confidence $\geq 0.25$ and compare against the ground truth tool set. This allows us to compute exact match accuracy, Jaccard similarity, top-1 accuracy, and per-tool precision/recall/F1 on the same basis as VLM-based classifiers. Results, including a per-tool comparison with Gemma (Table~\ref{tab:yolo_vs_gemma}), are reported in Section~\ref{subsec:yolo_performance}.

\subsection{External Dataset: CholecT50}
\label{subsec:cholect50_methods}

To evaluate generalization to an independent surgical domain, we use CholecT50~\citep{cholec}, a publicly available dataset of laparoscopic cholecystectomy procedures. CholecT50 comprises 50 videos with frame-level annotations for 6 surgical instruments (grasper, bipolar, hook, scissors, clipper, irrigator), 10 surgical verbs, 15 anatomical targets, and 100 instrument-verb-target triplets. We focus exclusively on instrument detection to maintain consistency with our primary evaluation.

The dataset contains 100,863 annotated frames. We perform an 80/20 train/validation split at the video level to prevent data leakage, yielding 80,940 training frames (40 videos) and 19,923 validation frames (10 videos). The majority class baseline---predicting the most common tool set (grasper, hook) for every frame---achieves 34.76\% exact match accuracy on the validation set.

We evaluate zero-shot performance using Gemma 3 27B, fine-tune with LoRA and a classification head using the same configuration as Section~\ref{subsec:classification_head_methods}, conduct a LoRA rank sweep ($r \in \{2, 4, 8, 16, 32, 64, 128, 256, 512, 1024\}$) using the same protocol as Section~\ref{subsec:rank_sweep_methods}, and train YOLOv12-m using the same setup as Section~\ref{subsec:yolo_methods}. Because CholecT50 ships only per-frame tool-presence labels, the YOLO training data on this dataset uses synthetic full-frame boxes following the protocol described in Section~\ref{subsec:yolo_methods}; YOLO is therefore being trained as a YOLO-backbone multi-label classifier on CholecT50, not as a true bounding-box detector. Results, including Table~\ref{tab:cholect50_yolo_vs_gemma} and Figure~\ref{fig:cholect50_accuracy_vs_rank}, are reported in Section~\ref{subsec:cholect50_results}.

\subsection{External Dataset: PitVis-2023}
\label{subsec:pitvis_methods}

To further test generalization---and in particular to evaluate on a second neurosurgical dataset distinct from SDSC-EEA---we use PitVis-2023~\citep{pitvis}, a publicly available dataset of endoscopic transsphenoidal pituitary surgery procedures released as part of the EndoVis 2023 PitVis Challenge. The official release advertises 25 videos, but the released annotation directory ships only 24 per-video annotation files: \texttt{annotations\_19.csv} is missing from the public dataset. We therefore exclude video~19 (which has no published labels) and use the 24 videos with both video and annotations, which carry frame-level annotations for 18 surgical instrument classes (bipolar forceps, cottle, cup forceps, dural scissors, freer elevator, haemostatic foam, irrigation syringe, kerrisons, micro-Doppler, nasal cutting forceps, pituitary rongeurs, retractable knife, ring curette, spatula dissector, stealth pointer, suction, surgical drill, tissue glue), as well as 14 surgical workflow steps. As with CholecT50, we focus exclusively on instrument detection.

The 24 annotated videos contain 115,562 annotated frames. We perform a video-level train/validation split (19 train videos, 5 validation videos) yielding 84,666 training frames and 30,896 validation frames, preventing within-procedure data leakage. The majority class baseline---predicting the most common tool set (the empty set, corresponding to no instruments visible) for every frame---achieves 39.63\% exact match accuracy on the validation set.

We evaluate zero-shot performance using Gemma 3 27B and the same eleven frontier VLMs evaluated on CholecT50, fine-tune with LoRA and a classification head using the same configuration as Section~\ref{subsec:classification_head_methods}, conduct a LoRA rank sweep ($r \in \{2, 4, 8, 16, 32, 64, 128, 256, 512, 1024\}$) using the same protocol as Section~\ref{subsec:rank_sweep_methods}, and train YOLOv12-m using the same setup as Section~\ref{subsec:yolo_methods}. As on CholecT50, PitVis-2023 ships only per-frame tool-presence labels, so the YOLO training data on this dataset uses synthetic full-frame boxes per the protocol in Section~\ref{subsec:yolo_methods} and YOLO operates as a YOLO-backbone multi-label classifier rather than a true bounding-box detector. Results, including Table~\ref{tab:pitvis_yolo_vs_gemma} and Figure~\ref{fig:pitvis_accuracy_vs_rank}, are reported in Section~\ref{subsec:pitvis_results}.

\subsection{External Dataset: SurgVU}
\label{subsec:surgvu_methods}

To evaluate generalization to a third independent surgical domain---and one that is markedly different from the endoscopic neurosurgical procedures of SDSC-EEA and PitVis-2023---we use SurgVU~\citep{surgvu}, a publicly available dataset of robotic-assisted surgery training sessions on porcine tissue, released as part of the EndoVis 2024 SurgVU Challenge. The dataset consists of 280 video clips from 155 training sessions, in which trainee and expert surgeons perform standardized exercises (e.g.\ suturing, uterine horn, suspensory ligaments) on a porcine model using the da~Vinci robotic platform. Tool presence labels are derived from temporal install/uninstall events automatically harvested from the robot arms and converted to per-frame multi-label tool presence. The accompanying paper~\citep{surgvu} highlights twelve main exercise tools (needle driver, cadiere forceps, prograsp forceps, monopolar curved scissors, bipolar forceps, stapler, force bipolar, vessel sealer, permanent cautery hook/spatula, clip applier, tip-up fenestrated grasper, grasping retractor); the released \texttt{tools.csv} additionally records five rare classes that were occasionally installed during the exercises (bipolar dissector, potts scissors, suction irrigator, synchroseal, tenaculum forceps), giving 17 distinct instrument classes in the released label set. We use the full 17-class taxonomy as released, retaining the rare classes for completeness; three of them (bipolar dissector, potts scissors, tenaculum forceps) have zero positive frames in our validation split and are noted as such in all per-tool tables. We focus exclusively on instrument detection.

We extract one frame every 30 seconds from the released videos. SurgVU videos contain burnt-in heads-up display (HUD) overlays that directly leak the ground-truth labels: a ``TRAINING INSTRUMENT'' warning naming the active tool is rendered along the top of every frame, and a four-arm da~Vinci tool readout (e.g.\ ``1~PROGRASP FORCEPS \ldots'') is rendered along the bottom. To prevent this leakage from trivializing the task for any model, we crop the top 10\% and bottom 10\% of pixels from every extracted frame before saving; the cropped JPEGs are what every model (zero-shot, fine-tuned, and YOLO) sees. The SurgVU release also includes a separate official validation set (\texttt{cat1\_test\_set\_public.zip}) that contains bounding-box annotations for only 8 of the 12 main tools and is intended for the MICCAI tool-detection challenge; we do not use that set, because doing so would prevent us from evaluating the same 17-class tool-presence taxonomy across train and validation. Instead, we perform a session-level 80/20 train/validation split of the 155 training sessions (124 train sessions, 31 validation sessions) yielding 81,751 training frames and 18,919 validation frames, preventing within-session data leakage. The majority class baseline---predicting the most common tool set (the empty set, corresponding to no instruments visible) for every frame---achieves 16.94\% exact match accuracy on the validation set.

We evaluate zero-shot performance using Gemma 3 27B and the same eleven frontier VLMs evaluated on CholecT50 and PitVis-2023, fine-tune with LoRA and a classification head using the same configuration as Section~\ref{subsec:classification_head_methods}, conduct a LoRA rank sweep ($r \in \{2, 4, 8, 16, 32, 64, 128, 256, 512, 1024\}$) using the same protocol as Section~\ref{subsec:rank_sweep_methods}, and train YOLOv12-m using the same setup as Section~\ref{subsec:yolo_methods}. The publicly released SurgVU labels are also tool-presence-only (the official \texttt{cat1} bounding-box validation set is held back for the MICCAI challenge and does not cover the full 17-class taxonomy or the training videos), so the YOLO training data on SurgVU again uses synthetic full-frame boxes per the protocol in Section~\ref{subsec:yolo_methods}, and YOLO operates as a YOLO-backbone multi-label classifier rather than a true bounding-box detector on this dataset. The Gemma and YOLO evaluations, together with five of the eleven frontier evaluations, are run on the full 18,919 validation frames drawn from the same 31 validation sessions; the remaining six frontier models (Claude Fable 5, Claude Fable 5.1, Gemini 3.8 Flash, GPT-5.6 Sol, Kimi K3, and Qwen3.8 Max) are evaluated on random subsamples of 1{,}000 of these frames and are marked with an asterisk ($^*$) in all results tables. Results, including Table~\ref{tab:surgvu_yolo_vs_gemma} and Figure~\ref{fig:surgvu_accuracy_vs_rank}, are reported in Section~\ref{subsec:surgvu_results}.

\subsection{Beyond Detection: Four Bounding-Box-Free Annotation Tasks}
\label{subsec:newtasks_methods}

Object detection is only one of the major annotation tasks performed on surgical video, and its bounding-box structure could favor detector architectures such as YOLO. To test whether our conclusions are an artifact of the task, we evaluate four bounding-box-free annotation tasks on four additional public datasets, chosen to cover the temporal-workflow and intent dimensions of surgical video annotation.

\paragraph{Datasets and splits.} PitVQA~\citep{he2024pitvqa} provides frame-level surgical workflow recognition---3 phases and 14 steps---in endoscopic pituitary surgery (109{,}099 frames from 25 videos, split by video into 84{,}332 training and 24{,}767 validation frames following the official 20/5 video split; the videos originate from the same public challenge as PitVis-2023, but the labels and task differ). SAR-RARP50~\citep{psychogyios2023sarrarp50} provides fine-grained surgical gesture recognition---8 suturing gestures---in robot-assisted radical prostatectomy; we score one mid-segment frame per contiguous gesture segment, split by operation (70/30, seed 42) into 1{,}563 training and 636 validation frames from 35 and 15 operations, respectively. CholecT50~\citep{cholec} additionally provides frame-level surgical action recognition through the verb component of its action-triplet annotations---a multi-label intent-analysis task in which the model must identify every action currently being performed among nine instrument--tissue actions (plus an explicit \texttt{idle} class). We reuse the video-level split from Section~\ref{subsec:cholect50_methods} (40 training and 10 validation videos; 80{,}940 training and 19{,}923 validation frames), so no video crosses splits and the task shares images, but not labels, with the tool-detection benchmark. DSAD~\citep{carstens2023dresden} provides multi-label anatomical-structure presence over 12 abdominal structures in robot-assisted rectal resection (official splits; 7{,}889 training and 1{,}978 validation frames). None of these tasks involves bounding boxes, removing the structural advantage that object detection confers on detector architectures. Full label ontologies and split details are given in Appendix~\ref{sec:appendix_newtasks}.

\paragraph{Models and protocol.} We use a uniform frame-level protocol: every model receives a single frame and the task's label ontology, so no model class benefits from temporal context or bounding-box supervision. We compare five model classes. (i)~Zero-shot open-weight VLMs: Gemma~3 27B-it, the strongest open-weight zero-shot model in the tool-detection evaluation, and Qwen3.8 27B, the most recent open-weight release of the Qwen-VL series~\citep{qwen2026qwen3827b}, prompted for JSON exactly as in the tool-detection protocol (prompt templates in Appendix~\ref{sec:appendix_newtasks_prompts}). (ii)~Eight frontier proprietary VLMs from the model generation current at the time of writing (Claude Fable~5, Claude Opus~5, Claude Sonnet~5, Gemini~3.1 Pro, Gemini~3.7 Flash, GPT-5.6 Luna, GPT-5.6 Sol, and GPT-5.6 Terra), evaluated with identical prompts through their APIs (all four datasets are public); on the three larger validation splits (PitVQA, CholecT50 action recognition, and DSAD) frontier models are scored on a random seed-42 subsample of 1{,}000 frames. (iii)~A surgical vision foundation model, LemonFM~\citep{che2026lemonlargeendoscopicmonocular} (ConvNeXt-Large backbone pretrained at scale on endoscopic video): we freeze the backbone and train a linear probe per task. (iv)~Gemma~3 27B fine-tuned with LoRA on each task's training split, with both output heads from Section~\ref{subsec:lora_methods}: the autoregressive JSON generator ($r=16$; two to three epochs, scaled inversely with training-set size) and the linear classification head ($r=128$; 10 epochs). (v)~Small supervised specialists: ResNet-50~\citep{he2016deep} (25.6M parameters, ImageNet-1K initialization) trained on each task's training split, with multi-label sigmoid heads for DSAD and CholecT50 action recognition, two softmax heads for PitVQA, and a single softmax head for SAR-RARP50---so the specialists on these tasks involve no detection components of any kind. Outputs failing schema validation are counted as errors, as above. Exact match requires the full predicted label set to equal the ground truth (for PitVQA, both the phase and the step); micro-averaged F1 and 95\% bootstrap confidence intervals ($B=1{,}000$) are computed as in Section~\ref{subsec:metrics}, and every model class is compared against a majority class baseline that predicts the most frequent training-split label(s) on every validation frame. Training configurations for every model class are listed in Appendix~\ref{sec:appendix_newtasks}. Results are reported in Section~\ref{subsec:newtasks}.

{\sloppy Authors maintain an up-to-date version of the benchmark leaderboard, including models evaluated after the present study was finalized, at \url{https://skblv.github.io/neurosurgery-video-eval-website/}.\par}

\subsection{Evaluation Metrics}
\label{subsec:metrics}

We report the following metrics throughout. Exact match accuracy is the percentage of frames where the predicted tool set exactly matches the ground truth; this is a strict metric that penalizes any false positive or false negative. Jaccard similarity is computed for each frame as $J = |P \cap G| / |P \cup G|$ where $P$ is the predicted set and $G$ is the ground truth set, and we report the mean across all frames. We also compute per-tool precision, recall, and F1 scores as standard binary classification metrics independently for each tool class. For models with continuous prediction scores (classification head), we additionally report ROC-AUC (area under the receiver operating characteristic curve) and AUPRC (area under the precision-recall curve) per tool class, as well as macro-averaged values across tools present in the validation set. Per-class accuracy for zero-prevalence classes is meaningless (a model predicting all negatives achieves 100\% accuracy) and is excluded from macro-averaged metrics.

To enable direct comparison between YOLO and VLM-based classifiers, we additionally report top-1 accuracy: the fraction of frames where the tool with the highest predicted probability is present in the ground truth set. Both YOLO (via class confidence scores) and the Gemma classifier (via sigmoid outputs) produce explicit per-tool probabilities, making this metric computable for both. However, top-1 accuracy cannot be computed for generative VLM outputs, which produce unordered tool lists without per-tool probability scores. This metric isolates the model's ability to identify the single most salient tool in each frame, a prerequisite for reliable surgical assistance.

For 95\% confidence intervals on exact match accuracy, we use bootstrap resampling with $B = 1{,}000$ iterations. For a dataset of $N$ frames, we resample $N$ observations with replacement from the binary correct/incorrect results and compute the mean for each of the $B$ bootstrap samples; the 2.5th and 97.5th percentiles form the confidence interval. For the four bounding-box-free tasks of Section~\ref{subsec:newtasks_methods}, we additionally report micro-averaged F1 over the task's label set, with bootstrap confidence intervals computed in the same way; for the single-label SAR-RARP50 task, exact match and micro-F1 coincide up to parse failures.

%% file: sections/3_results.tex
\section{Results}
\label{sec:results}

We present results in seven parts. Section~\ref{subsec:zero_shot} establishes the baseline: zero-shot VLMs fail to exceed a trivial majority class baseline despite three years of scaling. Given this failure, the next three sections ask whether adaptation can close the gap. Sections~\ref{subsec:lora_formatting} and~\ref{subsec:lora_classification_head} explore two parallel fine-tuning strategies---JSON generation and a classification head---that both improve substantially over zero-shot but plateau well below human-level accuracy. Section~\ref{subsec:lora_rank_sweep} then tests whether this plateau is due to insufficient capacity by scaling LoRA rank by nearly three orders of magnitude; training accuracy saturates near 99\% while validation accuracy remains below 40\%, indicating that the bottleneck is not model capacity. Section~\ref{subsec:yolo_performance} compares against YOLOv12-m, a specialized 26M-parameter object detection model that outperforms all VLM-based approaches with 1,000$\times$ fewer parameters. Section~\ref{subsec:robustness} replicates the key experiments on three independent public datasets---CholecT50 (laparoscopic cholecystectomy), PitVis-2023 (endoscopic pituitary surgery), and SurgVU (robotic-assisted training sessions on porcine tissue)---and finds the same broad patterns across all four surgical domains. Finally, Section~\ref{subsec:newtasks} tests whether these conclusions are an artifact of the detection task by evaluating four bounding-box-free annotation tasks---workflow, gesture, action, and anatomy recognition---and finds that the same specialist-over-generalist hierarchy replicates on every dataset.

\subsection{Zero-shot accuracy of open-weight models does not surpass the majority class baseline}
\label{subsec:zero_shot}

\paragraph{Takeaways}
Even for larger VLMs, in the zero-shot setting, performance stays at or near the majority-class baseline. Progress on general multimodal benchmarks and parameter scale does not transfer reliably to this surgical perception task.

\begin{figure}[H]
\centering
\includegraphics[width=0.75\textwidth]{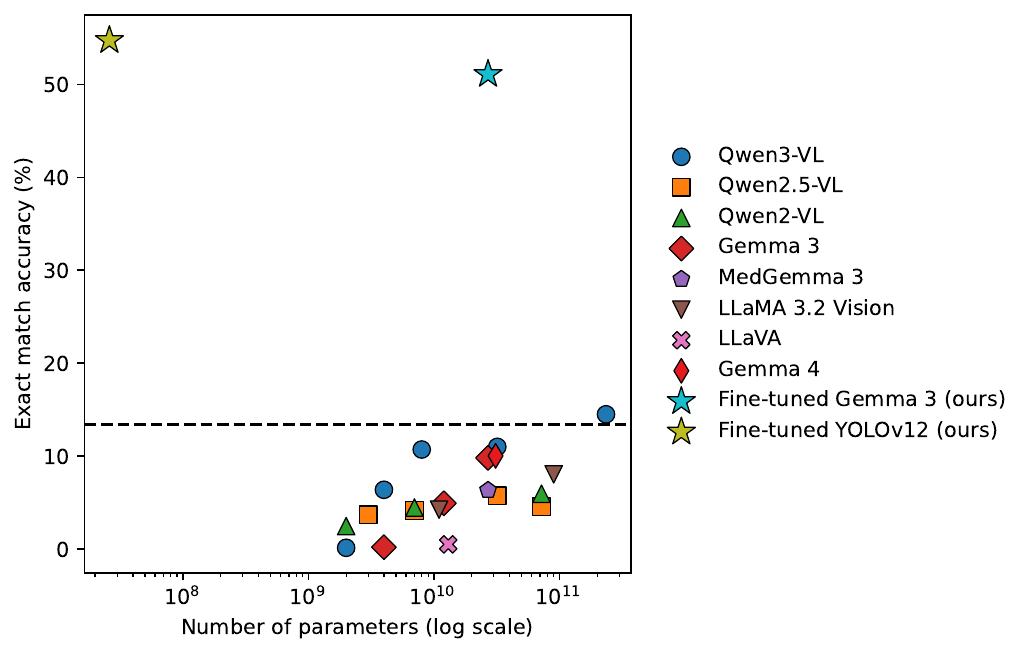}
\caption{Exact-match accuracy on the SDSC-EEA validation set ($n=20{,}016$) as a function of model parameter count. Colors and marker shapes denote model families. The black dashed line indicates the majority-class baseline (13.4\%). Accuracy exhibits a positive but strongly sublinear relationship with parameter count; the relationship is family-dependent, with Qwen models consistently outperforming similarly-sized Gemma and Llama models.}
\label{fig:accuracy_vs_params}
\end{figure}

\begin{figure}[H]
\centering
\includegraphics[width=0.75\textwidth]{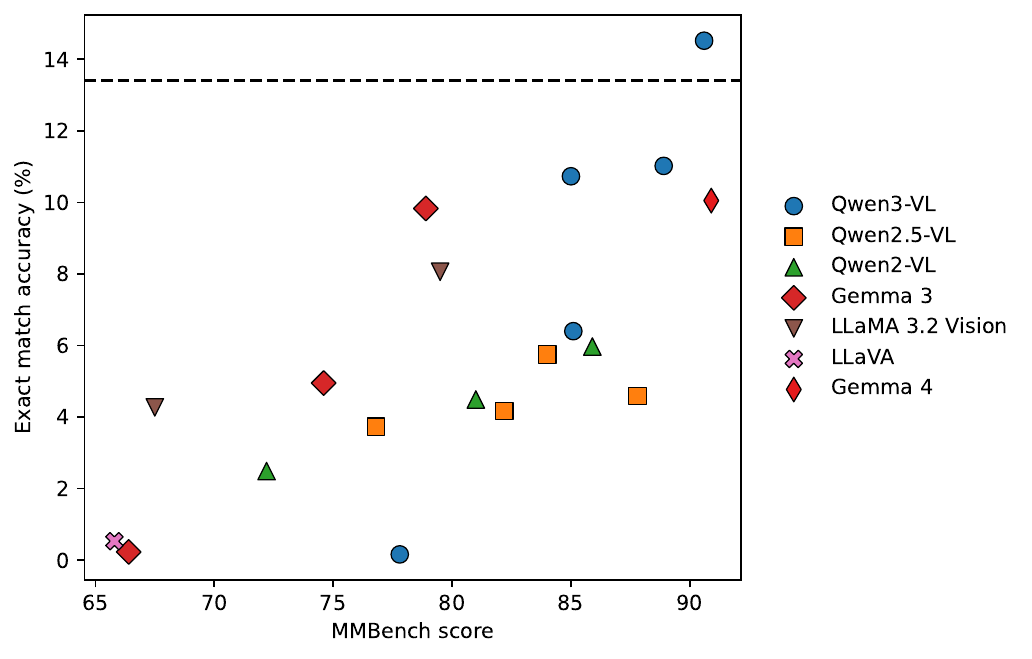}
\caption{Zero-shot exact-match accuracy on the SDSC-EEA validation set ($n=20{,}016$) plotted against MMBench score. Colors and marker shapes denote model families. The black dashed line indicates the majority-class baseline (13.4\%). Higher MMBench scores correlate weakly with higher tool detection accuracy: the best tool-detection model (Qwen3-VL-235B, MMBench 90.6) achieves only 14.52\%, while the highest MMBench scorer (Gemma 4 31B, MMBench 90.9) achieves just 10.05\%---both far below fine-tuned models (51.08\%, Section~\ref{subsec:lora_classification_head}).}
\label{fig:accuracy_vs_mmbench}
\end{figure}

\paragraph{Detailed Results.} We evaluate zero-shot tool detection performance across 20 open-weight vision-language models (Section~\ref{subsec:zero_shot_methods}) released between September 2023 and April 2026. Despite dramatic increases in model scale, from LLaVA 1.5 13B (2023) to Qwen3-VL-235B (2025) and Gemma 4 31B (2026), and substantial improvements on general vision benchmarks, no model meaningfully surpasses the majority class baseline on the validation set.

Table~\ref{tab:zero_shot} reports exact match accuracy for all models; no model meaningfully surpasses the majority class baseline. As shown in Figure~\ref{fig:accuracy_vs_mmbench}, higher MMBench scores are weakly correlated with higher performance on the tool detection benchmark in our dataset. However, even the best performing model on tool detection, Qwen3-VL-235B (MMBench 90.6, validation accuracy 14.52\%), significantly underperforms the fine-tuned Gemma 3 27B in Section~\ref{subsec:lora_classification_head} (51.08\% validation exact match accuracy). Strikingly, Gemma 4 31B---the highest MMBench scorer in our evaluation (90.9)---achieves only 10.05\% on tool detection, below the majority class baseline. This further suggests that there are surgical visual capabilities that go beyond what can be measured by multi-purpose benchmarks like MMBench.

Notably, MedGemma 3 27B-it, which is described as a model optimized for medicine, underperforms Gemma 3 27B-it---a sibling that MedGemma is based on---on the validation set (6.36\% vs.\ 9.83\%). Per-tool classification metrics (precision, recall, F1) for all 20 evaluated zero-shot models are provided in Appendix~\ref{sec:appendix_zero_shot_metrics}. Appendix~\ref{sec:appendix_validation_failures} shows representative failed outputs, which are dominated by hallucinated tool names rather than formatting errors.

\subsection{LoRA fine-tuning improves tool detection modestly but remains below human-level}
\label{subsec:lora_formatting}

\paragraph{Takeaways} Task-specific fine-tuning improves performance relative to zero-shot evaluation, but it does not close the generalization gap on held-out procedures.

\begin{figure}[H]
\centering
\includegraphics[width=\textwidth]{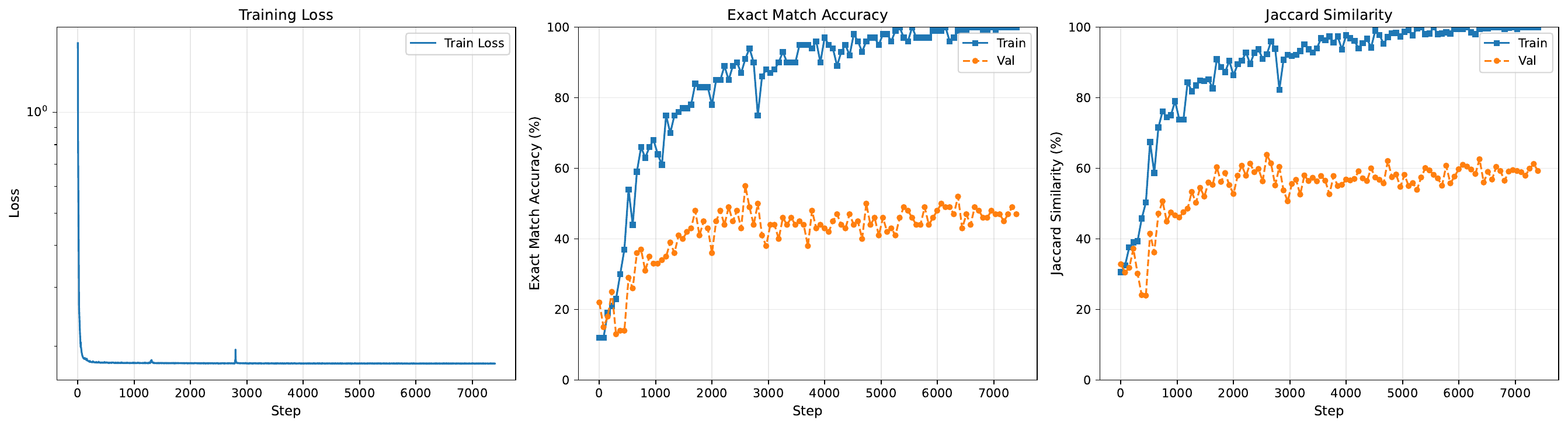}
\caption{Training dynamics for LoRA fine-tuning with JSON output on SDSC-EEA ($r=1024$). Left: Training loss (log scale) decreases steadily, confirming the model learns the structured output format. Center: Exact match accuracy. Right: Jaccard similarity. Both accuracy and Jaccard show a persistent gap between training and validation performance, indicating limited generalization to held-out procedures. Metrics are computed on fixed random subsets of 100 frames from each set, evaluated 100 times throughout training.}
\label{fig:s19_loss}
\end{figure}

\paragraph{Detailed Results} Given that zero-shot models fail at surgical tool detection regardless of scale, we next ask whether task-specific fine-tuning can bridge the gap. We fine-tune Gemma 3 27B with LoRA adapters to generate structured JSON predictions (Section~\ref{subsec:lora_methods}). Figure~\ref{fig:s19_loss} shows training and validation loss curves over 10 epochs.

\begin{landscape}
\begin{table}[H]
\centering
\caption{Zero-shot tool detection exact match accuracy (\%) on SDSC-EEA for all evaluated VLMs with 95\% bootstrap confidence intervals. Train ($n=47{,}618$ frames, 53 procedures), validation ($n=20{,}016$ frames, 13 procedures), and full dataset. Output validation failures are counted as incorrect predictions. The majority class baseline, which predicts the most common tool set for every frame, achieves 13.41\% exact match accuracy on the validation set. The pre-training baseline (Gemma 3 27B) achieves 9.83\% validation exact match accuracy (95\% CI: 9.43\%--10.21\%). Validation accuracy ranges from 0.11\% (Qwen3-VL-2B) to 14.52\% (Qwen3-VL-235B-A22B-Thinking); only Qwen3-VL-235B marginally surpasses the majority class baseline. Gemma 4 31B-it, a 2026 model with strong general benchmark improvements over Gemma 3, achieves 10.05\% validation exact match accuracy---comparable to Gemma 3 27B (9.83\%). Output validation failure rates range from 0.8\% (Gemma 3 27B, Qwen3-VL-8B) to 41.7\% (Qwen2-VL-2B).}
\label{tab:zero_shot}
\small
\begin{tabular}{lrllllll}
\toprule
 & Params & \multicolumn{2}{c}{Train} & \multicolumn{2}{c}{Validation} & \multicolumn{2}{c}{Full} \\
\cmidrule(lr){3-4} \cmidrule(lr){5-6} \cmidrule(lr){7-8}
Model & (B) & EM \% & 95\% CI & EM \% & 95\% CI & EM \% & 95\% CI \\
\midrule
Qwen3-VL-235B-A22B-Thinking & 235 & 17.04 & 16.69--17.39 & 14.52 & 14.06--15.02 & 16.28 & 16.01--16.55 \\
Qwen3-VL-32B-Instruct & 32 & 16.02 & 15.70--16.34 & 11.04 & 10.57--11.45 & 14.58 & 14.32--14.84 \\
Qwen3-VL-8B-Instruct & 8 & 12.08 & 11.80--12.38 & 10.73 & 10.32--11.13 & 11.68 & 11.46--11.92 \\
Qwen3-VL-4B-Instruct & 4 & 11.10 & 10.82--11.38 & 6.37 & 6.04--6.70 & 9.72 & 9.51--9.94 \\
Qwen3-VL-2B-Instruct & 2 & 0.29 & 0.24--0.34 & 0.11 & 0.07--0.16 & 0.24 & 0.20--0.27 \\
\midrule
Qwen2.5-VL-72B-Instruct & 72 & 10.27 & 10.01--10.54 & 4.56 & 4.27--4.87 & 8.59 & 8.40--8.79 \\
Qwen2.5-VL-32B-Instruct & 32 & 10.08 & 9.83--10.36 & 5.43 & 5.13--5.74 & 8.72 & 8.52--8.93 \\
Qwen2.5-VL-7B-Instruct & 7 & 6.54 & 6.33--6.75 & 3.48 & 3.25--3.75 & 5.64 & 5.49--5.82 \\
Qwen2.5-VL-3B-Instruct & 3 & 5.43 & 5.23--5.64 & 2.90 & 2.68--3.14 & 4.69 & 4.53--4.86 \\
\midrule
Qwen2-VL-72B-Instruct & 72 & 8.70 & 8.46--8.95 & 5.70 & 5.39--6.00 & 7.82 & 7.64--8.01 \\
Qwen2-VL-7B-Instruct & 7 & 5.13 & 4.93--5.32 & 3.35 & 3.09--3.60 & 4.61 & 4.46--4.76 \\
Qwen2-VL-2B-Instruct & 2 & 2.29 & 2.15--2.42 & 1.48 & 1.31--1.64 & 2.05 & 1.94--2.15 \\
\midrule
Gemma 3 27B-it & 27 & 5.61 & 5.40--5.81 & 9.83 & 9.43--10.21 & 6.85 & 6.66--7.02 \\
Gemma 3 12B-it & 12 & 5.20 & 5.01--5.39 & 4.95 & 4.65--5.25 & 5.13 & 4.97--5.29 \\
Gemma 3 4B-it & 4 & 0.56 & 0.50--0.63 & 0.23 & 0.16--0.30 & 0.46 & 0.42--0.51 \\
\midrule
MedGemma 3 27B-it & 27 & 5.41 & 5.20--5.60 & 6.36 & 6.02--6.70 & 5.68 & 5.51--5.86 \\
\midrule
Llama-3.2-90B-Vision & 90 & 10.62 & 10.33--10.91 & 8.06 & 7.68--8.45 & 9.85 & 9.62--10.09 \\
Llama-3.2-11B-Vision & 11 & 0.33 & 0.28--0.38 & 0.19 & 0.13--0.25 & 0.29 & 0.25--0.33 \\
\midrule
LLaVA-1.5-13B & 13 & 1.66 & 1.55--1.77 & 0.53 & 0.44--0.64 & 1.33 & 1.23--1.41 \\
\midrule
Gemma 4 31B-it & 31 & 18.44 & 18.09--18.80 & 10.05 & 9.65--10.46 & 15.94 & 15.67--16.22 \\
\midrule
Majority class baseline & -- & 11.49 & -- & 13.41 & -- & 12.06 & -- \\
\bottomrule
\end{tabular}
\end{table}
\end{landscape}

After 10 epochs, the fine-tuned model achieves 47.63\% exact match accuracy (95\% CI: 46.97\%--48.34\%) and 57.34\% Jaccard similarity on the validation set ($n=20{,}016$). This represents a substantial improvement over both the majority class baseline (13.41\% exact match, 31.91\% Jaccard) and the pre-training baseline (9.83\% exact match, 25.98\% Jaccard).

Table~\ref{tab:s19_tool_metrics} in Appendix~\ref{sec:appendix_s19_metrics} shows per-tool precision and recall. The model learns to detect several tools with high F1 scores (Suction Coagulator: 0.989, Drill: 0.876, Suction: 0.809) but completely fails on others (Suction microdebrider: 0\% recall despite 497 ground truth instances in validation). This discrepancy arises from the procedure-based train/validation split: tools that appear predominantly in validation procedures were rarely seen during training (Table~\ref{tab:tool_distribution}). For example, Suction microdebrider has only 94 training instances versus 497 in validation, and Aspirating dissector has 88 training instances versus 2{,}319 in validation. 

Qualitative analysis reveals that fine-tuned models produce syntactically correct JSON outputs with valid tool names (eliminating output validation failures common in zero-shot outputs), but generalization to unseen tool distributions remains poor.

\subsection{LoRA with classification head learns in-sample but fails to generalize out-of-sample}
\label{subsec:lora_classification_head}

\paragraph{Takeaways}
Dedicated classification objectives are more effective than autoregressive JSON generation for surgical tool detection, yielding the strongest VLM-based performance in our study. The train-validation gap remains.

\begin{figure}[H]
\centering
\includegraphics[width=\textwidth]{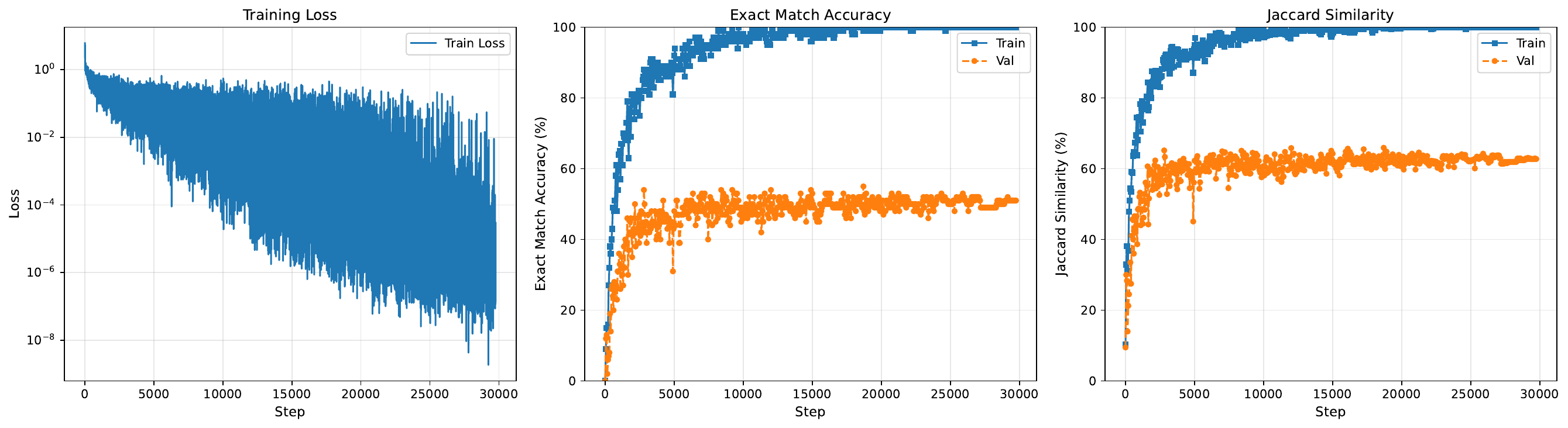}
\caption{Training dynamics for LoRA fine-tuning with classification head on SDSC-EEA ($r=1024$). Left: Training loss (log scale). Center: Exact match accuracy. Right: Jaccard similarity. The classification head achieves the highest validation accuracy among all VLM-based methods (51.08\%), outperforming JSON generation at the same LoRA rank (47.63\%, Figure~\ref{fig:s19_loss}). The persistent train--validation gap reflects limited generalization to held-out procedures. Metrics are computed on fixed random subsets of 100 frames from each set, approximately 100 times throughout training.}
\label{fig:s16_loss}
\end{figure}

\paragraph{Detailed Results} Having established that LoRA fine-tuning with JSON generation improves over zero-shot baselines, we test whether a dedicated classification objective can do better. We replace JSON generation with a linear classification head trained with binary cross-entropy loss (Section~\ref{subsec:lora_methods}). The classification head produces per-tool probability scores, enabling threshold-independent metrics such as ROC-AUC and AUPRC that are not available from discrete JSON outputs. Figure~\ref{fig:s16_loss} shows training dynamics.

After 10 epochs, the fine-tuned model achieves 51.08\% exact match accuracy (95\% CI: 50.39\%--51.81\%) and 61.33\% Jaccard similarity on the validation set ($n=20{,}016$), substantially outperforming both the majority class baseline (13.41\%) and the pre-training baseline (9.83\%). The model also achieves 80.5\% macro-averaged ROC-AUC and 37.6\% macro-averaged AUPRC across the 23 tool classes present in the validation set. Table~\ref{tab:s16_roc_auc} in Appendix~\ref{sec:appendix_s16_metrics} shows per-tool ROC-AUC and AUPRC.

This approach achieves the highest validation accuracy among all VLM-based methods, outperforming JSON generation with the same LoRA rank (Section~\ref{subsec:lora_formatting}, 47.63\% exact match), suggesting that explicit multi-label classification objectives are more effective than autoregressive generation for this task.

\subsection{Scaling LoRA adapter rank does not meaningfully improve out-of-sample accuracy}
\label{subsec:lora_rank_sweep}

\paragraph{Takeaways}
Increasing LoRA rank improves training accuracy but produces only limited gains on held-out procedures. This suggests that the main bottleneck is not insufficient adapter capacity or compute, but failure to generalize under distribution shift.

\begin{figure}[H]
\centering
\includegraphics[width=0.8\textwidth]{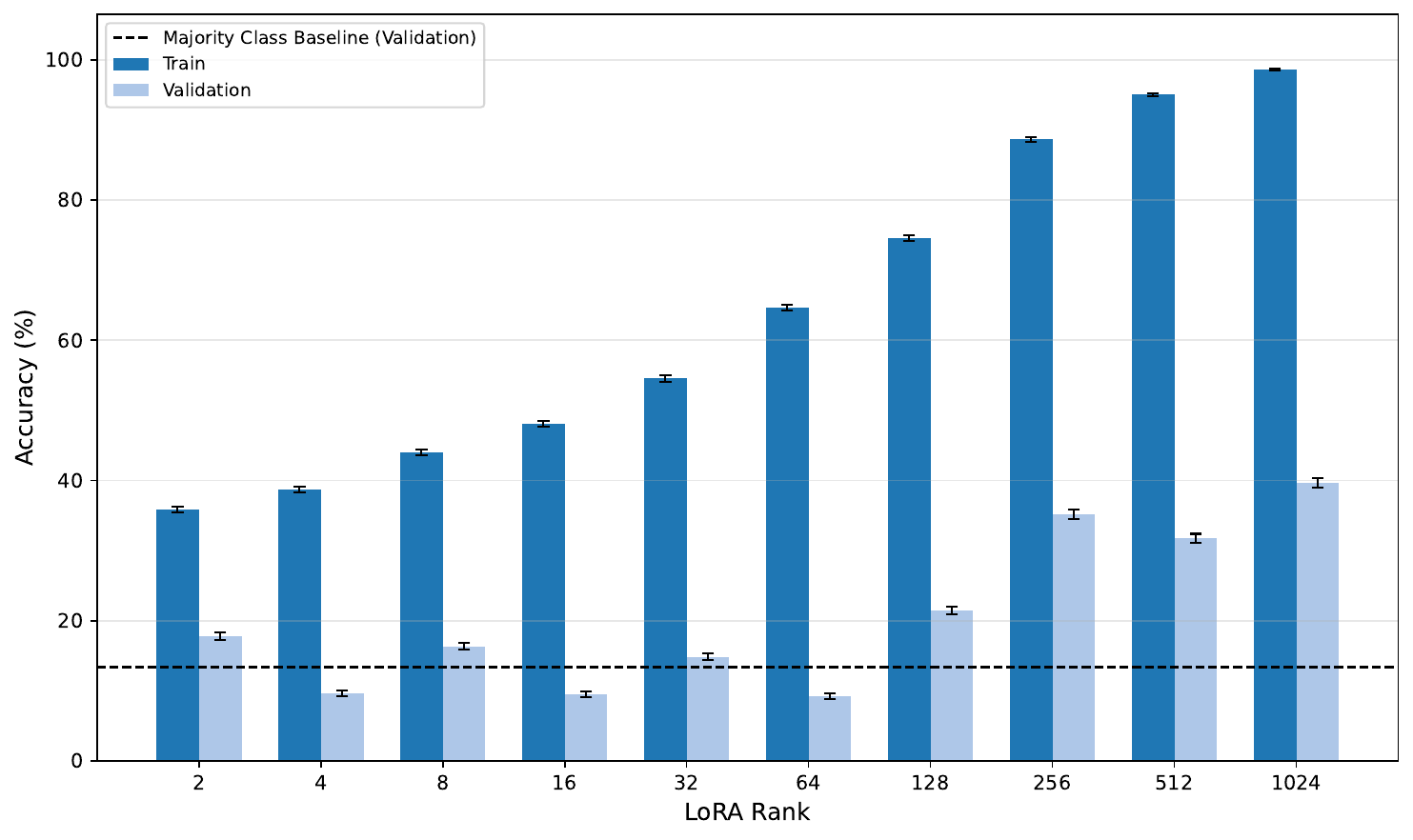}
\caption{Exact match accuracy vs.\ LoRA rank on SDSC-EEA. Gemma 3 27B with LoRA adapters and a linear classification head, trained for 3 epochs at each rank ($r \in \{2, 4, \ldots, 1024\}$). Training accuracy (dark blue) increases monotonically from 35.9\% to 98.6\%, while validation accuracy (light blue) remains below 40\% across all ranks. The widening gap demonstrates that scaling adapter capacity alone cannot overcome the procedure-level distribution shift. Error bars: 95\% bootstrap CIs ($B=1{,}000$).}
\label{fig:accuracy_vs_rank}
\end{figure}

\paragraph{Detailed Results} The experiments in Sections \ref{subsec:lora_formatting}-\ref{subsec:lora_classification_head} use a single, large LoRA rank ($r=1024$). A natural question is whether the validation accuracy gap reflects insufficient model capacity. We sweep LoRA rank from 2 to 1,024, increasing trainable parameters by nearly three orders of magnitude (Table~\ref{tab:rank_sweep}; methodology in Section~\ref{subsec:lora_methods}). Figure~\ref{fig:accuracy_vs_rank} shows accuracy as a function of rank.

\begin{table}[H]
\centering
\caption{LoRA rank sweep on SDSC-EEA: configurations with trainable parameter counts for each rank.}
\label{tab:rank_sweep}
\begin{tabular}{rr}
\toprule
Rank & Params (M) \\
\midrule
2 & 4.7 \\
4 & 9.4 \\
8 & 18.8 \\
16 & 37.5 \\
32 & 75.0 \\
64 & 150.0 \\
128 & 300.0 \\
256 & 600.0 \\
512 & 1200.1 \\
1024 & 2400.2 \\
\bottomrule
\end{tabular}
\end{table}

Training accuracy increases monotonically with rank, from 35.9\% (95\% CI: 35.5\%--36.3\%) at $r=2$ to 98.6\% (95\% CI: 98.5\%--98.7\%) at $r=1024$. Validation accuracy peaks at $r=1024$ with 39.6\% (95\% CI: 39.0\%--40.3\%), though the relationship is non-monotonic.

\subsection{Specialized supervised model}
\label{subsec:yolo_performance}

\paragraph{Takeaways}
For this narrow perceptual task, a small specialized vision model outperforms all VLM-based approaches while using orders of magnitude fewer parameters. This suggests the surgical AI performance is currently limited less by larger generalist models than by the availability of task-specific data.

\paragraph{Detailed Results} The VLM-based approaches above require a 27B-parameter model, with corresponding training time and inference cost. A natural question is whether a small, specialized model can match that performance at a fraction of the cost. YOLOv12-m (Section~\ref{subsec:yolo_methods}), with only 26M parameters---over 1,000$\times$ fewer---achieves 54.73\% exact match accuracy (95\% CI: 54.03\%--55.44\%) and 64.00\% Jaccard similarity (95\% CI: 63.37\%--64.58\%) on the validation set ($n=20{,}016$), with 70.06\% top-1 accuracy (95\% CI: 69.43\%--70.70\%), outperforming the best VLM (Gemma 3 27B with classification head, $r=1024$; 51.08\% exact match, 61.33\% Jaccard). We select YOLO as a natural baseline for this task given its established success in object detection. However, YOLO is trained with bounding box supervision, while VLMs receive only set-level labels. To verify that YOLO's advantage is not solely due to this richer supervisory signal, we train a ResNet-50 (23.6M parameters) using the same set-level labels as VLMs---without any bounding box information. This CNN achieves 39.6\% exact match accuracy (95\% CI: 38.9\%--40.3\%), outperforming all zero-shot VLMs (Appendix~\ref{sec:appendix_resnet}).

Table~\ref{tab:yolo_vs_gemma} compares per-tool metrics between the two models. ROC-AUC is computed from Gemma's sigmoid outputs and YOLO's maximum detection confidence per class.

\begin{table}[H]
\centering
\caption{Per-tool comparison: YOLOv12-m vs.\ Gemma 3 27B (classification head) on the SDSC-EEA validation set ($n=20{,}016$). YOLO achieves higher F1 and recall on all 15 tools, while Gemma achieves higher precision on 4 tools. For ROC-AUC, the two models are complementary: YOLO leads on 7 tools, Gemma on 8. Sorted by ground truth count ($N$). Best per row in bold.}
\label{tab:yolo_vs_gemma}
\footnotesize
\begin{tabular}{l r cc cc cc cc}
\toprule
 & & \multicolumn{2}{c}{Precision} & \multicolumn{2}{c}{Recall} & \multicolumn{2}{c}{F1} & \multicolumn{2}{c}{ROC-AUC} \\
\cmidrule(lr){3-4} \cmidrule(lr){5-6} \cmidrule(lr){7-8} \cmidrule(lr){9-10}
Tool & $N$ & YOLO & Gemma & YOLO & Gemma & YOLO & Gemma & YOLO & Gemma \\
\midrule
Suction & 10685 & \textbf{.732} & .673 & \textbf{.963} & .885 & \textbf{.832} & .764 & \textbf{.875} & .819 \\
Rongeur & 2790 & .948 & \textbf{.960} & \textbf{.716} & .222 & \textbf{.816} & .361 & .866 & \textbf{.920} \\
Cotton Patty & 2143 & \textbf{.877} & .576 & \textbf{.819} & .706 & \textbf{.847} & .635 & \textbf{.981} & .929 \\
Drill & 2116 & .943 & \textbf{.945} & \textbf{.959} & .790 & \textbf{.951} & .861 & \textbf{.984} & .983 \\
Rhoton Dissector & 1462 & \textbf{.554} & .540 & \textbf{.882} & .590 & \textbf{.680} & .564 & \textbf{.945} & .920 \\
Surgical Knife & 1422 & \textbf{.947} & .920 & \textbf{.904} & .049 & \textbf{.925} & .092 & \textbf{.953} & .939 \\
Suction Coagulator & 1188 & .982 & \textbf{1.00} & \textbf{.995} & .622 & \textbf{.988} & .767 & .998 & \textbf{1.00} \\
Backbiting rongeur & 1041 & .852 & .741 & \textbf{.243} & .019 & \textbf{.378} & .038 & .641 & \textbf{.918} \\
Scissor & 996 & .458 & \textbf{.622} & \textbf{.673} & .136 & \textbf{.545} & .223 & \textbf{.840} & .766 \\
Surgicel & 739 & \textbf{1.00} & .971 & \textbf{.635} & .628 & \textbf{.776} & .763 & .825 & \textbf{.908} \\
Curette & 708 & .842 & \textbf{.949} & \textbf{.468} & .239 & \textbf{.601} & .382 & .744 & \textbf{.940} \\
Grasper & 509 & \textbf{.289} & .032 & \textbf{.686} & .020 & \textbf{.406} & .024 & \textbf{.840} & .708 \\
Bipolar Forceps & 263 & \textbf{.752} & .000 & \textbf{.346} & .000 & \textbf{.474} & .000 & .675 & \textbf{.723} \\
Straight Forceps & 173 & .454 & \textbf{.642} & \textbf{.775} & .249 & \textbf{.573} & .358 & .908 & \textbf{.943} \\
Irrigation & 112 & .809 & \textbf{1.00} & \textbf{.339} & .018 & \textbf{.478} & .035 & .701 & \textbf{.776} \\
\bottomrule
\end{tabular}
\end{table}

\subsection{Robustness}\label{subsec:robustness}

\subsubsection{Performance on CholecT50}
\label{subsec:cholect50_results}

\paragraph{Takeaways}
The results on SDSC-EEA reproduce on CholecT50: the broad pattern that zero-shot performance is poor, that fine-tuning is necessary, and that smaller models outperform VLMs at a fraction of the size. Additionally, frontier proprietary models from the GPT, Gemini, and Claude families underperform a fine-tuned open-weight LLM and a specialised computer vision model.

\paragraph{Detailed Results} To assess whether our findings generalize beyond neurosurgery, we evaluate on CholecT50, an independent laparoscopic cholecystectomy dataset with 6 instrument classes (Section~\ref{subsec:cholect50_methods}). Zero-shot Gemma 3 27B achieves 6.87\% exact match accuracy (95\% CI: 6.55\%--7.22\%), which is below the majority class baseline (34.76\%).

Fine-tuning Gemma 3 27B with LoRA ($r=128$) and a classification head reaches 83.02\% exact match accuracy (95\% CI: 82.52\%--83.56\%) and 88.79\% Jaccard similarity (95\% CI: 88.43\%--89.18\%). YOLOv12-m achieves 81.37\% exact match accuracy (95\% CI: 80.87\%--81.92\%) and 88.00\% Jaccard similarity (95\% CI: 87.62\%--88.34\%), with 93.80\% top-1 accuracy (95\% CI: 93.45\%--94.12\%). Table~\ref{tab:cholect50_yolo_vs_gemma} compares per-tool metrics between the two fine-tuned models.

Per-tool metrics for the zero-shot setting are in Appendix~\ref{sec:appendix_cholect50}: grasper achieves the highest F1 (0.627), while bipolar has 12,096 false positives vs.\ 838 true positives.

Additionally, since CholecT50 is a public dataset, we can evaluate the performance of eleven frontier VLMs from the GPT \citep{openai2026gpt54,openai2026gpt56sol}, Gemini 3 \citep{gemini2026gemini31pro}, Claude \citep{anthropic2026opus46,anthropic2026fable5}, Kimi \citep{moonshot2026kimik3}, and Qwen \citep{qwen2026qwen38max} families\footnote{We cannot evaluate SDSC-EEA on proprietary frontier models because we cannot send private SDSC-EEA data to third-party APIs. All evaluations on SDSC-EEA were conducted locally on open-source VLMs.} using the same prompt template and validation frames. Six of these models---Claude Fable 5, Claude Fable 5.1, Gemini 3.8 Flash, GPT-5.6 Sol, Kimi K3, and Qwen3.8 Max---were evaluated on a random subsample of 1{,}000 validation frames per dataset rather than the full validation set; they are marked with an asterisk ($^*$) in all tables. The CholecT50 column of the consolidated cross-dataset summary in Table~\ref{tab:cross_dataset_em} reports exact-match accuracy for every model evaluated on this dataset; the corresponding 95\% bootstrap confidence intervals are in Appendix~\ref{sec:appendix_cross_dataset_ci} (Table~\ref{tab:cholect50_all_models_ci}).

Four frontier models---Gemini 3.8 Flash$^*$ (73.40\%), Qwen3.8 Max$^*$ (70.90\%), Gemini 3 Flash Preview (69.15\%), and Gemini 3.1 Pro Preview (66.21\%)---achieve the highest zero-shot exact match accuracy, far surpassing the open-weight Gemma~3 27B (6.87\%) and approaching the fine-tuned models. Second, performance varies dramatically across model families: Qwen3.8 Max$^*$ and all three Google models exceed 66\%, Claude Fable 5.1$^*$ and Kimi K3$^*$ reach 61.70\%, GPT-5.6 Sol$^*$, Claude Fable 5$^*$, and Claude Opus fall in the 52--56\% range, while GPT-5.4 and Claude Sonnet fall below the majority class baseline. Third, the fine-tuned models (83\% and 81\%) still outperform even the best zero-shot API model (Gemini 3.8 Flash$^*$, 73.40\%) by 8--10 percentage points, confirming that task-specific training remains valuable even as frontier models improve.

\begin{table}[H]
\centering
\caption{Per-tool comparison: YOLOv12-m vs.\ Gemma 3 27B (classification head) on CholecT50 validation set ($n=19{,}923$). Unlike SDSC-EEA (Table~\ref{tab:yolo_vs_gemma}), results are mixed: Gemma achieves higher F1 on 4 of 6 tools; YOLO leads on irrigator and bipolar. Gemma has higher precision on all 6 tools; YOLO has higher recall on 5 of 6. Sorted by Gemma F1. Best per row in bold.}
\label{tab:cholect50_yolo_vs_gemma}
\footnotesize
\begin{tabular}{l cc cc cc cc}
\toprule
 & \multicolumn{2}{c}{Precision} & \multicolumn{2}{c}{Recall} & \multicolumn{2}{c}{F1} & \multicolumn{2}{c}{ROC-AUC} \\
\cmidrule(lr){2-3} \cmidrule(lr){4-5} \cmidrule(lr){6-7} \cmidrule(lr){8-9}
Tool & YOLO & Gemma & YOLO & Gemma & YOLO & Gemma & YOLO & Gemma \\
\midrule
hook & .953 & \textbf{.972} & \textbf{.986} & .977 & .969 & \textbf{.974} & \textbf{.992} & .989 \\
clipper & .902 & \textbf{.936} & \textbf{.910} & .895 & .906 & \textbf{.915} & .967 & \textbf{.989} \\
grasper & .860 & \textbf{.899} & \textbf{.953} & .927 & .904 & \textbf{.913} & \textbf{.958} & .941 \\
irrigator & .910 & \textbf{.929} & \textbf{.819} & .703 & \textbf{.862} & .800 & .955 & \textbf{.969} \\
bipolar & \textbf{.920} & .944 & \textbf{.776} & .743 & \textbf{.842} & .831 & .920 & \textbf{.959} \\
scissors & .884 & \textbf{.888} & .578 & \textbf{.599} & .699 & \textbf{.715} & .825 & \textbf{.947} \\
\bottomrule
\end{tabular}
\end{table}

\begin{figure}[H]
\centering
\includegraphics[width=0.8\textwidth]{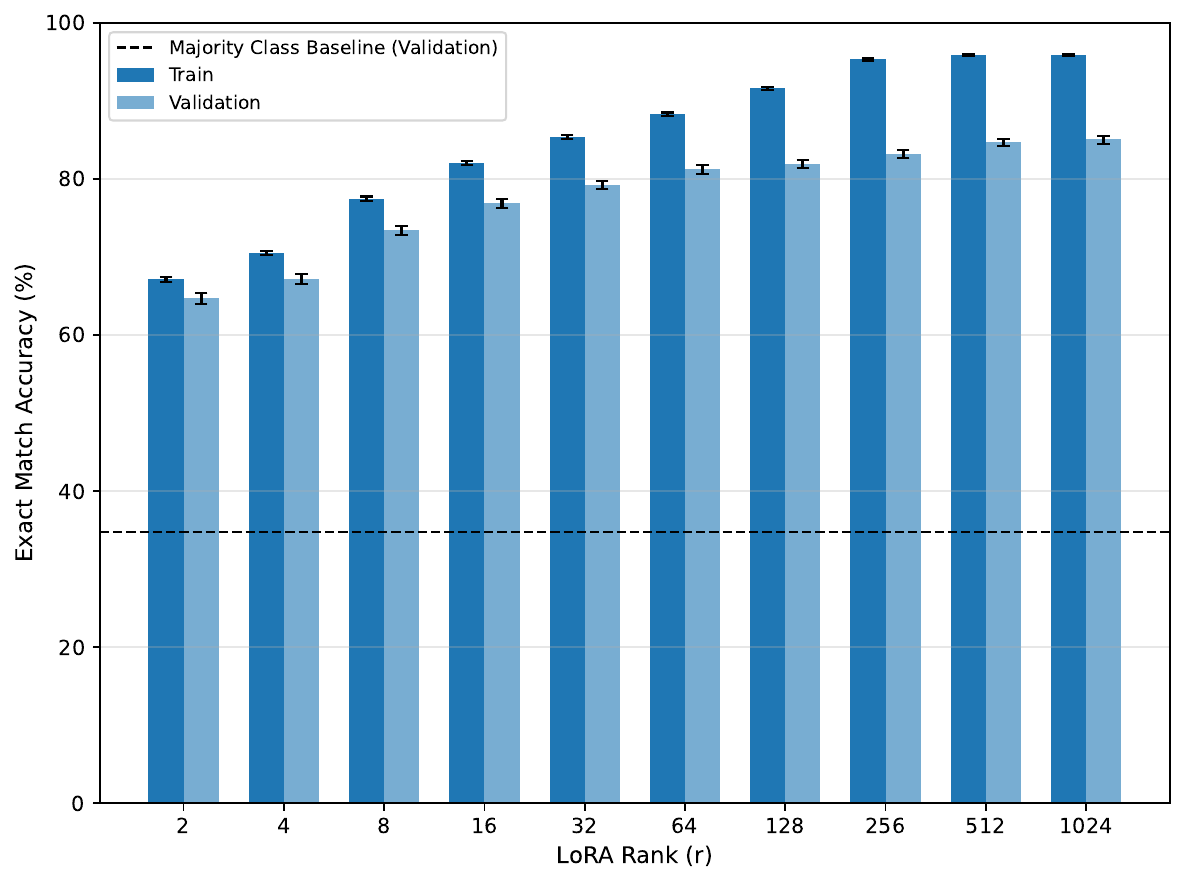}
\caption{Exact match accuracy vs.\ LoRA rank on CholecT50. Unlike SDSC-EEA (Figure~\ref{fig:accuracy_vs_rank}), validation accuracy increases monotonically across all ranks, reaching 85.1\% at $r=1024$, with a much smaller train--validation gap. The lower tool diversity (6 vs.\ 31 classes) and more uniform video-level distribution make CholecT50 more amenable to LoRA fine-tuning. Dashed line: majority class baseline (34.76\%). Error bars: 95\% bootstrap CIs ($B=1{,}000$).}
\label{fig:cholect50_accuracy_vs_rank}
\end{figure}

We additionally sweep LoRA rank from 2 to 1,024 on CholecT50 using the same protocol as Section~\ref{subsec:lora_rank_sweep}. Figure~\ref{fig:cholect50_accuracy_vs_rank} shows accuracy as a function of rank. Training accuracy increases from 67.1\% (95\% CI: 66.8\%--67.5\%) at $r=2$ to 95.9\% (95\% CI: 95.8\%--96.1\%) at $r=1024$. Validation accuracy increases from 64.7\% (95\% CI: 64.0\%--65.4\%) at $r=2$ to 85.1\% (95\% CI: 84.5\%--85.6\%) at $r=1024$. Unlike SDSC-EEA, where validation accuracy exhibits a non-monotonic relationship with rank and remains below 40\% even at $r=1024$, CholecT50 validation accuracy increases monotonically across all tested ranks and reaches 85.1\% at $r=1024$.

\subsubsection{Performance on PitVis-2023}
\label{subsec:pitvis_results}

\paragraph{Takeaways}
The patterns observed on SDSC-EEA and CholecT50 reproduce on a third independent dataset---PitVis-2023, an endoscopic pituitary neurosurgery benchmark with 18 instrument classes. Zero-shot open-weight VLMs underperform a trivial baseline; closed-weight frontier VLMs improve over Gemma but remain far below fine-tuned models; LoRA fine-tuning of Gemma 3 27B and a specialized YOLOv12-m both substantially outperform every zero-shot approach, including all eleven frontier models.

\paragraph{Detailed Results} To further test whether our findings generalize, we evaluate on PitVis-2023 \citep{pitvis}, an endoscopic transsphenoidal pituitary surgery dataset with 18 instrument classes (Section~\ref{subsec:pitvis_methods}). PitVis is closer in surgical domain to SDSC-EEA than CholecT50 (both are endoscopic neurosurgical procedures), but is a fully independent dataset collected at a different institution with a different surgical team and label taxonomy. Following the same protocol as Section~\ref{subsec:cholect50_results}, we evaluate zero-shot Gemma 3 27B, the same eleven frontier VLMs, fine-tuned Gemma 3 27B with LoRA and a classification head, and YOLOv12-m on a video-level 19/5 train/validation split (84,666 train frames, 30,896 validation frames). The majority class baseline---predicting the most common tool set (the empty set, corresponding to no instruments visible) for every frame---achieves 39.63\% exact match accuracy on the validation set.

Zero-shot Gemma 3 27B achieves 28.46\% exact match accuracy (95\% CI: 27.97\%--28.94\%), again below the majority class baseline. Fine-tuning Gemma 3 27B with LoRA ($r=128$) and a classification head reaches 84.77\% exact match accuracy and 87.14\% Jaccard similarity on the validation set, with a macro-averaged ROC-AUC of 0.966 and macro AUPRC of 0.691 across the 18 tool classes. YOLOv12-m achieves 82.78\% exact match accuracy and 86.50\% Jaccard similarity, with 89.26\% top-1 accuracy and a macro-averaged ROC-AUC of 0.853. The PitVis-2023 column of Table~\ref{tab:cross_dataset_em} reports exact-match accuracy for every model evaluated on this dataset; the corresponding 95\% bootstrap confidence intervals are in Appendix~\ref{sec:appendix_cross_dataset_ci} (Table~\ref{tab:pitvis_all_models_ci}).

The ranking of model families partially mirrors the CholecT50 results (Table~\ref{tab:cross_dataset_em}): Claude Fable 5.1$^*$ (58.70\%), Gemini 3.8 Flash$^*$ (58.40\%), Gemini 3.1 Pro (57.65\%), and Claude Fable 5$^*$ (57.20\%) are the strongest zero-shot models, followed by Gemini 3 Flash, Qwen3.8 Max$^*$, the two Claude 4.6 models, GPT-5.4, and Kimi K3$^*$, all clustered between 48\% and 53\%. GPT-5.6 Sol$^*$ (39.40\%) is the weakest frontier model and sits at the majority class baseline (39.63\%). Every other frontier model exceeds the open-weight Gemma 3 27B by a wide margin (more than 20 percentage points), but every frontier model also remains at least 26 percentage points below the fine-tuned Gemma 3 27B. The fine-tuned open-weight model and YOLOv12-m again outperform all zero-shot approaches, including the frontier VLMs.

\begin{table}[H]
\centering
\caption{Per-tool comparison: YOLOv12-m vs.\ Gemma 3 27B (LoRA + classification head) on the PitVis-2023 validation set ($n=30{,}896$). Sorted by ground truth count ($N$) in descending order. Best per row in bold.}
\label{tab:pitvis_yolo_vs_gemma}
\footnotesize
\begin{tabular}{l r cc cc cc cc}
\toprule
 & & \multicolumn{2}{c}{Precision} & \multicolumn{2}{c}{Recall} & \multicolumn{2}{c}{F1} & \multicolumn{2}{c}{ROC-AUC} \\
\cmidrule(lr){3-4} \cmidrule(lr){5-6} \cmidrule(lr){7-8} \cmidrule(lr){9-10}
Tool & $N$ & YOLO & Gemma & YOLO & Gemma & YOLO & Gemma & YOLO & Gemma \\
\midrule
suction & 11971 & .838 & \textbf{.886} & \textbf{.972} & .958 & .900 & \textbf{.921} & \textbf{.979} & .974 \\
ring\_curette & 4314 & \textbf{.975} & .966 & .780 & \textbf{.851} & .866 & \textbf{.905} & .938 & \textbf{.988} \\
kerrisons & 3567 & .877 & \textbf{.891} & \textbf{.770} & .748 & .820 & \textbf{.814} & \textbf{.943} & .939 \\
pituitary\_rongeurs & 909 & .665 & \textbf{.845} & \textbf{.503} & .414 & .573 & \textbf{.555} & .816 & \textbf{.921} \\
spatula\_dissector & 412 & .471 & \textbf{.719} & \textbf{.396} & .279 & .430 & \textbf{.402} & .706 & \textbf{.965} \\
nasal\_cutting\_forceps & 374 & .567 & \textbf{.709} & \textbf{.610} & .436 & .588 & \textbf{.540} & .806 & \textbf{.924} \\
stealth\_pointer & 353 & .833 & \textbf{.851} & \textbf{.507} & .518 & .630 & \textbf{.644} & .898 & \textbf{.963} \\
irrigation\_syringe & 286 & .483 & \textbf{.801} & .598 & .591 & .534 & \textbf{.680} & .892 & \textbf{.964} \\
micro\_doppler & 251 & .967 & \textbf{.969} & \textbf{.813} & .745 & .883 & \textbf{.842} & .920 & \textbf{.991} \\
cup\_forceps & 241 & .181 & \textbf{.500} & \textbf{.710} & .257 & .288 & \textbf{.340} & .908 & \textbf{.941} \\
freer\_elevator & 226 & .661 & \textbf{.639} & \textbf{.681} & .549 & \textbf{.671} & .591 & .852 & \textbf{.968} \\
haemostatic\_foam & 179 & .903 & .896 & .726 & \textbf{.721} & \textbf{.805} & .799 & .880 & \textbf{.963} \\
retractable\_knife & 136 & .695 & \textbf{.800} & \textbf{.485} & .235 & .571 & \textbf{.364} & .794 & \textbf{.917} \\
cottle & 130 & .574 & \textbf{.687} & \textbf{.862} & .354 & \textbf{.689} & .467 & .945 & \textbf{.994} \\
dural\_scissors & 129 & \textbf{.594} & .452 & \textbf{.295} & .109 & \textbf{.394} & .175 & .647 & \textbf{.984} \\
surgical\_drill & 80 & .896 & \textbf{1.00} & \textbf{.538} & .763 & .672 & \textbf{.865} & .775 & \textbf{.991} \\
tissue\_glue & 63 & .732 & \textbf{.839} & \textbf{.952} & .825 & \textbf{.828} & .832 & .984 & \textbf{.998} \\
bipolar\_forceps & 49 & .842 & \textbf{1.00} & \textbf{.327} & .082 & \textbf{.471} & .151 & .663 & \textbf{.999} \\
\bottomrule
\end{tabular}
\end{table}

Per-tool comparison (Table~\ref{tab:pitvis_yolo_vs_gemma}) reveals a similar pattern to CholecT50: Gemma achieves higher precision on most tools while YOLO achieves higher recall, but neither model dominates uniformly across F1. Gemma's classification head produces stronger ROC-AUC on 17 of 18 tools (the only exception being suction).

\begin{figure}[H]
\centering
\includegraphics[width=0.8\textwidth]{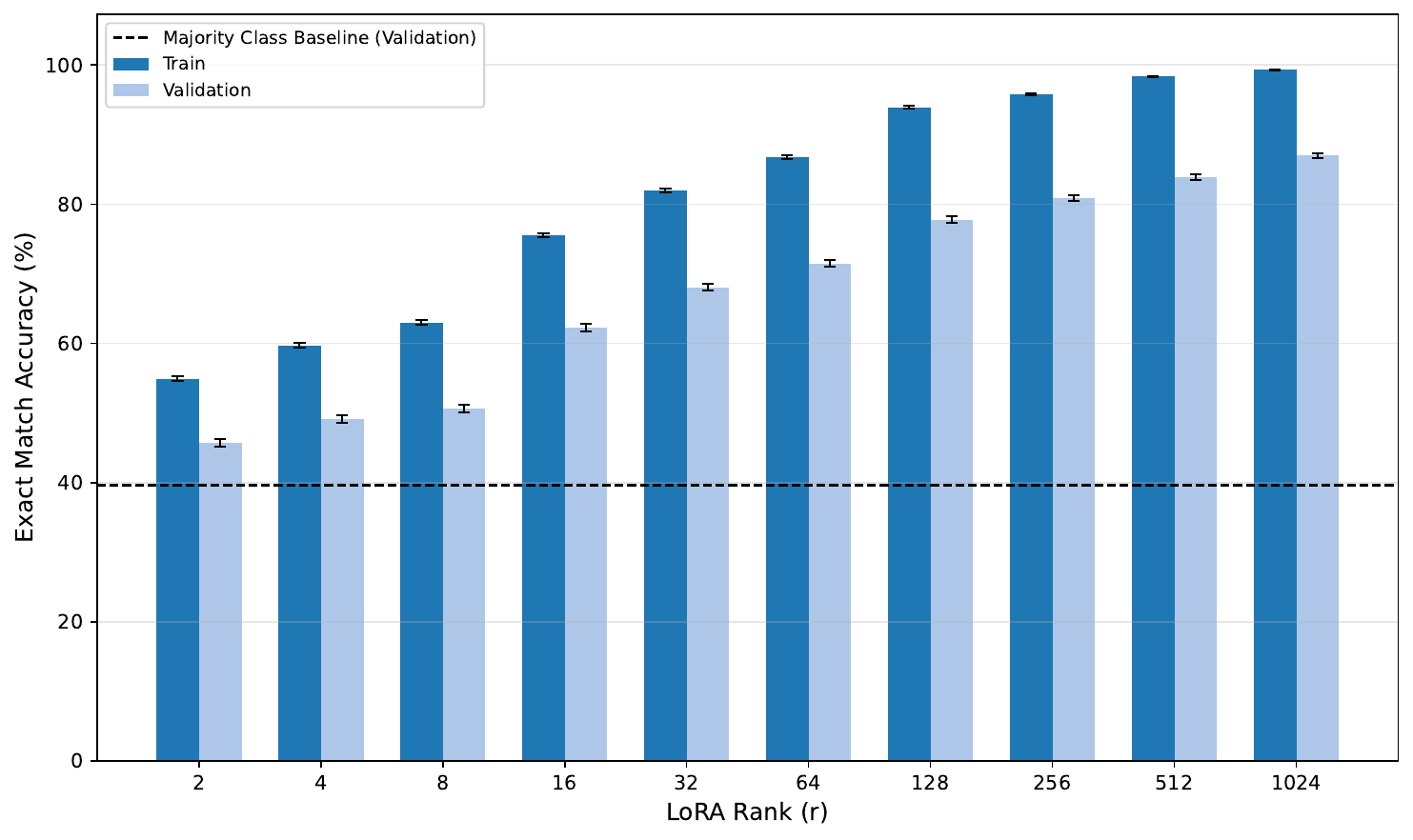}
\caption{Exact match accuracy vs.\ LoRA rank on PitVis-2023. As on CholecT50 (Figure~\ref{fig:cholect50_accuracy_vs_rank}) and unlike SDSC-EEA (Figure~\ref{fig:accuracy_vs_rank}), validation accuracy increases monotonically across all tested ranks, reaching 86.97\% at $r=1024$. Dashed line: majority class baseline (39.63\%). Error bars: 95\% bootstrap CIs ($B=1{,}000$).}
\label{fig:pitvis_accuracy_vs_rank}
\end{figure}

We additionally sweep LoRA rank from 2 to 1,024 on PitVis-2023 using the same protocol as Section~\ref{subsec:lora_rank_sweep}. Figure~\ref{fig:pitvis_accuracy_vs_rank} shows accuracy as a function of rank. Training accuracy increases monotonically from 54.9\% (95\% CI: 54.58\%--55.27\%) at $r=2$ to 99.3\% (95\% CI: 99.27\%--99.39\%) at $r=1024$. Validation accuracy increases monotonically from 45.7\% (95\% CI: 45.14\%--46.23\%) at $r=2$ to 86.97\% (95\% CI: 86.63\%--87.34\%) at $r=1024$. As on CholecT50, validation accuracy on PitVis-2023 increases smoothly with rank---in contrast to SDSC-EEA, where validation accuracy plateaus below 40\% regardless of rank.

\subsubsection{Performance on SurgVU}
\label{subsec:surgvu_results}

\paragraph{Takeaways}
The patterns from the three previous sections also hold on a fourth independent dataset and a markedly different surgical domain---SurgVU, a public benchmark of robotic-assisted surgery training sessions on porcine tissue with 17 released instrument classes. Zero-shot Gemma~3 27B again fails to surpass the trivial majority class baseline; the eleven frontier VLMs do somewhat better but, unlike on CholecT50 and PitVis-2023, only five of the eleven (Claude Sonnet 4.6, Gemini 3.1 Pro, Claude Fable 5.1$^*$, Claude Fable 5$^*$, and Gemini 3.8 Flash$^*$) clearly exceed the baseline, and even the best of them remains roughly 28~percentage points below the fine-tuned open-weight model and YOLOv12-m. LoRA fine-tuning of Gemma~3 27B and a specialized YOLOv12-m both substantially outperform every zero-shot approach. However, in contrast to CholecT50 and PitVis-2023 and consistent with our SDSC-EEA findings, validation accuracy plateaus around 50\% across LoRA ranks even as training accuracy climbs above 80\%, indicating that the procedure-level distribution shift on SurgVU is closer in difficulty to SDSC-EEA than to the other two public datasets.

\paragraph{Detailed Results} To further test whether our findings generalize beyond endoscopic neurosurgery (SDSC-EEA, PitVis-2023) and laparoscopic cholecystectomy (CholecT50), we evaluate on SurgVU~\citep{surgvu}, a public dataset of robotic-assisted surgery training sessions on porcine tissue (Section~\ref{subsec:surgvu_methods}). The dataset is collected on a different hardware platform (da~Vinci robot vs.\ endoscopic camera) and in a different setting (training exercises on porcine tissue vs.\ live patient procedures) than the other three benchmarks. Following the same protocol as Sections~\ref{subsec:cholect50_results}--\ref{subsec:pitvis_results}, we evaluate zero-shot Gemma~3 27B, the same eleven frontier VLMs, fine-tuned Gemma~3 27B with LoRA and a classification head, and YOLOv12-m on a session-level 124/31 train/validation split of the 155 SurgVU training sessions (81,751 train frames, 18,919 validation frames). The majority class baseline---predicting the most common tool set (the empty set, corresponding to no instruments visible) for every frame---achieves 16.94\% exact match accuracy on the validation set.

Zero-shot Gemma~3 27B achieves 2.90\% exact match accuracy (95\% CI: 2.66\%--3.16\%), well below the majority class baseline. Fine-tuning Gemma~3 27B with LoRA ($r=128$) and a classification head reaches 50.61\% exact match accuracy and 67.52\% Jaccard similarity on the validation set, with a macro-averaged ROC-AUC of 0.740 and macro AUPRC of 0.502 across the 14 tool classes present in the validation set. YOLOv12-m achieves 51.75\% exact match accuracy and 68.72\% Jaccard similarity, with 80.17\% top-1 accuracy and a macro-averaged ROC-AUC of 0.731. The SurgVU column of Table~\ref{tab:cross_dataset_em} reports exact-match accuracy for every model evaluated on this dataset; the corresponding 95\% bootstrap confidence intervals are in Appendix~\ref{sec:appendix_cross_dataset_ci} (Table~\ref{tab:surgvu_all_models_ci}).

The eleven frontier VLMs perform substantially worse on SurgVU than on either CholecT50 or PitVis-2023 (Table~\ref{tab:cross_dataset_em}), and the family-level ranking changes markedly across datasets. Claude Sonnet 4.6 (23.05\%, 95\% CI 22.46--23.65) and Gemini 3.1 Pro Preview (22.46\%, 95\% CI 21.86--23.02) are the only two frontier models that clearly exceed the 16.94\% baseline by more than 5~percentage points. Claude Fable 5.1$^*$ (21.00\%, 95\% CI 18.50--23.70), Claude Fable 5$^*$ (20.09\%, 95\% CI 17.75--22.34), and Gemini 3.8 Flash$^*$ (19.80\%, 95\% CI 17.40--22.40) also exceed the baseline---none of their 95\% CIs overlap 16.94\%---though by smaller margins of roughly 3--4~percentage points. GPT-5.4 (17.64\%, 95\% CI 17.10--18.24) edges marginally above the baseline---its 95\% CI does not overlap 16.94\%, but the margin is only 0.7~percentage points. Claude Opus 4.6 (17.22\%, 95\% CI 16.73--17.77), GPT-5.6 Sol$^*$ (17.40\%, 95\% CI 15.20--19.90), Qwen3.8 Max$^*$ (17.20\%, 95\% CI 14.90--19.70), and Kimi K3$^*$ (16.39\%, 95\% CI 13.68--19.25) are statistically indistinguishable from the baseline (their CIs overlap 16.94\%). Gemini 3 Flash Preview (11.69\%, 95\% CI 11.23--12.16) falls more than 5~percentage points below the baseline. Compared with PitVis-2023, every frontier model loses 22--41~percentage points of exact match accuracy on SurgVU, with Gemini 3 Flash dropping the most (from 52.62\% to 11.69\%). Despite this large gap, every frontier model still substantially outperforms zero-shot Gemma~3 27B (2.90\%), so the broad pattern from CholecT50 and PitVis-2023 holds: frontier VLMs improve over zero-shot Gemma but remain far below fine-tuned open-weight models and the small specialized object detector. Even the strongest frontier model on SurgVU (Claude Sonnet 4.6) trails the fine-tuned Gemma~3 27B by 27.6~percentage points and YOLOv12-m by 28.7~percentage points.

Per-tool comparison between YOLOv12-m and Gemma~3 27B + LoRA classification head is shown in Table~\ref{tab:surgvu_yolo_vs_gemma}. As on CholecT50 and PitVis-2023, Gemma achieves higher precision on most tools (8 of 10 tools with non-zero F1), while YOLO achieves higher recall (8 of 10 tools). YOLO leads on F1 on 6 of 10 tools (notably bipolar forceps, cadiere forceps, grasping retractor) while Gemma leads on F1 for 4 tools (clip applier, force bipolar, permanent cautery hook/spatula, vessel sealer). Both models entirely fail on the rarest tools (suction irrigator, synchroseal, tip-up fenestrated grasper)---each with fewer than 200 ground truth instances in the validation set---reflecting the heavy long-tail in the SurgVU instrument distribution.

\begin{table}[H]
\centering
\caption{Per-tool comparison: YOLOv12-m vs.\ Gemma 3 27B (LoRA + classification head) on the SurgVU validation set ($n=18{,}919$). Tools with zero ground truth instances in the validation set (bipolar dissector, potts scissors, tenaculum forceps) are omitted. Sorted by ground truth count ($N$) in descending order. Best per row in bold; rows where neither model produces a non-zero score are not bolded.}
\label{tab:surgvu_yolo_vs_gemma}
\footnotesize
\begin{tabular}{l r cc cc cc cc}
\toprule
 & & \multicolumn{2}{c}{Precision} & \multicolumn{2}{c}{Recall} & \multicolumn{2}{c}{F1} & \multicolumn{2}{c}{ROC-AUC} \\
\cmidrule(lr){3-4} \cmidrule(lr){5-6} \cmidrule(lr){7-8} \cmidrule(lr){9-10}
Tool & $N$ & YOLO & Gemma & YOLO & Gemma & YOLO & Gemma & YOLO & Gemma \\
\midrule
needle driver & 7946 & .845 & .834 & \textbf{.746} & .735 & \textbf{.792} & .781 & \textbf{.867} & .859 \\
cadiere forceps & 7928 & .797 & \textbf{.808} & \textbf{.933} & .863 & \textbf{.860} & .835 & \textbf{.937} & .912 \\
bipolar forceps & 6980 & .817 & \textbf{.866} & \textbf{.889} & .864 & \textbf{.851} & .865 & \textbf{.934} & .928 \\
monop. curved scissors & 6906 & .880 & \textbf{.914} & \textbf{.739} & .716 & .803 & \textbf{.803} & .873 & \textbf{.881} \\
prograsp forceps & 3341 & \textbf{.690} & .695 & \textbf{.540} & .503 & \textbf{.606} & .584 & .763 & \textbf{.738} \\
grasping retractor & 3096 & .750 & .718 & \textbf{.719} & .671 & \textbf{.734} & .694 & \textbf{.840} & .897 \\
force bipolar & 2091 & .743 & .736 & \textbf{.190} & .153 & \textbf{.302} & .253 & \textbf{.582} & .735 \\
per. caut. hook/spatula & 1275 & \textbf{.535} & .764 & .230 & \textbf{.253} & .321 & \textbf{.380} & .611 & \textbf{.665} \\
vessel sealer & 984 & \textbf{.522} & .710 & \textbf{.714} & .716 & .603 & \textbf{.713} & .875 & \textbf{.932} \\
clip applier & 967 & \textbf{.476} & .669 & .113 & \textbf{.092} & .182 & \textbf{.162} & .588 & \textbf{.609} \\
stapler & 444 & \textbf{.785} & .892 & .723 & \textbf{.651} & .753 & \textbf{.753} & .888 & \textbf{.955} \\
tip-up fenest. grasper & 352 & .000 & .000 & .000 & .000 & .000 & .000 & .500 & \textbf{.363} \\
suction irrigator & 62 & .000 & .000 & .000 & .000 & .000 & .000 & \textbf{.500} & .605 \\
synchroseal & 10 & .000 & .000 & .000 & .000 & .000 & .000 & \textbf{.477} & .282 \\
\bottomrule
\end{tabular}
\end{table}

\begin{figure}[H]
\centering
\includegraphics[width=0.8\textwidth]{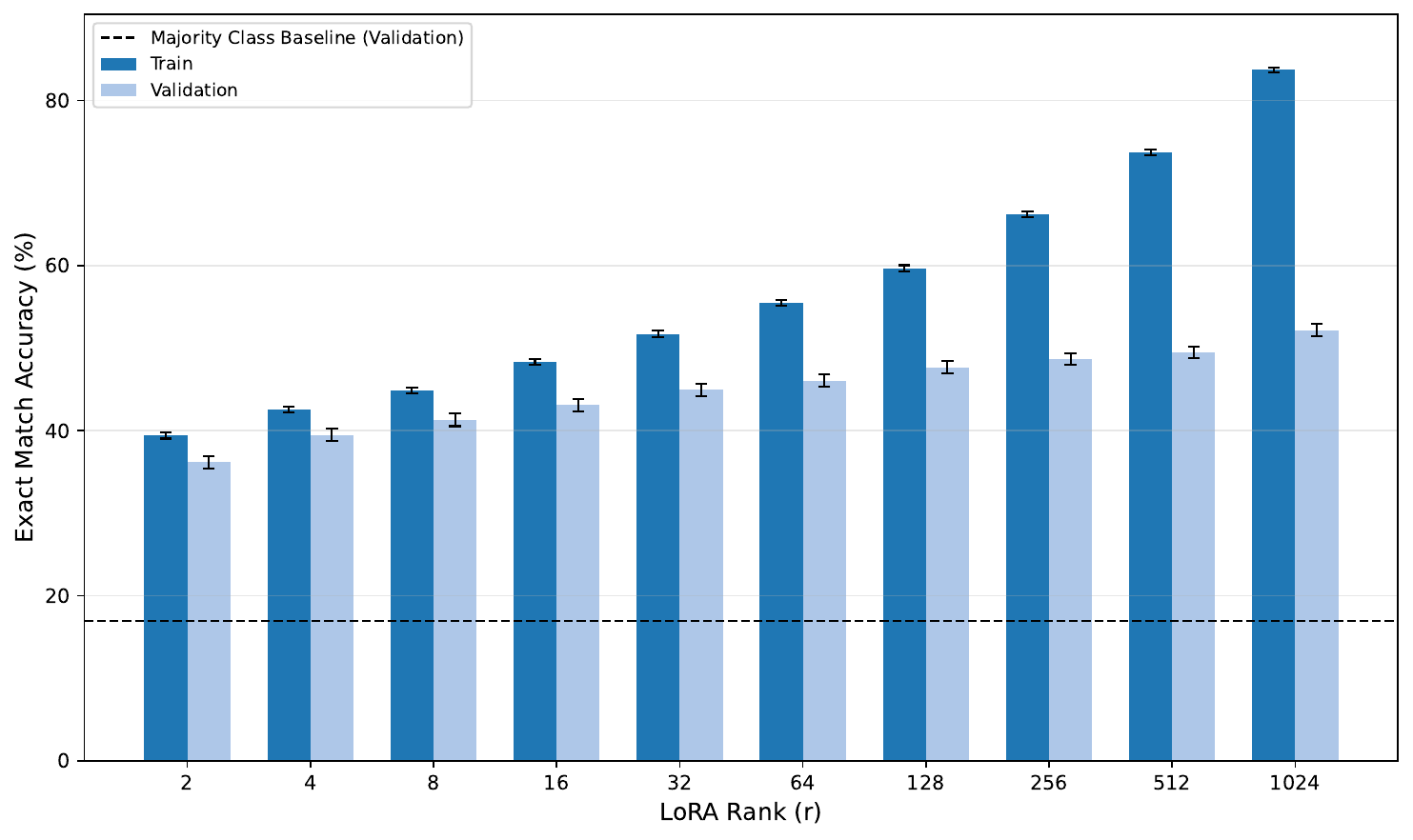}
\caption{Exact match accuracy vs.\ LoRA rank on SurgVU. As on SDSC-EEA (Figure~\ref{fig:accuracy_vs_rank}) and unlike CholecT50 (Figure~\ref{fig:cholect50_accuracy_vs_rank}) and PitVis-2023 (Figure~\ref{fig:pitvis_accuracy_vs_rank}), the train--validation gap widens substantially as rank increases. Validation accuracy reaches 52.19\% at $r=1024$, while training accuracy reaches 83.73\%. Dashed line: majority class baseline (16.94\%). Error bars: 95\% bootstrap CIs ($B=1{,}000$).}
\label{fig:surgvu_accuracy_vs_rank}
\end{figure}

We additionally sweep LoRA rank from 2 to 1,024 on SurgVU using the same protocol as Section~\ref{subsec:lora_rank_sweep}. Figure~\ref{fig:surgvu_accuracy_vs_rank} shows accuracy as a function of rank. Training accuracy increases monotonically from 39.4\% (95\% CI: 39.04\%--39.79\%) at $r=2$ to 83.7\% (95\% CI: 83.44\%--83.99\%) at $r=1024$. Validation accuracy also increases with rank but at a much slower pace---from 36.2\% at $r=2$ to 52.2\% at $r=1024$---and the gap between training and validation accuracy widens substantially as rank grows. This pattern mirrors the SDSC-EEA rank sweep more closely than the smooth scaling observed on CholecT50 and PitVis-2023, suggesting that the session-level distribution shift on SurgVU---where individual training sessions vary substantially in instrument repertoire---is a more meaningful generalization barrier than raw model capacity.

\subsection{Replication on Bounding-Box-Free Annotation Tasks}
\label{subsec:newtasks}

\paragraph{Takeaways}
The specialist-over-generalist hierarchy is not an artifact of object detection. On four annotation tasks that involve no bounding boxes---workflow recognition, gesture recognition, action recognition, and anatomy presence---small supervised specialists or a task-adapted open-weight VLM lead every dataset, a frozen surgical foundation model with a linear probe follows, and all ten frontier VLMs trail the top two models on every dataset, with only one clearing the majority class baseline on gesture recognition. Task-specific supervision is absorbed by both small specialists and an adapted 27B VLM; it is not absorbed by frontier scale alone.

\begin{figure}[H]
\centering
\includegraphics[width=\textwidth]{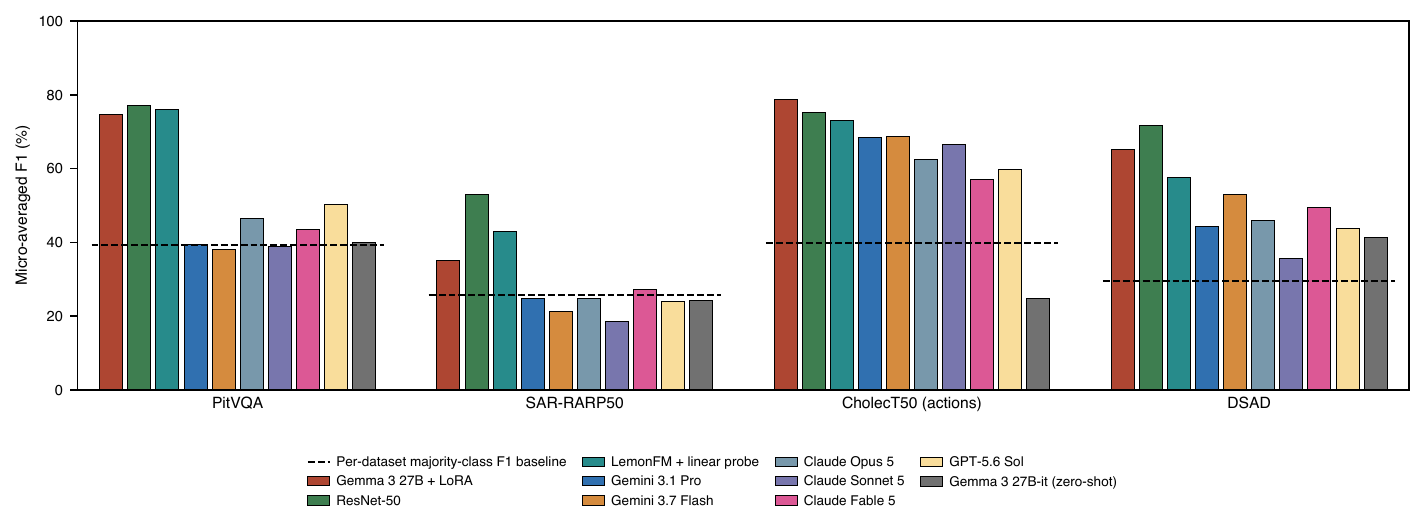}
\caption{The specialist-over-generalist hierarchy replicates on four bounding-box-free surgical annotation tasks. Validation micro-averaged F1 on PitVQA frame-level workflow recognition (3 phases + 14 steps, scored jointly; $n = 24{,}767$ frames from 5 held-out videos), SAR-RARP50 suturing-gesture recognition (8 classes; $n = 636$), CholecT50 surgical action recognition (9 actions + idle, multi-label; $n = 19{,}923$ frames from 10 held-out videos), and DSAD anatomical-structure presence (12 structures, multi-label; $n = 1{,}978$). The models shown mirror the model families of Table~\ref{tab:cross_task_em}, with each frontier family represented by the generation evaluated on these tasks (Gemini~3.1 Pro, Gemini~3.7 Flash, Claude Opus~5, Claude Sonnet~5, Claude Fable~5, GPT-5.6 Sol): Gemma~3 27B fine-tuned with the LoRA classification head, the ResNet-50 supervised specialist, the LemonFM surgical foundation model with a linear probe, the six frontier VLMs, and zero-shot Gemma~3 27B-it. The remaining evaluated models (Claude Fable 5.1, Gemini 3.8 Flash, GPT-5.6 Luna, GPT-5.6 Terra, zero-shot Qwen3.8 27B, and the JSON-head fine-tune) are reported in Tables~\ref{tab:pitvqa_new_ci}--\ref{tab:dsad_new_ci}. Dashed lines indicate per-dataset majority class baselines. Frontier VLMs were evaluated on random seed-42 subsamples ($n = 1{,}000$) of the PitVQA, CholecT50-action, and DSAD validation splits. Values are point estimates; 95\% bootstrap confidence intervals ($B=1{,}000$) for every model are reported in Tables~\ref{tab:pitvqa_new_ci}--\ref{tab:dsad_new_ci}.}
\label{fig:new_tasks_f1}
\end{figure}

\paragraph{Detailed Results} Object detection is only one of the major annotation tasks performed on surgical video, and its bounding-box structure could favor detector architectures such as YOLO. We therefore evaluate four bounding-box-free annotation tasks under a uniform single-frame protocol across five model classes (Section~\ref{subsec:newtasks_methods}): frame-level workflow (phase + step) recognition on PitVQA, fine-grained suturing-gesture recognition on SAR-RARP50, multi-label surgical action recognition on CholecT50, and multi-label anatomy presence on DSAD. Figure~\ref{fig:new_tasks_f1} reports micro-averaged F1 for the leading models of each class, and Table~\ref{tab:cross_task_em} reports exact-match accuracy for every evaluated model---the bounding-box-free companion to the tool-detection summary in Table~\ref{tab:cross_dataset_em}. Both metrics with 95\% bootstrap confidence intervals for all models are reported in Appendix~\ref{sec:appendix_newtasks} (Tables~\ref{tab:pitvqa_new_ci}--\ref{tab:dsad_new_ci}).

Our findings on these tasks, which are less likely to be biased towards object detection, align with the tool-detection results. First, small supervised specialists lead three of the four datasets: 77.1\% micro F1 on PitVQA workflow recognition (majority baseline, 39.4\%), 53.0\% on SAR-RARP50 gestures (baseline, 25.6\%), and 71.6\% on DSAD anatomy presence (baseline, 29.7\%). On CholecT50 action recognition the LoRA classification head leads at 78.8\%, 3.6 points above the specialist (75.2\%; baseline, 39.8\%) and 2.2 points above the JSON-head fine-tune (76.6\%). Second, a frozen surgical foundation model with a task-specific linear probe (LemonFM) ranks second on PitVQA and SAR-RARP50 (76.1\% and 42.9\%) and third on DSAD (57.6\%), where the classification head is second (65.2\%); on CholecT50 actions it ranks fourth (72.9\%), behind both LoRA heads and the specialist---evidence that surgical-domain pretraining, rather than scale, is what transfers most reliably when a large VLM is not adapted to the task. Third, every frontier VLM trails the top two models on every dataset: the best frontier score is 50.3\% on PitVQA (26.8 points behind the specialist; GPT-5.6 Sol), 30.0\% on SAR-RARP50 (23.0 points behind, and the only frontier score there whose 95\% CI clears the majority baseline; Claude Fable~5.1), 70.1\% on CholecT50 actions (8.7 points behind the classification head; Gemini~3.8 Flash), and 54.4\% on DSAD (17.2 points behind; Gemini~3.8 Flash). Both zero-shot open-weight VLMs sit at or below the majority baseline on PitVQA and SAR-RARP50, and diverge sharply elsewhere---Gemma~3 27B is the stronger on DSAD (41.3\% vs.\ a 29.7\% baseline) yet collapses to 24.9\% on CholecT50 actions, where Qwen3.8 27B reaches 53.4\%; on every dataset both trail the linear probe and the supervised specialist. LoRA fine-tuning of Gemma~3 27B narrows the gap and, on action recognition, reverses it. With the autoregressive JSON head, fine-tuning helps on PitVQA, raising micro F1 from 40.1\% zero-shot to 64.4\%, and on CholecT50 actions from 24.9\% to 76.6\%---but on the two smaller training splits it changes micro F1 by at most $+0.3$ points (DSAD) and otherwise degrades it (SAR-RARP50). The linear classification head absorbs the same supervision more efficiently (78.8\% on CholecT50 actions, 74.6\% on PitVQA, 65.2\% on DSAD, 35.2\% on SAR-RARP50). Task-specific supervision is thus absorbed by both small specialists and an adapted 27B VLM; it is not absorbed by frontier scale alone.

CholecT50 action recognition is the friendliest of the four tasks for frontier VLMs---identifying what an instrument is doing to tissue sits closer to the general visual-reasoning distribution than naming a workflow step or an anatomical plane---and the best frontier model accordingly clears the majority baseline by 30 points (70.1\% vs.\ 39.8\%). The ordering still places every frontier VLM below surgical-domain training: the frozen surgical foundation model with a linear probe reaches 72.9\%, the 25.6M-parameter specialist 75.2\%, the LoRA JSON head 76.6\%, and the LoRA classification head 78.8\%, while zero-shot Gemma~3 27B manages only 24.9\%, 15 points below the trivial baseline. Even on the task most favorable to generalists, surgical-domain training---whether pretraining a foundation model, fitting a small specialist, or adapting an open-weight VLM---remains worth more than frontier scale.
\subsection{Cross-Dataset Summary}
\label{subsec:cross_dataset_summary}

\paragraph{Takeaways}
The same pattern holds across all eight datasets and both task families: a small supervised specialist or a task-adapted open-weight VLM leads every dataset, every frontier VLM trails the leaders by a wide margin, and zero-shot open-weight VLMs sit at or below the trivial majority class baseline on most datasets. Because the two task families were evaluated with different specialists (YOLOv12-m vs.\ ResNet-50) and different frontier generations, we report them as two tables rather than one grid: Table~\ref{tab:cross_dataset_em} for the four tool-detection benchmarks and Table~\ref{tab:cross_task_em} for the four bounding-box-free tasks. Closed-weight frontier VLMs were not evaluated on SDSC-EEA because the data cannot be transmitted to third-party APIs for privacy reasons.

\paragraph{Detailed Results}
Tables~\ref{tab:cross_dataset_em} and~\ref{tab:cross_task_em} consolidate exact-match results from Sections~\ref{subsec:lora_classification_head}, \ref{subsec:yolo_performance}, \ref{subsec:robustness} and~\ref{subsec:newtasks}. The per-dataset tables with 95\% bootstrap confidence intervals are provided in Appendix~\ref{sec:appendix_cross_dataset_ci} for the four tool-detection benchmarks (Tables~\ref{tab:sdsc_eea_all_models_ci}, \ref{tab:cholect50_all_models_ci}, \ref{tab:pitvis_all_models_ci}, \ref{tab:surgvu_all_models_ci}) and in Appendix~\ref{sec:appendix_newtasks} for the four bounding-box-free tasks (Tables~\ref{tab:pitvqa_new_ci}--\ref{tab:dsad_new_ci}).

\begin{table}[H]
\centering
\caption{Tool detection: exact-match accuracy (\%) on the validation split of each of the four tool-detection benchmarks. Rows index models; columns index datasets. Entries marked --- were not evaluated: closed-weight frontier VLMs cannot be run on SDSC-EEA because the data is private and cannot be sent to third-party APIs, and the LoRA JSON head was run only on SDSC-EEA. Best per column in bold. Per-dataset 95\% bootstrap confidence intervals are in Appendix~\ref{sec:appendix_cross_dataset_ci}. The models marked $\boldsymbol{^*}$ were evaluated on a random seed-42 subsample of 1{,}000 validation frames per dataset rather than the full validation set, to fit within our API budget. Table~\ref{tab:cross_task_em} is the companion summary for the four bounding-box-free tasks.}
\label{tab:cross_dataset_em}
\small
\begin{tabular}{lcccc}
\toprule
Model & SDSC-EEA & CholecT50 & PitVis-2023 & SurgVU \\
\midrule
Gemma 3 27B + LoRA cls.\ head & 51.08          & \textbf{83.02} & \textbf{84.77} & 50.61 \\
Gemma 3 27B + LoRA (JSON)     & 47.63          & ---            & ---            & --- \\
YOLOv12-m                     & \textbf{54.73} & 81.37          & 82.78          & \textbf{51.75} \\
\midrule
Claude Fable 5.1$^*$          & ---            & 61.70          & 58.70          & 21.00 \\
Gemini 3.8 Flash$^*$          & ---            & 73.40          & 58.40          & 19.80 \\
Gemini 3.1 Pro Preview        & ---            & 66.21          & 57.65          & 22.46 \\
Claude Fable 5$^*$            & ---            & 54.70          & 57.20          & 20.09 \\
Gemini 3 Flash Preview        & ---            & 69.15          & 52.62          & 11.69 \\
Qwen3.8 Max$^*$               & ---            & 70.90          & 52.16          & 17.20 \\
Claude Opus 4.6               & ---            & 52.37          & 50.74          & 17.22 \\
Claude Sonnet 4.6             & ---            & 30.73          & 50.03          & 23.05 \\
GPT-5.4                       & ---            & 32.09          & 49.35          & 17.64 \\
Kimi K3$^*$                   & ---            & 61.70          & 48.80          & 16.39 \\
GPT-5.6 Sol$^*$               & ---            & 55.80          & 39.40          & 17.40 \\
\midrule
Gemma 3 27B-it (zero-shot)    & 9.83           & 6.87           & 28.46          & 2.90 \\
\midrule
Majority class baseline       & 13.41          & 34.76          & 39.63          & 16.94 \\
\bottomrule
\end{tabular}
\end{table}

\begin{table}[H]
\centering
\caption{Bounding-box-free tasks: exact-match accuracy (\%) on the validation split of each of the four bounding-box-free annotation tasks. Rows index models; columns index datasets, with each dataset's annotation task in parentheses. All four datasets are public and every model was evaluated on all four, so there are no missing entries. Exact match requires the full predicted label set to equal the ground truth---for PitVQA, both the phase and the step---so it is a strict all-or-nothing criterion on the two multi-label tasks (CholecT50 action recognition and DSAD). Best per column in bold. Micro-averaged F1 for the same models is shown in Figure~\ref{fig:new_tasks_f1}, and both metrics with 95\% bootstrap confidence intervals are in Appendix~\ref{sec:appendix_newtasks}. Every frontier VLM was evaluated on a random seed-42 subsample of 1{,}000 frames, except on SAR-RARP50, whose 636-frame validation split is scored in full for every model. Table~\ref{tab:cross_dataset_em} is the companion summary for the four tool-detection benchmarks.}
\label{tab:cross_task_em}
\small
\begin{tabular}{lcccc}
\toprule
                              & PitVQA         & SAR-RARP50     & CholecT50      & DSAD \\
Model                         & (workflow)     & (gesture)      & (action)       & (anatomy) \\
\midrule
ResNet-50 (supervised)        & \textbf{66.42} & \textbf{52.99} & 52.94          & \textbf{30.59} \\
Gemma 3 27B + LoRA cls.\ head & 64.62          & 35.22          & \textbf{60.65} & 27.86 \\
LemonFM + linear probe        & 63.27          & 42.92          & 49.76          & 17.29 \\
Gemma 3 27B + LoRA (JSON)     & 50.93          & 22.48          & 60.00          & 10.82 \\
\midrule
GPT-5.6 Sol$^*$               & 32.10          & 23.90          & 30.40          & 5.20 \\
Claude Opus 5$^*$             & 30.60          & 24.84          & 40.20          & 2.60 \\
GPT-5.6 Terra$^*$             & 30.40          & 21.86          & 34.70          & 5.10 \\
GPT-5.6 Luna$^*$              & 29.90          & 22.33          & 29.60          & 3.30 \\
Gemini 3.8 Flash$^*$          & 24.60          & 21.50          & 39.80          & 14.10 \\
Gemini 3.7 Flash$^*$          & 23.70          & 21.23          & 32.20          & 12.10 \\
Claude Sonnet 5$^*$           & 23.20          & 18.55          & 45.40          & 6.10 \\
Gemini 3.1 Pro Preview$^*$    & 22.00          & 24.84          & 37.80          & 7.80 \\
Claude Fable 5$^*$            & 18.10          & 27.20          & 35.70          & 9.80 \\
Claude Fable 5.1$^*$          & 14.60          & 30.00          & 31.30          & 11.00 \\
\midrule
Gemma 3 27B-it (zero-shot)    & 25.06          & 24.21          & 1.54           & 1.62 \\
Qwen3.8 27B (zero-shot)       & 12.32          & 23.27          & 31.34          & 2.28 \\
\midrule
Majority class baseline       & 29.73          & 25.63          & 18.46          & 1.42 \\
\bottomrule
\end{tabular}
\end{table}

Across all eight datasets---SDSC-EEA, CholecT50, PitVis-2023, SurgVU, PitVQA, SAR-RARP50, CholecT50 actions, and DSAD---we observe consistent patterns: zero-shot open-weight VLMs fail to surpass simple baselines, frontier closed-weight VLMs improve substantially but remain well below fine-tuned models on every public dataset, and a small specialized object detection model is competitive with or better than the fine-tuned 27B-parameter VLM at roughly three orders of magnitude smaller scale (26M vs.\ 27B parameters, $\approx 1{,}000\times$ fewer). The absolute accuracy of frontier VLMs varies dramatically across the three public datasets (best frontier model: 73\%$^*$ on CholecT50, 59\%$^*$ on PitVis-2023, 23\% on SurgVU), but on every dataset the gap to the fine-tuned open-weight model and to YOLOv12-m is at least 8~percentage points and as large as 28--29~percentage points on SurgVU. The two datasets where validation accuracy increases smoothly with LoRA rank (CholecT50 and PitVis-2023) are characterized by a relatively small instrument vocabulary and a more uniform tool distribution across cases; the two datasets where validation accuracy plateaus despite increasing rank (SDSC-EEA and SurgVU) feature both a longer-tailed tool distribution and stronger procedure- or session-level distribution shift between train and validation splits.

%% file: sections/4_discussion.tex
\section{Discussion}
\label{sec:discussion}

Our results on surgical tool detection demonstrate that significant obstacles still exist when training generally-capable AI architectures to perform specialized medical tasks. As shown in Section~\ref{subsec:zero_shot} (Table~\ref{tab:zero_shot}, Figures~\ref{fig:accuracy_vs_params}--\ref{fig:accuracy_vs_mmbench}), 20 open-weight VLMs spanning 2B--235B parameters fail to surpass a trivial majority class baseline on surgical tool detection, despite steady gains on general benchmarks. This underperformance is not driven by output format failures: larger and more recent models produce parseable predictions on more than 99\% of frames. For example, the format parsing failure rates with Gemma 3 27B-it is 0.84\%, 0.86\% with Qwen3-VL-32B-Instruct, and 0.97\% with Gemma 4 31B-it (full per-model rates in Appendix~\ref{sec:appendix_format_failures}). Fine-tuning closes part of the gap---Section~\ref{subsec:lora_classification_head} achieves 51.08\% exact match accuracy with a classification head---but a persistent train--validation gap (Figure~\ref{fig:s16_loss}) and the rank sweep in Section~\ref{subsec:lora_rank_sweep} (Figure~\ref{fig:accuracy_vs_rank}) confirm that scaling adapter capacity alone does not resolve the generalization bottleneck. Thus, our results suggest advancing surgical AI may require more task-specialized approaches and---contrary to conventional wisdom---may not be directly solvable by the ``scaling law'' approach of increasing computation or architecture size \citep{hestness2017deep, kaplan2020scalinglawsneurallanguage,ho2025rosetta}.

Meanwhile, Section~\ref{subsec:yolo_performance} shows that YOLOv12-m, a 26M-parameter model, outperforms all VLM-based approaches (Table~\ref{tab:yolo_vs_gemma}), and Section~\ref{subsec:robustness} reproduce the same pattern---including on the same eleven frontier VLMs from the GPT, Gemini, Claude, Kimi, and Qwen families on every public dataset---across three independent public datasets: CholecT50, PitVis-2023, and SurgVU. Section~\ref{subsec:cross_dataset_summary} consolidates the headline exact-match accuracies into two cross-dataset tables, one per task family (Tables~\ref{tab:cross_dataset_em} and~\ref{tab:cross_task_em}); per-tool comparisons against YOLOv12-m on each public dataset are in Tables~\ref{tab:cholect50_yolo_vs_gemma}, \ref{tab:pitvis_yolo_vs_gemma}, and~\ref{tab:surgvu_yolo_vs_gemma}, and rank sweeps in Figures~\ref{fig:cholect50_accuracy_vs_rank}, \ref{fig:pitvis_accuracy_vs_rank}, and~\ref{fig:surgvu_accuracy_vs_rank}; all four datasets except SDSC-EEA are available in the public domain and may have been included in the training of the underlying VLMs. Taken together, these results show that specialized computer vision models match or outperform VLMs at one thousandth of the cost measured in the number of parameters, which is directly proportional to compute and latency. The efficiency and superiority of these specialized models indicate that the next critical advances in Surgical AI will most likely arise from a focused large-scale community effort to defragment data, achieve consensus and labeling at scale, and make data and labels available in an administrative/operational context that would facilitate better training of such specialized models.

Nor are these findings an artifact of the detection task. On four annotation tasks that involve no bounding boxes---workflow recognition, gesture recognition, action recognition, and anatomy presence---the hierarchy replicates (Section~\ref{subsec:newtasks}): small supervised specialists or a task-adapted open-weight VLM lead, a frozen surgical foundation model with a linear probe follows, and ten frontier VLMs from the latest model generation trail the leaders on every dataset (Figure~\ref{fig:new_tasks_f1}, Tables~\ref{tab:pitvqa_new_ci}--\ref{tab:dsad_new_ci}). Our thesis aligns with the position of a recent NEJM AI editorial arguing that specialized models remain necessary in medical AI \citep{hollon2026benchmarking}.

\paragraph{Complementarity with Related Works.} We surveyed widely used medical text and vision-language benchmarks in Appendix~\ref{sec:appendix_benchmark_survey}'s Table~\ref{tab:medical_benchmark_survey}, including MMLU medical and biology subsets \citep{hendryckstest2021}, PubMedQA \citep{jin2019pubmedqa}, MedQA \citep{jin2021disease}, MedMCQA \citep{pmlr-v174-pal22a}, SLAKE \citep{liu2021slakesemanticallylabeledknowledgeenhanceddataset}, PMC-VQA \citep{zhang2024pmcvqavisualinstructiontuning}, OmniMedVQA \citep{hu2024omnimedvqanewlargescalecomprehensive}, MedXpertQA and MedXpertQA-MM \citep{zuo2025medxpertqabenchmarkingexpertlevelmedical}, MultiMedEval \citep{pmlr-v250-royer24a}, and MedFrameQA \citep{yu2026medframeqamultiimagemedicalvqa}. These resources cover text, radiology, pathology, dermatology, ophthalmology, endoscopy, microscopy, documents, charts, and other medical data types, but they do not include surgical video or intraoperative surgical tool-recognition modalities. This absence helps explain why strong performance on broad medical benchmarks may not transfer to the operating-room perception tasks evaluated in this paper.

Recent surgery-specific benchmarks and datasets, including CholecT50~\citep{cholec}, PitVis-2023~\citep{pitvis}, SurgVU~\citep{surgvu}, SurgXBench~\citep{cheng2025surgxbenchexplainablevisionlanguagemodel}, SUREON~\citep{perez2026sureonbenchmarkvisionlanguagemodelsurgical}, and Surg$\Sigma$/Surg-DB~\citep{zeng2026surgsigmaspectrumlargescalemultimodal}, have substantially advanced evaluation of surgical video understanding by targeting capabilities that broad medical benchmarks often miss, including fine-grained instrument recognition and localization, instrument--verb--target interactions, workflow/phase/action understanding, safety assessment, and surgical reasoning. At the same time, most broad medical benchmarks summarized in Table~\ref{tab:medical_benchmark_survey} do not directly evaluate intraoperative surgical video understanding. This distinction matters because strong performance on general medical or multimodal leaderboards is not sufficient evidence of competence in operative perception. Our results provide a complementary view: even recent generalist VLMs remain weak on SDSC-EEA, and the same pattern is observed across CholecT50, PitVis-2023, and SurgVU, whereas task-specific training and smaller specialized vision models perform substantially better.

For a broader perspective on VLMs in Surgical AI, we recommend the concurrent work of \citet{rau2025systematicevaluationlargevisionlanguage}, who systematically benchmark 11 state-of-the-art VLMs across 13 datasets and 17 visual-understanding tasks spanning laparoscopic, robotic, and open procedures, organized by complexity into scene comprehension, surgical-progression understanding, and safety/performance assessment. They additionally study in-context learning (one, three, or five examples per class) and out-of-domain comparisons against task-specific supervised models. They find that VLMs can sometimes outperform supervised models when deployed outside the supervised model's training distribution, that few-shot prompting can boost task performance up to threefold, and that spatial localization and temporal reasoning remain difficult. In another related and concurrent study, \citet{poudel2026evaluatinglargevisionlanguagemodels} evaluate three open-weight VLMs (Qwen2.5-VL, LLaVA-1.5, and InternVL3.5) for surgical tool detection on the GraSP robot-assisted prostatectomy dataset and likewise find that zero-shot performance is poor for two of the three models and that LoRA fine-tuning substantially reduces detection errors across all of them. Very much complementing both \citet{rau2025systematicevaluationlargevisionlanguage}'s and \citet{poudel2026evaluatinglargevisionlanguagemodels}'s findings, this paper conducts extensive experiments (utilizing 35 distinct vision--language models, a surgical foundation model, and two specialist architectures across seven surgical datasets and five annotation-task families) to assess how potential future scaling in computational power, model size, and finetuning methodology could impact the performance of both VLM and specialized AI models.

For a review of surgical AI applications and the broader translational landscape, we also recommend \citet{annurev:/content/journals/10.1146/annurev-bioeng-110824-114236}'s survey. The review summarizes how AI methods have been applied to laparoscopic video for surgical scene understanding, workflow analysis, and intraoperative decision support, and highlights persistent barriers around data, modeling, performance evaluation, and clinical translation. Our study complements this perspective by providing a direct empirical benchmark of current generalist vision-language models on surgical tool detection, showing that the practical capabilities needed for reliable surgical perception remain limited even under substantial model and compute scaling.

\cite{che2026lemonlargeendoscopicmonocular} consider large-scale surgery-specific visual pretraining. LEMON aggregates a large curated corpus of endoscopic surgical video, and LemonFM---a vision foundation model---uses self-supervised pretraining on this corpus to improve performance across multiple modalities. This result is consistent with our central finding: surgical perception failures are not a consequence of insufficient model scale and domain-specific data is needed. Our own evaluation reinforces the point: a frozen LemonFM backbone with a per-task linear probe outranks every frontier VLM on all four bounding-box-free tasks (Section~\ref{subsec:newtasks}), even though it is orders of magnitude smaller than the frontier models it beats.

Together, these works and ours paint a nuanced picture of the opportunities and obstacles towards building reliable AI for surgery. For example, \cite{rau2025systematicevaluationlargevisionlanguage} map what today's VLMs can do across tasks with in-context learning; \cite{cheng2025surgxbenchexplainablevisionlanguagemodel} use explainable AI methods to show why VLMs fail in surgical tasks; \cite{annurev:/content/journals/10.1146/annurev-bioeng-110824-114236} present a review of surgical understanding in AI methods for laparoscopic video analysis; \cite{che2026lemonlargeendoscopicmonocular} demonstrate the value of large-scale, surgery-specific endoscopic pretraining through LEMONFM; and our paper explores the potential and limitations of model, compute, and data scaling for the future development of surgical AI.

\paragraph{Specificity vs Generalism.}  Our results suggest that small specialized models tend to outperform large generalist VLMs when the target task is well-specified and structurally complex---narrow in scope yet demanding fine-grained, consistent perception---as is the case for intraoperative surgical tool detection. This view is complementary to evidence that, when the objective is breadth of capability rather than narrow specialization, generalist VLMs can be the more attractive choice: \citet{rau2025systematicevaluationlargevisionlanguage}, for instance, report out-of-domain experiments in which contemporary VLMs generalize across heterogeneous surgical datasets and tasks more gracefully than task-specific supervised models. Read together, the two perspectives are consistent and motivate a division of labor in which specialized models are deployed within their narrow regime of strength while VLMs are reserved for tasks that require broader, more flexible reasoning across surgical contexts.

Our results also suggest that one way to reconcile generality with performance is to treat the VLM as an orchestrator that can fit or select specialized perception modules on demand. The complementary strengths visible in Table~\ref{tab:yolo_vs_gemma}---where YOLO leads on recall and F1 across all tools while Gemma leads on ROC-AUC for 8 of 15 tools---suggest that hybrid systems combining both model types could outperform either alone. A promising direction for future research is exploring the best approaches for build and improving such hierarchical systems, with the generalist model delegating to specialized models for high-precision subtasks.

\paragraph{Need for Community-Driven Progress: Perspectives from the SDSC.} Some authors of this paper are members of the \citeauthor{sdsc2026}, a nonprofit dedicated to advancing open, collaborative, and clinically-grounded approaches to surgical AI. Our experiences in developing collaborative AI tools for surgeons suggest that assembling large-scale data, ontologies, and labels is a critical prerequisite step for building useful clinical tools. This involves building curated datasets, domain-specific innovations, and efficient annotation frameworks. Moreover, such an effort must be supported by a community- and consensus-focused effort, led by coalitions of aligned organizations. By fostering multi-institutional collaboration, standardizing data-sharing protocols, and developing open access tools, SDSC and similar organizations seek to rapidly advance surgical AI to achieve clinical relevance, equitable access, and real world impact.

The SDSC believes that surgical AI is constrained less by model scale than by the availability of clinically relevant data. Despite advances in foundation models, both the SDSC's practical experience and the results in this paper indicate that performance on basic perceptual tasks remains limited under realistic distribution shift---as evidenced by the widening train--validation gap across LoRA ranks (Section~\ref{subsec:lora_rank_sweep}, Figure~\ref{fig:accuracy_vs_rank}) and the uneven per-tool recall driven by procedure-level tool imbalance (Section~\ref{subsec:lora_formatting}, Table~\ref{tab:tool_distribution}). This indicates that significant gaps remain in domain-specific data coverage and suggests that improvements will depend on the development of large-scale, standardized surgical datasets that capture variability across procedures, institutions, and patient populations. In this setting, the SDSC and similar organizations can play an important role in enabling cross-institutional data aggregation and establishing shared standards. 

In addition to the need for specialized data, this paper's findings indicate that the most promising path forward may not be pushing towards more powerful end-to-end AI models, but rather developing hybrid systems in which generalist models are complemented by specialized components, consistent with the strong performance of smaller task-specific models (Sections~\ref{subsec:yolo_performance}, \ref{subsec:robustness}). Accordingly, the research and development of such hybrid models has been a key focus at the SDSC \citep{masson2024ai, cook2025enhancing, cook20251255}.

\section{Exploratory Next Steps: Natural Language vs. Operating Room}

As an exploratory next step, in Appendix~\ref{sec:appendix_gemma_responses}, we conduct a preliminary experiment where LLMs appear to give nearly entirely correct responses to questions relating to pituitary tumor surgery. However, as we have demonstrated earlier, the same models fail at a simple task of tool detection. This result is not surprising: consider that neurosurgeons train primarily through practice. The Accreditation Council for Graduate Medical Education (ACGME) requires 7 years of residency, typically completed after 2 years of rotations in medical school, compared to only 2 years of classroom and anatomy lab education \cite{acgme}. The idea that tacit knowledge in such jobs is more important than what can be written down is not new. In the context of job market automation, this is often referred to as Polanyi's paradox \cite{NBERw20485}. From this preliminary experiment, we hypothesize that the challenge with medical AI is that the data for pretraining foundation models lacks information from the operating room.

%% file: sections/5_limitations.tex
\section{Limitations}
\label{sec:limitations}

This study has several limitations. First, although we extend the evaluation beyond tool detection to workflow, gesture, action, and anatomy recognition (Section~\ref{subsec:newtasks}), every evaluation is frame-level: no model receives temporal context, and we do not evaluate higher-order capabilities such as decision support or anomaly detection. It is possible that VLMs offer greater advantages on these more abstract tasks, where language-mediated reasoning plays a larger role. On the four bounding-box-free tasks, frontier VLMs were also scored on 1{,}000-frame subsamples of the three larger validation splits, so their confidence intervals are wider than those of the locally evaluated models. Second, our VLM evaluation focuses on open-weight models with a specific prompting and decoding setup. Stronger closed-source models, alternative prompting strategies, or more extensive instruction tuning could yield different results. Third, the degree to which our conclusions generalize to other surgical specialties, institutions, and recording conditions remains an open question, although the consistency of the takeaways on CholecT50, PitVis-2023, and SurgVU with those that we found on our own data suggests the broad pattern holds across four distinct surgical domains, including endoscopic neurosurgery, videos of laparoscopic cholecystectomy surgery, and robotic-assisted training sessions on porcine tissue. Fourth, the SurgVU labels we use are not direct visual ground truth: they are derived from temporal install/uninstall events automatically harvested from the da~Vinci robot arms (Section~\ref{subsec:surgvu_methods}), which means the per-frame label records ``tool installed and active on an arm during this time interval'' rather than ``tool visible in this specific cropped frame.'' A tool can therefore be labelled present while being momentarily off-screen, occluded by tissue or by another instrument, or entirely outside the cropped field of view used by every model. This label-source mismatch likely depresses zero-shot VLM scores (which can only score frames they actually see) more than it does fine-tuned models (which can learn session- and task-level instrument priors that compensate for transient invisibility); this asymmetry should be kept in mind when interpreting the SurgVU gap between zero-shot and fine-tuned models. CholecT50 and PitVis-2023 ship per-frame human visual annotations and are not subject to this issue. Fifth, while we did our best to conduct scaling experiments within our computational means, it remains possible that future models may show non-linear ``emergent'' jumps in performance if model size and training duration scale past a yet-to-be-discovered threshold \citep{wei2022emergentabilitieslargelanguage}.

%% file: sections/6_conclusion.tex
\section{Conclusion}
\label{sec:conclusion}

In this paper, we evaluate how much recent progress in large vision-language models and scaling can bring us closer towards better Surgical AI with surgical tool detection in endoscopic endonasal neurosurgery as a case study supported by additional robustness check experiments in other surgical domains. Section~\ref{subsec:zero_shot} shows that across 20 open-weight VLMs spanning 2B to 235B parameters, zero-shot performance on held-out procedures remains at or near a trivial majority class baseline, despite large gains on general benchmarks like MMBench. Section~\ref{subsec:lora_classification_head} demonstrates that LoRA fine-tuning with a classification head improves performance substantially, reaching 51.08\% exact match accuracy, but a persistent train--validation gap reflects limited generalization under procedure-level distribution shift. Section~\ref{subsec:lora_rank_sweep} rules out insufficient capacity as the cause: scaling LoRA rank by nearly three orders of magnitude saturates training accuracy near 99\% while validation accuracy remains below 40\%. Section~\ref{subsec:yolo_performance} shows that YOLOv12-m, a 26M-parameter object detection model---over 1,000$\times$ smaller than the VLM---outperforms all VLM-based approaches at a fraction of the training time and inference cost. Section~\ref{subsec:robustness} replicates these findings, including comparisons with the same eleven frontier VLMs from the GPT, Gemini, Claude, Kimi, and Qwen families on every public dataset, on three independent public datasets---CholecT50 (laparoscopic cholecystectomy), PitVis-2023 (endoscopic pituitary neurosurgery), and SurgVU (robotic-assisted training sessions on porcine tissue)---confirming that the same patterns hold across all four surgical domains: zero-shot open-weight VLMs underperform trivial baselines, closed-weight frontier VLMs improve substantially but remain well below fine-tuned models on every dataset (with the gap widening to 28--29~percentage points on SurgVU, where only five of the eleven frontier models clearly exceed the trivial baseline), and a small specialized model is competitive with or better than the fine-tuned 27B-parameter VLM at roughly three orders of magnitude smaller scale (26M vs.\ 27B parameters, $\approx 1{,}000\times$ fewer). Section~\ref{subsec:newtasks} shows that this hierarchy is not an artifact of the detection task: on four bounding-box-free annotation tasks---workflow recognition (PitVQA), gesture recognition (SAR-RARP50), action recognition (CholecT50), and anatomy presence (DSAD)---small supervised specialists or a task-adapted open-weight VLM lead every dataset, a frozen surgical foundation model with a linear probe follows, and all ten frontier VLMs trail the top two models on every dataset. Across 37 distinct vision--language models, seven surgical datasets, and five annotation-task families, the same conclusion holds.

Our findings suggest that progress toward reliable surgical AI is likely more constrained by limited amounts of specialized data than by the scale of potential AI architectures and training resources. Small specialized models can outperform large foundation models on narrow surgical tasks while being orders of magnitude more efficient. Thus, future efforts to pool and label surgical data across institutions will be crucial to improving Surgical AI capabilities.

%% file: sections/7_appendix.tex
\section{Per-Dataset Exact-Match Accuracy with Confidence Intervals}
\label{sec:appendix_cross_dataset_ci}

This appendix provides the per-dataset companions to the consolidated cross-dataset summary in Table~\ref{tab:cross_dataset_em}. For each dataset, models are reported with their parameter count, validation exact-match accuracy, and 95\% bootstrap confidence interval ($B=1{,}000$). Tables~\ref{tab:sdsc_eea_all_models_ci}--\ref{tab:surgvu_all_models_ci} use the same row ordering and grouping as Table~\ref{tab:cross_dataset_em} so the two views can be read in parallel. Closed-weight frontier VLMs were not evaluated on SDSC-EEA because the data is private and cannot be transmitted to third-party APIs; rows for those models are accordingly absent from Table~\ref{tab:sdsc_eea_all_models_ci}.

\begin{table}[H]
\centering
\caption{Tool detection exact match accuracy (\%) on the SDSC-EEA validation set ($n=20{,}016$ frames, 13 procedures) with 95\% bootstrap confidence intervals ($B=1{,}000$). Closed-weight frontier VLMs were not evaluated on SDSC-EEA (the data cannot be sent to third-party APIs). The majority class baseline predicts the most common tool set for every frame. Best in bold.}
\label{tab:sdsc_eea_all_models_ci}
\small
\begin{tabular}{lrll}
\toprule
Model & Params (B) & EM \% & 95\% CI \\
\midrule
YOLOv12-m                     & 0.026 & \textbf{54.73} & \textbf{54.03--55.44} \\
Gemma 3 27B + LoRA cls.\ head & 27    & 51.08          & 50.39--51.81 \\
Gemma 3 27B-it (zero-shot)    & 27    & 9.83           & 9.43--10.21 \\
\midrule
Majority class baseline       & ---   & 13.41          & --- \\
\bottomrule
\end{tabular}
\end{table}

\begin{table}[H]
\centering
\caption{Tool detection exact match accuracy (\%) on the CholecT50 validation set ($n=19{,}923$ frames, 6 instrument classes) with 95\% bootstrap confidence intervals ($B=1{,}000$). The majority class baseline predicts the most common tool set for every frame. Output validation failures are counted as incorrect predictions. Best in bold. $^*$Evaluated on a random subsample of 1{,}000 validation frames.}
\label{tab:cholect50_all_models_ci}
\small
\begin{tabular}{lrll}
\toprule
Model & Params (B) & EM \% & 95\% CI \\
\midrule
Gemma 3 27B + LoRA cls.\ head & 27    & \textbf{83.02} & \textbf{82.52--83.56} \\
YOLOv12-m                     & 0.026 & 81.37          & 80.87--81.92 \\
Gemini 3.8 Flash$^*$          & ---   & 73.40          & 70.80--75.90 \\
Qwen3.8 Max$^*$               & ---   & 70.90          & 68.00--73.60 \\
Gemini 3 Flash Preview        & ---   & 69.15          & 68.49--69.73 \\
Gemini 3.1 Pro Preview        & ---   & 66.21          & 65.58--66.88 \\
Claude Fable 5.1$^*$          & ---   & 61.70          & 58.80--64.80 \\
Kimi K3$^*$                   & ---   & 61.70          & 58.90--64.90 \\
GPT-5.6 Sol$^*$               & ---   & 55.80          & 52.90--58.90 \\
Claude Fable 5$^*$            & ---   & 54.70          & 51.60--57.70 \\
Claude Opus 4.6               & ---   & 52.37          & 51.67--53.03 \\
GPT-5.4                       & ---   & 32.09          & 31.40--32.72 \\
Claude Sonnet 4.6             & ---   & 30.73          & 30.07--31.37 \\
Gemma 3 27B-it (zero-shot)    & 27    & 6.87           & 6.55--7.22 \\
\midrule
Majority class baseline       & ---   & 34.76          & --- \\
\bottomrule
\end{tabular}
\end{table}

\begin{table}[H]
\centering
\caption{Tool detection exact match accuracy (\%) on the PitVis-2023 validation set ($n=30{,}896$ frames, 18 instrument classes) with 95\% bootstrap confidence intervals ($B=1{,}000$). The majority class baseline predicts the most common tool set (the empty set) for every frame. Output validation failures are counted as incorrect predictions. Best in bold. $^*$Evaluated on a random subsample of 1{,}000 validation frames.}
\label{tab:pitvis_all_models_ci}
\small
\begin{tabular}{lrll}
\toprule
Model & Params (B) & EM \% & 95\% CI \\
\midrule
Gemma 3 27B + LoRA cls.\ head & 27    & \textbf{84.77} & \textbf{84.36--85.16} \\
YOLOv12-m                     & 0.026 & 82.78          & 82.36--83.20 \\
Claude Fable 5.1$^*$          & ---   & 58.70          & 55.60--61.60 \\
Gemini 3.8 Flash$^*$          & ---   & 58.40          & 55.20--61.60 \\
Gemini 3.1 Pro Preview        & ---   & 57.65          & 57.11--58.20 \\
Claude Fable 5$^*$            & ---   & 57.20          & 54.30--60.50 \\
Gemini 3 Flash Preview        & ---   & 52.62          & 52.08--53.18 \\
Qwen3.8 Max$^*$               & ---   & 52.16          & 48.95--55.37 \\
Claude Opus 4.6               & ---   & 50.74          & 50.18--51.31 \\
Claude Sonnet 4.6             & ---   & 50.03          & 49.49--50.55 \\
GPT-5.4                       & ---   & 49.35          & 48.81--49.92 \\
Kimi K3$^*$                   & ---   & 48.80          & 45.60--51.90 \\
GPT-5.6 Sol$^*$               & ---   & 39.40          & 36.30--42.60 \\
Gemma 3 27B-it (zero-shot)    & 27    & 28.46          & 27.97--28.94 \\
\midrule
Majority class baseline       & ---   & 39.63          & --- \\
\bottomrule
\end{tabular}
\end{table}

\begin{table}[H]
\centering
\caption{Tool detection exact match accuracy (\%) on the SurgVU validation set ($n=18{,}919$ frames, 17 instrument classes) with 95\% bootstrap confidence intervals ($B=1{,}000$). The majority class baseline predicts the most common tool set (the empty set) for every frame. Output validation failures are counted as incorrect predictions. Best in bold. $^*$Evaluated on a random subsample of 1{,}000 validation frames.}
\label{tab:surgvu_all_models_ci}
\small
\begin{tabular}{lrll}
\toprule
Model & Params (B) & EM \% & 95\% CI \\
\midrule
YOLOv12-m                     & 0.026 & \textbf{51.75} & \textbf{50.97--52.51} \\
Gemma 3 27B + LoRA cls.\ head & 27    & 50.61          & 49.84--51.39 \\
Claude Sonnet 4.6             & ---   & 23.05          & 22.46--23.65 \\
Gemini 3.1 Pro Preview        & ---   & 22.46          & 21.86--23.02 \\
Claude Fable 5.1$^*$          & ---   & 21.00          & 18.50--23.70 \\
Claude Fable 5$^*$            & ---   & 20.09          & 17.75--22.34 \\
Gemini 3.8 Flash$^*$          & ---   & 19.80          & 17.40--22.40 \\
GPT-5.4                       & ---   & 17.64          & 17.10--18.24 \\
GPT-5.6 Sol$^*$               & ---   & 17.40          & 15.20--19.90 \\
Claude Opus 4.6               & ---   & 17.22          & 16.73--17.77 \\
Qwen3.8 Max$^*$               & ---   & 17.20          & 14.90--19.70 \\
Kimi K3$^*$                   & ---   & 16.39          & 13.68--19.25 \\
Gemini 3 Flash Preview        & ---   & 11.69          & 11.23--12.16 \\
Gemma 3 27B-it (zero-shot)    & 27    & 2.90           & 2.66--3.16 \\
\midrule
Majority class baseline       & ---   & 16.94          & --- \\
\bottomrule
\end{tabular}
\end{table}

\section{Survey of Medical Benchmarks}
\label{sec:appendix_benchmark_survey}

Table~\ref{tab:medical_benchmark_survey} summarizes the benchmark modalities and dataset sizes used to assess whether prominent medical AI benchmarks include surgical modalities.

\begin{table}[p]
\centering
\caption{Survey of prominent medical AI benchmarks and their reported modalities.}
\label{tab:medical_benchmark_survey}
\scriptsize
\setlength{\tabcolsep}{3pt}
\renewcommand{\arraystretch}{1.12}
\begin{tabular}{@{}p{0.15\textwidth} p{0.33\textwidth} p{0.24\textwidth} r@{}}
\toprule
Benchmark & Modalities & $N$ rows & Year \\
\midrule
MMLU (Medicine + Biology subsets)\newline \citep{hendryckstest2021} & Text; Anatomy, Clinical Knowledge, College Biology, College Medicine, Medical Genetics, Professional Medicine & 1{,}089 & 2021 \\
PubMedQA\newline \citep{jin2019pubmedqa} & Text & 273{,}518 & 2019 \\
MedQA\newline \citep{jin2021disease} & Text & 61{,}097 & 2021 \\
MedMCQA\newline \citep{pmlr-v174-pal22a} & Text & 193{,}155 & 2022 \\
SLAKE\newline \citep{liu2021slakesemanticallylabeledknowledgeenhanceddataset} & CT, MRI, X-Ray & 14{,}028 & 2021 \\
PMC-VQA\newline \citep{zhang2024pmcvqavisualinstructiontuning} & Various modalities/diseases; no fixed closed modality list in the paper & 227k QA / 149k images & 2023 \\
OmniMedVQA\newline \citep{hu2024omnimedvqanewlargescalecomprehensive} & Colposcopy, CT, Digital Photography, Fundus Photography, Infrared Reflectance Imaging, MR/MRI, OCT, Dermoscopy, Endoscopy, Microscopy Images, X-Ray, Ultrasound & 127{,}995 QA / 118{,}010 images & 2024 \\
MedXpertQA\newline \citep{zuo2025medxpertqabenchmarkingexpertlevelmedical} & Text & 2{,}450 & 2025 \\
MedXpertQA-MM\newline \citep{zuo2025medxpertqabenchmarkingexpertlevelmedical} & Radiology, Pathology, Medical Optical Imaging, Photos, Vital Signs, Diagrams, Documents, Charts, Tables, Others & 2{,}000 QA / 2{,}839 images & 2025 \\
MultiMedEval\newline \citep{pmlr-v250-royer24a} & General Medicine, X-Ray, CT, Microscope, Dermatology, Mammology, OCT, Ultrasound, Fundus Camera, Pathology, Radiology & 72{,}834 & 2024 \\
MedFrameQA\newline \citep{yu2026medframeqamultiimagemedicalvqa} & CT, MRI, Ultrasound, X-ray & 2{,}851 QA pairs / 9{,}237 frames / 3{,}420 videos & 2025 \\
\bottomrule
\end{tabular}
\end{table}

\section{Tool Distribution Across Train/Validation Splits}
\label{sec:appendix_tool_distribution}

Table~\ref{tab:tool_distribution} shows the number of frames containing each tool in the training set ($n=47{,}618$ frames, 53 procedures) and validation set ($n=20{,}016$ frames, 13 procedures). Because the split is performed at the procedure level, the per-tool distribution across splits is highly uneven. Several tools appear almost exclusively in one split: for example, Aspirating dissector has 88 training instances versus 2{,}319 validation instances, and Sonopet pineapple tip has 1{,}991 training instances versus zero in validation.

\begin{table}[H]
\centering
\caption{Number of frames containing each tool in the training ($n=47{,}618$) and validation ($n=20{,}016$) splits, with percentage of frames in each split. Sorted by total count in descending order.}
\label{tab:tool_distribution}
\footnotesize
\begin{tabular}{l rr rr r}
\toprule
Tool & Train $N$ & Train \% & Val $N$ & Val \% & Total \\
\midrule
Suction & 29{,}692 & 62.4 & 10{,}685 & 53.4 & 40{,}377 \\
Cotton Patty & 8{,}822 & 18.5 & 2{,}143 & 10.7 & 10{,}965 \\
Grasper & 6{,}812 & 14.3 & 509 & 2.5 & 7{,}321 \\
Rhoton Dissector & 4{,}249 & 8.9 & 1{,}462 & 7.3 & 5{,}711 \\
Curette & 4{,}829 & 10.1 & 708 & 3.5 & 5{,}537 \\
Rongeur & 1{,}730 & 3.6 & 2{,}790 & 13.9 & 4{,}520 \\
Scissor & 2{,}928 & 6.1 & 996 & 5.0 & 3{,}924 \\
Drill & 1{,}639 & 3.4 & 2{,}116 & 10.6 & 3{,}755 \\
Surgicel & 2{,}847 & 6.0 & 739 & 3.7 & 3{,}586 \\
Surgical Knife & 1{,}364 & 2.9 & 1{,}422 & 7.1 & 2{,}786 \\
Suction Coagulator & 1{,}400 & 2.9 & 1{,}188 & 5.9 & 2{,}588 \\
Aspirating dissector & 88 & 0.2 & 2{,}319 & 11.6 & 2{,}407 \\
Bipolar Forceps & 1{,}934 & 4.1 & 263 & 1.3 & 2{,}197 \\
Sonopet pineapple tip & 1{,}991 & 4.2 & 0 & 0.0 & 1{,}991 \\
Straight Forceps & 1{,}602 & 3.4 & 173 & 0.9 & 1{,}775 \\
Doppler & 1{,}592 & 3.3 & 0 & 0.0 & 1{,}592 \\
unknown & 1{,}172 & 2.5 & 23 & 0.1 & 1{,}195 \\
Backbiting rongeur & 144 & 0.3 & 1{,}041 & 5.2 & 1{,}185 \\
Tissue shaver & 934 & 2.0 & 57 & 0.3 & 991 \\
Cottle Elevator & 855 & 1.8 & 0 & 0.0 & 855 \\
Through cutting forceps & 255 & 0.5 & 542 & 2.7 & 797 \\
Suction microdebrider & 94 & 0.2 & 497 & 2.5 & 591 \\
Floseal Gel & 502 & 1.1 & 40 & 0.2 & 542 \\
Collagen Matrix & 280 & 0.6 & 0 & 0.0 & 280 \\
Floseal Applicator & 199 & 0.4 & 40 & 0.2 & 239 \\
Monopolar Electrocautery & 190 & 0.4 & 0 & 0.0 & 190 \\
Local Anesthesia Needle & 139 & 0.3 & 0 & 0.0 & 139 \\
Straight Curette & 102 & 0.2 & 16 & 0.1 & 118 \\
Needle & 31 & 0.1 & 0 & 0.0 & 31 \\
Tisseel Applicator & 23 & 0.0 & 0 & 0.0 & 23 \\
\bottomrule
\end{tabular}
\end{table}

\section{Zero-Shot Evaluation Prompt Template}
\label{sec:appendix_prompt}

The following prompt template is used for zero-shot tool detection evaluation across all vision-language models.

\begin{promptbox}
Identify all surgical tools visible in this surgical video frame.

Valid tool names:
- Aspirating dissector
- Backbiting rongeur
- Bipolar Forceps
- Collagen Matrix
- Cotton Patty
- Cottle Elevator
- Curette
- Doppler
- Drill
- Floseal Applicator
- Floseal Gel
- Grasper
- Irrigation
- Local Anesthesia Needle
- Monopolar Electrocautery
- Needle
- Rhoton Dissector
- Rongeur
- Scissor
- Sonopet pineapple tip
- Straight Curette
- Straight Forceps
- Suction
- Suction Coagulator
- Suction microdebrider
- Surgicel
- Surgical Knife
- Tisseel Applicator
- Tissue shaver
- Through cutting forceps
- unknown

Return your answer as a JSON object with this exact format:
{"detected_tools": ["Tool Name 1", "Tool Name 2"]}

Use ONLY the exact tool names from the list above.
If a tool appears multiple times, list it only once.
If no tools are visible, return: {"detected_tools": []}

Return ONLY valid JSON, nothing else.
\end{promptbox}

\section{Output Validation Methodology}
\label{sec:appendix_output_validation}

Model outputs are parsed by extracting the first valid JSON object from the response text using regex matching. An output is classified as an output validation failure if it meets any of the following conditions: (1) the response does not contain valid JSON or is missing the \texttt{detected\_tools} key (JSON failure), (2) the \texttt{detected\_tools} value is not an array of strings (schema failure), or (3) the array contains tool names that do not exactly match any entry in the provided ontology of 31 valid tool names (ontology failure, e.g., misspellings, capitalization mismatches, or hallucinated tool names). All output validation failures are treated as empty predictions (\texttt{detected\_tools: []}). This ensures that a model's inability to follow the output format or correctly name tools is penalized rather than silently excluded. Representative failure examples are shown in Appendix~\ref{sec:appendix_validation_failures}.

\section{Output Validation Failure Rates by Model on SDSC-EEA}
\label{sec:appendix_format_failures}

Table~\ref{tab:format_failures} reports the percentage of attempted validation frames on which each zero-shot vision-language model produced an output that could not be parsed under the rules in Appendix~\ref{sec:appendix_output_validation}; such frames are coerced to \texttt{detected\_tools: []} for all downstream metrics.

\begin{table}[H]
\centering
\caption{Zero-shot output validation failure rates on SDSC-EEA.}
\label{tab:format_failures}
\small
\begin{tabular}{lr}
\toprule
Model & Format failure \% \\
\midrule
Qwen3-VL-235B-A22B-Thinking & 2.72 \\
Qwen3-VL-32B-Instruct       & 0.86 \\
Qwen3-VL-8B-Instruct        & 0.84 \\
Qwen3-VL-4B-Instruct        & 1.17 \\
Qwen3-VL-2B-Instruct        & 32.75 \\
\midrule
Qwen2.5-VL-72B-Instruct     & 1.60 \\
Qwen2.5-VL-32B-Instruct     & 6.55 \\
Qwen2.5-VL-7B-Instruct      & 16.65 \\
Qwen2.5-VL-3B-Instruct      & 21.94 \\
\midrule
Qwen2-VL-72B-Instruct       & 5.30 \\
Qwen2-VL-7B-Instruct        & 26.59 \\
Qwen2-VL-2B-Instruct        & 41.73 \\
\midrule
Gemma 3 27B-it              & 0.84 \\
Gemma 3 12B-it              & 0.85 \\
Gemma 3 4B-it               & 1.01 \\
\midrule
MedGemma 3 27B-it           & 2.86 \\
\midrule
Llama-3.2-90B-Vision        & 0.01 \\
Llama-3.2-11B-Vision        & 95.72 \\
\midrule
LLaVA-1.5-13B               & 3.26 \\
\midrule
Gemma 4 31B-it              & 0.97 \\
\bottomrule
\end{tabular}
\end{table}

\section{LoRA Fine-Tuning Configuration}
\label{sec:appendix_lora_config}

All fine-tuning experiments use a fixed random seed of 42 for reproducibility. LoRA adapters~\citep{hu2021loralowrankadaptationlarge} are applied to the query, key, value, and output projection matrices in both the language model and vision encoder attention layers (\texttt{q\_proj}, \texttt{k\_proj}, \texttt{v\_proj}, \texttt{o\_proj}, \texttt{out\_proj}).

\paragraph{JSON Generation.}
LoRA rank $r = 1024$, scaling factor $\alpha = 2048$, dropout 0.05. Training: 10 epochs, learning rate $2 \times 10^{-5}$, effective batch size 64 (per-GPU batch size 1 $\times$ 8 gradient accumulation steps $\times$ 8 H200 GPUs), bfloat16 precision. Gradient checkpointing is used to reduce memory consumption. Training is distributed using PyTorch DDP~\citep{li2020pytorchdistributedexperiencesaccelerating}. Training and evaluation required 80 wall-clock hours (640 GPU-hours on H200 GPUs). During training, exact match accuracy and Jaccard similarity are periodically evaluated on fixed random subsets of 100 training and 100 validation frames.

\paragraph{Classification Head.}
The base model processes the image and prompt, and we apply mean pooling over the final hidden states (excluding padding tokens) to obtain a fixed-dimensional representation. A single linear layer (no hidden layers) maps this representation to 31 output logits (one per tool class), trained with binary cross-entropy loss averaged across all tool classes. At inference, we apply a sigmoid activation and threshold at 0.5 to obtain binary predictions. LoRA rank $r = 1024$, $\alpha = 2048$, dropout 0.05. Training: 10 epochs, learning rate $5 \times 10^{-6}$, effective batch size 32 (per-GPU batch size 1 $\times$ 4 gradient accumulation steps $\times$ 8 H200 GPUs).

\paragraph{Rank Sweep.}
We sweep LoRA ranks $r \in \{2, 4, 8, 16, 32, 64, 128, 256, 512, 1024\}$, setting $\alpha = 2r$ for each. Training: 3 epochs per configuration, effective batch size 32, with other settings matching the classification head configuration above. The full sweep required approximately 62 wall-clock hours (492 GPU-hours on H200 GPUs). Trainable parameters scale linearly with rank, from 4.7M at $r=2$ to 2.4B at $r=1024$. For each configuration, we report training and validation exact match accuracy with 95\% confidence intervals.

\section{Per-Tool Metrics for LoRA Fine-Tuning with JSON Output}
\label{sec:appendix_s19_metrics}

Table~\ref{tab:s19_tool_metrics} shows per-tool classification metrics on the validation set ($n=20{,}016$ frames) for Gemma 3 27B fine-tuned with LoRA to produce JSON outputs.

\begin{table}[H]
\centering
\caption{Per-tool classification metrics for Gemma 3 27B with LoRA fine-tuning (JSON output). TP = true positives, FP = false positives, FN = false negatives, TN = true negatives.}
\label{tab:s19_tool_metrics}
\small
\begin{tabular}{lrrrrrrrr}
\toprule
Tool & TP & FP & FN & TN & Accuracy & Precision & Recall & F1 \\
\midrule
Aspirating dissector & 0 & 0 & 2319 & 17697 & 0.884 & 0.000 & 0.000 & 0.000 \\
Backbiting rongeur & 141 & 44 & 900 & 18931 & 0.953 & 0.762 & 0.135 & 0.230 \\
Bipolar Forceps & 0 & 103 & 263 & 19650 & 0.982 & 0.000 & 0.000 & 0.000 \\
Cottle Elevator & 0 & 271 & 0 & 19745 & 0.987 & 0.000 & 0.000 & 0.000 \\
Cotton Patty & 1729 & 639 & 414 & 17234 & 0.947 & 0.730 & 0.807 & 0.767 \\
Curette & 103 & 3 & 605 & 19305 & 0.970 & 0.972 & 0.145 & 0.253 \\
Doppler & 0 & 31 & 0 & 19985 & 0.999 & 0.000 & 0.000 & 0.000 \\
Drill & 1637 & 109 & 479 & 17791 & 0.971 & 0.938 & 0.774 & 0.848 \\
Floseal Applicator & 0 & 17 & 40 & 19959 & 0.997 & 0.000 & 0.000 & 0.000 \\
Floseal Gel & 2 & 90 & 38 & 19886 & 0.994 & 0.022 & 0.050 & 0.030 \\
Grasper & 60 & 666 & 449 & 18841 & 0.944 & 0.083 & 0.118 & 0.097 \\
Irrigation & 9 & 104 & 103 & 19800 & 0.990 & 0.080 & 0.080 & 0.080 \\
Monopolar Electrocautery & 0 & 831 & 0 & 19185 & 0.959 & 0.000 & 0.000 & 0.000 \\
Rhoton Dissector & 799 & 3086 & 663 & 15468 & 0.813 & 0.206 & 0.546 & 0.299 \\
Rongeur & 522 & 1509 & 2268 & 15717 & 0.811 & 0.257 & 0.187 & 0.217 \\
Scissor & 530 & 703 & 466 & 18317 & 0.942 & 0.430 & 0.532 & 0.475 \\
Sonopet pineapple tip & 0 & 1 & 0 & 20015 & 1.000 & 0.000 & 0.000 & 0.000 \\
Straight Forceps & 44 & 406 & 129 & 19437 & 0.973 & 0.098 & 0.254 & 0.141 \\
Suction & 9411 & 4289 & 1274 & 5042 & 0.722 & 0.687 & 0.881 & 0.772 \\
Surgicel & 347 & 36 & 392 & 19241 & 0.979 & 0.906 & 0.470 & 0.619 \\
Through cutting forceps & 1 & 11 & 541 & 19463 & 0.972 & 0.083 & 0.002 & 0.004 \\
Tissue shaver & 0 & 0 & 57 & 19959 & 0.997 & 0.000 & 0.000 & 0.000 \\
unknown & 2 & 1223 & 21 & 18770 & 0.938 & 0.002 & 0.087 & 0.003 \\
\bottomrule
\end{tabular}
\end{table}

\section{Per-Tool Metrics for LoRA Fine-Tuning with Classification Head}
\label{sec:appendix_s16_metrics}

Table~\ref{tab:s16_tool_metrics} shows per-tool classification metrics on the validation set ($n=20{,}016$ frames) for Gemma 3 27B fine-tuned with LoRA and a linear classification head.

\begin{table}[H]
\centering
\caption{Per-tool classification metrics for Gemma 3 27B with LoRA fine-tuning and classification head. TP = true positives, FP = false positives, FN = false negatives, TN = true negatives.}
\label{tab:s16_tool_metrics}
\small
\begin{tabular}{lrrrrrrrr}
\toprule
Tool & TP & FP & FN & TN & Accuracy & Precision & Recall & F1 \\
\midrule
Aspirating dissector & 0 & 0 & 2319 & 17697 & 0.884 & 0.000 & 0.000 & 0.000 \\
Backbiting rongeur & 20 & 7 & 1021 & 18968 & 0.949 & 0.741 & 0.019 & 0.038 \\
Bipolar Forceps & 0 & 10 & 263 & 19743 & 0.986 & 0.000 & 0.000 & 0.000 \\
Collagen Matrix & 0 & 0 & 0 & 20016 & 1.000 & 0.000 & 0.000 & 0.000 \\
Cottle Elevator & 0 & 3 & 0 & 20013 & 1.000 & 0.000 & 0.000 & 0.000 \\
Cotton Patty & 1513 & 1112 & 630 & 16761 & 0.913 & 0.576 & 0.706 & 0.635 \\
Curette & 169 & 9 & 539 & 19299 & 0.973 & 0.949 & 0.239 & 0.382 \\
Doppler & 0 & 1 & 0 & 20015 & 1.000 & 0.000 & 0.000 & 0.000 \\
Drill & 1672 & 98 & 444 & 17802 & 0.973 & 0.945 & 0.790 & 0.861 \\
Floseal Applicator & 0 & 0 & 40 & 19976 & 0.998 & 0.000 & 0.000 & 0.000 \\
Floseal Gel & 2 & 1 & 38 & 19975 & 0.998 & 0.667 & 0.050 & 0.093 \\
Grasper & 10 & 307 & 499 & 19200 & 0.960 & 0.032 & 0.020 & 0.024 \\
Irrigation & 2 & 0 & 110 & 19904 & 0.995 & 1.000 & 0.018 & 0.035 \\
Local Anesthesia Needle & 0 & 1 & 0 & 20015 & 1.000 & 0.000 & 0.000 & 0.000 \\
Monopolar Electrocautery & 0 & 0 & 0 & 20016 & 1.000 & 0.000 & 0.000 & 0.000 \\
Needle & 0 & 0 & 0 & 20016 & 1.000 & 0.000 & 0.000 & 0.000 \\
Rhoton Dissector & 863 & 735 & 599 & 17819 & 0.933 & 0.540 & 0.590 & 0.564 \\
Rongeur & 620 & 26 & 2170 & 17200 & 0.890 & 0.960 & 0.222 & 0.361 \\
Scissor & 135 & 82 & 861 & 18938 & 0.953 & 0.622 & 0.136 & 0.223 \\
Sonopet pineapple tip & 0 & 0 & 0 & 20016 & 1.000 & 0.000 & 0.000 & 0.000 \\
Straight Curette & 0 & 0 & 16 & 20000 & 0.999 & 0.000 & 0.000 & 0.000 \\
Straight Forceps & 43 & 24 & 130 & 19819 & 0.992 & 0.642 & 0.249 & 0.358 \\
Suction & 9452 & 4604 & 1233 & 4727 & 0.708 & 0.673 & 0.885 & 0.764 \\
Suction Coagulator & 739 & 0 & 449 & 18828 & 0.978 & 1.000 & 0.622 & 0.767 \\
Suction microdebrider & 0 & 0 & 497 & 19519 & 0.975 & 0.000 & 0.000 & 0.000 \\
Surgical Knife & 69 & 6 & 1353 & 18588 & 0.932 & 0.920 & 0.049 & 0.092 \\
Surgicel & 464 & 14 & 275 & 19263 & 0.986 & 0.971 & 0.628 & 0.763 \\
Through cutting forceps & 0 & 1 & 542 & 19473 & 0.973 & 0.000 & 0.000 & 0.000 \\
Tisseel Applicator & 0 & 0 & 0 & 20016 & 1.000 & 0.000 & 0.000 & 0.000 \\
Tissue shaver & 2 & 4 & 55 & 19955 & 0.997 & 0.333 & 0.035 & 0.064 \\
unknown & 0 & 26 & 23 & 19967 & 0.998 & 0.000 & 0.000 & 0.000 \\
\bottomrule
\end{tabular}
\end{table}

Table~\ref{tab:s16_roc_auc} shows per-tool ROC-AUC and AUPRC.

\begin{table}[H]
\centering
\caption{Per-tool ROC-AUC and AUPRC for Gemma 3 27B with LoRA fine-tuning and classification head.}
\label{tab:s16_roc_auc}
\small
\begin{tabular}{lrr}
\toprule
Tool & ROC-AUC & AUPRC \\
\midrule
Suction Coagulator & 1.000 & 0.995 \\
Drill & 0.983 & 0.935 \\
Straight Forceps & 0.943 & 0.351 \\
Curette & 0.940 & 0.600 \\
Surgical Knife & 0.939 & 0.578 \\
Cotton Patty & 0.929 & 0.699 \\
Rongeur & 0.920 & 0.729 \\
Rhoton Dissector & 0.920 & 0.540 \\
Backbiting rongeur & 0.918 & 0.482 \\
Surgicel & 0.908 & 0.701 \\
Floseal Gel & 0.878 & 0.351 \\
Straight Curette & 0.841 & 0.008 \\
Suction & 0.819 & 0.847 \\
Through cutting forceps & 0.797 & 0.092 \\
Irrigation & 0.776 & 0.117 \\
Scissor & 0.766 & 0.259 \\
Tissue shaver & 0.737 & 0.099 \\
Bipolar Forceps & 0.723 & 0.043 \\
Grasper & 0.708 & 0.047 \\
Aspirating dissector & 0.655 & 0.158 \\
unknown & 0.618 & 0.001 \\
Floseal Applicator & 0.494 & 0.002 \\
Suction microdebrider & 0.298 & 0.016 \\
\midrule
Macro Average & 0.805 & 0.376 \\
\bottomrule
\end{tabular}
\end{table}

The model achieves high ROC-AUC ($>0.9$) for tools well-represented in training (Suction Coagulator, Drill, Straight Forceps, Curette, Surgical Knife, Cotton Patty, Rongeur, Rhoton Dissector, Backbiting rongeur, Surgicel), but lower values for tools with limited training data or those appearing predominantly in validation procedures.

Our per-tool evaluation metrics (ROC-AUC and AUPRC) are reported for the 23 surgical instruments that appear in the validation set.

\section{Zero-Shot Output Validation Failure Examples}
\label{sec:appendix_validation_failures}

Output validation failures in zero-shot evaluation are not merely JSON formatting issues. Table~\ref{tab:validation_failures} shows representative failed outputs from Qwen2-VL-2B-Instruct, the model with the highest output validation failure rate (41.7\%). In most cases, the model produces syntactically valid JSON but hallucinates tool names that do not exist in the provided list, such as ``Stirrup Curtain,'' ``Parallel Shears,'' ``Microlaryngeal electrodes,'' and ``Semitendinosus skin dissection.'' These hallucinated names are not surgical instruments and indicate a fundamental failure in visual recognition, not a formatting limitation.

\begin{table}[H]
\centering
\caption{Representative output validation failure examples from Qwen2-VL-2B-Instruct zero-shot evaluation. Each row shows the model's raw output and the failure category. Outputs are from randomly sampled frames (seed=42).}
\label{tab:validation_failures}
\small
\begin{tabular}{p{7.5cm} p{5.5cm}}
\toprule
Model Output & Failure Category \\
\midrule
\texttt{\{"detected\_tools": ["SCissor"]\}} & Ontology: misspelled tool name (``SCissor'' vs.\ ``Scissor'') \\
\addlinespace
\texttt{["Curette", "Floseal Gel", "Straight Curette", "Scissors", "Single- Precision Scissors", "Suction microdebrider"]} & JSON + Ontology: missing JSON wrapper; ``Scissors'' and ``Single- Precision Scissors'' not in valid list \\
\addlinespace
\texttt{\{"detected\_tools": ["Backbiting Rongeur", "Rhoton Dissector"]\}} & Ontology: capitalization mismatch (``Rongeur'' vs.\ ``rongeur'') \\
\addlinespace
\texttt{\{"detected\_tools": ["Collagen Matrix", "Curette", "Tourniquet"]\}} & Ontology: hallucinated tool (``Tourniquet'') \\
\addlinespace
\texttt{\{"detected\_tools": ["Local Anesthesia Needle"]]} & JSON: malformed JSON (extra closing bracket) \\
\addlinespace
\texttt{\{"detected\_tools": ["Aspirating dissector", "Froseal Applicator", "Froseal Gel", "Grasper", "Intraoperative Hemostat", "Rhoton Dissector"]\}} & Ontology: misspelled tools (``Froseal'') and hallucinated tool (``Intraoperative Hemostat'') \\
\addlinespace
\texttt{\{"detected\_tools": ["Curette", "Rhoton Dissector", "Stirrup Curtain", "Parallel Shears"]\}} & Ontology: hallucinated tools (``Stirrup Curtain,'' ``Parallel Shears'') \\
\addlinespace
\texttt{\{"detected\_tools": ["Microlaryngeal electrodes", "Laryngeal transducers", "Semitendinosus skin dissection", "Subglottic surgery", "Endoscopic cautery", "Suture", "Hydrofracture"]\}} & Ontology: all 7 predicted tools are hallucinated; none are surgical instruments \\
\bottomrule
\end{tabular}
\end{table}

\section{YOLOv12-m Training Configuration}
\label{sec:appendix_yolo_config}

Table~\ref{tab:yolo_config} shows the training configuration for YOLOv12-m used in Section~\ref{subsec:yolo_performance}. All hyperparameters use YOLO default values; no hyperparameter search was performed.

\begin{table}[H]
\centering
\caption{YOLOv12-m training configuration (300-epoch run with best exact match set accuracy).}
\label{tab:yolo_config}
\small
\begin{tabular}{ll}
\toprule
Parameter & Value \\
\midrule
Model & YOLOv12-m (26M parameters) \\
Pretrained weights & \texttt{yolo12m.pt} (COCO) \\
Epochs & 300 \\
Batch size & 32 total (4/GPU $\times$ 8 GPUs) \\
Image size & 1280$\times$1280 \\
Hardware & 8$\times$L40S GPUs \\
Training time & 11.3 wall-clock hours (90.6 GPU-hours) \\
Learning rate schedule & Cosine \\
Mixed precision & AMP (automatic) \\
Early stopping patience & 20 epochs \\
Mosaic close epoch & 10 \\
Data caching & Disabled (disk-based) \\
Confidence threshold (eval) & 0.25 \\
Random seed & 42 \\
\bottomrule
\end{tabular}
\end{table}

\section{Per-Tool Metrics for CholecT50 Evaluation}
\label{sec:appendix_cholect50}

Tables~\ref{tab:cholect50_zeroshot_metrics}--\ref{tab:cholect50_yolo_roc_auc} show per-tool classification metrics on the CholecT50 validation set ($n=19{,}923$ frames) for zero-shot Gemma 3 27B, fine-tuned Gemma 3 27B, and YOLOv12-m.

\begin{table}[H]
\centering
\caption{Per-tool classification metrics for Gemma 3 27B \textbf{zero-shot} on CholecT50. TP = true positives, FP = false positives, FN = false negatives, TN = true negatives.}
\label{tab:cholect50_zeroshot_metrics}
\small
\begin{tabular}{lrrrrrrrr}
\toprule
Tool & TP & FP & FN & TN & Accuracy & Precision & Recall & F1 \\
\midrule
grasper & 6978 & 2740 & 5552 & 4653 & 0.584 & 0.718 & 0.557 & 0.627 \\
hook & 2092 & 883 & 8651 & 8297 & 0.522 & 0.703 & 0.195 & 0.305 \\
bipolar & 838 & 12096 & 246 & 6743 & 0.381 & 0.065 & 0.773 & 0.120 \\
irrigator & 247 & 4355 & 696 & 14625 & 0.747 & 0.054 & 0.262 & 0.089 \\
clipper & 59 & 686 & 627 & 18551 & 0.934 & 0.079 & 0.086 & 0.083 \\
scissors & 163 & 3993 & 313 & 15454 & 0.784 & 0.039 & 0.342 & 0.070 \\
\bottomrule
\end{tabular}
\end{table}

In the zero-shot setting, bipolar has 12,096 false positives, irrigator has 4,355, and scissors has 3,993. Hook has a recall of 0.195. Grasper achieves the highest F1 (0.627).

\begin{table}[H]
\centering
\caption{Per-tool classification metrics for Gemma 3 27B \textbf{fine-tuned} (LoRA + classification head) on CholecT50. TP = true positives, FP = false positives, FN = false negatives, TN = true negatives.}
\label{tab:cholect50_finetuned_metrics}
\small
\begin{tabular}{lrrrrrrrr}
\toprule
Tool & TP & FP & FN & TN & Accuracy & Precision & Recall & F1 \\
\midrule
grasper & 11614 & 1307 & 916 & 6086 & 0.888 & 0.899 & 0.927 & 0.913 \\
hook & 10491 & 306 & 252 & 8874 & 0.972 & 0.972 & 0.977 & 0.974 \\
bipolar & 805 & 48 & 279 & 18791 & 0.984 & 0.944 & 0.743 & 0.831 \\
irrigator & 663 & 51 & 280 & 18929 & 0.983 & 0.929 & 0.703 & 0.800 \\
clipper & 614 & 42 & 72 & 19195 & 0.994 & 0.936 & 0.895 & 0.915 \\
scissors & 285 & 36 & 191 & 19411 & 0.989 & 0.888 & 0.599 & 0.715 \\
\bottomrule
\end{tabular}
\end{table}

After fine-tuning, the model achieves a macro ROC-AUC of 0.966 and macro AUPRC of 0.883. Table~\ref{tab:cholect50_roc_auc} shows per-tool ROC-AUC and AUPRC.

\begin{table}[H]
\centering
\caption{Per-tool ROC-AUC and AUPRC for Gemma 3 27B fine-tuned on CholecT50. Sorted by ROC-AUC in descending order.}
\label{tab:cholect50_roc_auc}
\small
\begin{tabular}{lrr}
\toprule
Tool & ROC-AUC & AUPRC \\
\midrule
hook & 0.989 & 0.985 \\
clipper & 0.989 & 0.909 \\
irrigator & 0.969 & 0.836 \\
grasper & 0.941 & 0.960 \\
bipolar & 0.959 & 0.856 \\
scissors & 0.947 & 0.753 \\
\midrule
Macro Average & 0.966 & 0.883 \\
\bottomrule
\end{tabular}
\end{table}

After fine-tuning, hook achieves the highest F1 (0.974), and all tools achieve F1 $>$ 0.7. The largest change from zero-shot to fine-tuned is for bipolar (F1: 0.120 $\to$ 0.831) and irrigator (F1: 0.089 $\to$ 0.800).

\begin{table}[H]
\centering
\caption{Per-tool classification metrics for \textbf{YOLOv12-m} on CholecT50.}
\label{tab:cholect50_yolo_metrics}
\small
\begin{tabular}{lrrrrrrrr}
\toprule
Tool & TP & FP & FN & TN & Accuracy & Precision & Recall & F1 \\
\midrule
grasper & 11938 & 1945 & 592 & 5448 & 0.873 & 0.860 & 0.953 & 0.904 \\
hook & 10590 & 517 & 153 & 8663 & 0.966 & 0.953 & 0.986 & 0.969 \\
bipolar & 841 & 73 & 243 & 18766 & 0.984 & 0.920 & 0.776 & 0.842 \\
irrigator & 772 & 76 & 171 & 18904 & 0.988 & 0.910 & 0.819 & 0.862 \\
clipper & 624 & 68 & 62 & 19169 & 0.993 & 0.902 & 0.910 & 0.906 \\
scissors & 275 & 36 & 201 & 19411 & 0.988 & 0.884 & 0.578 & 0.699 \\
\bottomrule
\end{tabular}
\end{table}

For YOLOv12-m, scissors has the lowest recall and F1 among all tools (476 validation instances). YOLOv12-m achieves higher F1 than Gemma on irrigator (0.862 vs.\ 0.800), while Gemma achieves higher F1 on grasper (0.913 vs.\ 0.904).

\begin{table}[H]
\centering
\caption{Per-tool ROC-AUC for YOLOv12-m on CholecT50 (using maximum detection confidence per class as the continuous score).}
\label{tab:cholect50_yolo_roc_auc}
\small
\begin{tabular}{lr}
\toprule
Tool & ROC-AUC \\
\midrule
hook & 0.992 \\
clipper & 0.967 \\
grasper & 0.958 \\
irrigator & 0.955 \\
bipolar & 0.920 \\
scissors & 0.825 \\
\midrule
Macro Average & 0.936 \\
\bottomrule
\end{tabular}
\end{table}

\section{Per-Tool Metrics for PitVis-2023 Evaluation}
\label{sec:appendix_pitvis}

Tables~\ref{tab:pitvis_zeroshot_metrics}--\ref{tab:pitvis_yolo_roc_auc} show per-tool classification metrics on the PitVis-2023 validation set ($n=30{,}896$ frames) for zero-shot Gemma 3 27B, fine-tuned Gemma 3 27B (LoRA + classification head), and YOLOv12-m.

\begin{table}[H]
\centering
\caption{Per-tool classification metrics for Gemma 3 27B \textbf{zero-shot} on PitVis-2023. TP = true positives, FP = false positives, FN = false negatives, TN = true negatives. Sorted by ground truth count ($N$) in descending order.}
\label{tab:pitvis_zeroshot_metrics}
\footnotesize
\begin{tabular}{lrrrrrrrr}
\toprule
Tool & $N$ & TP & FP & FN & TN & Precision & Recall & F1 \\
\midrule
suction & 11971 & 9353 & 13487 & 2618 & 5438 & 0.410 & 0.781 & 0.537 \\
ring\_curette & 4314 & 375 & 416 & 3939 & 26166 & 0.474 & 0.087 & 0.147 \\
kerrisons & 3567 & 42 & 116 & 3525 & 27213 & 0.266 & 0.012 & 0.023 \\
pituitary\_rongeurs & 909 & 49 & 1196 & 860 & 28791 & 0.039 & 0.054 & 0.045 \\
spatula\_dissector & 412 & 0 & 56 & 412 & 30428 & 0.000 & 0.000 & 0.000 \\
nasal\_cutting\_forceps & 374 & 0 & 0 & 374 & 30522 & 0.000 & 0.000 & 0.000 \\
stealth\_pointer & 353 & 0 & 0 & 353 & 30543 & 0.000 & 0.000 & 0.000 \\
irrigation\_syringe & 286 & 6 & 134 & 280 & 30476 & 0.043 & 0.021 & 0.028 \\
micro\_doppler & 251 & 0 & 15 & 251 & 30630 & 0.000 & 0.000 & 0.000 \\
cup\_forceps & 241 & 3 & 230 & 238 & 30425 & 0.013 & 0.012 & 0.013 \\
freer\_elevator & 226 & 1 & 267 & 225 & 30403 & 0.004 & 0.004 & 0.004 \\
haemostatic\_foam & 179 & 12 & 182 & 167 & 30535 & 0.062 & 0.067 & 0.064 \\
retractable\_knife & 136 & 1 & 339 & 135 & 30421 & 0.003 & 0.007 & 0.004 \\
cottle & 130 & 0 & 0 & 130 & 30766 & 0.000 & 0.000 & 0.000 \\
dural\_scissors & 129 & 0 & 42 & 129 & 30725 & 0.000 & 0.000 & 0.000 \\
surgical\_drill & 80 & 0 & 18 & 80 & 30798 & 0.000 & 0.000 & 0.000 \\
tissue\_glue & 63 & 0 & 1 & 63 & 30832 & 0.000 & 0.000 & 0.000 \\
bipolar\_forceps & 49 & 19 & 5380 & 30 & 25467 & 0.004 & 0.388 & 0.007 \\
\bottomrule
\end{tabular}
\end{table}

In the zero-shot setting, only suction achieves an F1 above 0.1, with severe over-prediction (5{,}380 false positives for bipolar\_forceps and 13{,}487 for suction). For 11 of 18 tools, the model produces zero correct detections.

\begin{table}[H]
\centering
\caption{Per-tool classification metrics for Gemma 3 27B \textbf{fine-tuned} (LoRA + classification head) on PitVis-2023. Sorted by ground truth count ($N$) in descending order.}
\label{tab:pitvis_finetuned_metrics}
\footnotesize
\begin{tabular}{lrrrrrrrr}
\toprule
Tool & $N$ & TP & FP & FN & TN & Precision & Recall & F1 \\
\midrule
suction & 11971 & 11473 & 1484 & 498 & 17441 & 0.886 & 0.958 & 0.921 \\
ring\_curette & 4314 & 3671 & 131 & 643 & 26451 & 0.966 & 0.851 & 0.905 \\
kerrisons & 3567 & 2669 & 325 & 898 & 27004 & 0.891 & 0.748 & 0.814 \\
pituitary\_rongeurs & 909 & 376 & 69 & 533 & 29918 & 0.845 & 0.414 & 0.555 \\
spatula\_dissector & 412 & 115 & 45 & 297 & 30439 & 0.719 & 0.279 & 0.402 \\
nasal\_cutting\_forceps & 374 & 163 & 67 & 211 & 30455 & 0.709 & 0.436 & 0.540 \\
stealth\_pointer & 353 & 183 & 32 & 170 & 30511 & 0.851 & 0.518 & 0.644 \\
irrigation\_syringe & 286 & 169 & 42 & 117 & 30568 & 0.801 & 0.591 & 0.680 \\
micro\_doppler & 251 & 187 & 6 & 64 & 30639 & 0.969 & 0.745 & 0.842 \\
cup\_forceps & 241 & 62 & 62 & 179 & 30593 & 0.500 & 0.257 & 0.340 \\
freer\_elevator & 226 & 124 & 70 & 102 & 30600 & 0.639 & 0.549 & 0.591 \\
haemostatic\_foam & 179 & 129 & 15 & 50 & 30702 & 0.896 & 0.721 & 0.799 \\
retractable\_knife & 136 & 32 & 8 & 104 & 30752 & 0.800 & 0.235 & 0.364 \\
cottle & 130 & 46 & 21 & 84 & 30745 & 0.687 & 0.354 & 0.467 \\
dural\_scissors & 129 & 14 & 17 & 115 & 30750 & 0.452 & 0.109 & 0.175 \\
surgical\_drill & 80 & 61 & 0 & 19 & 30816 & 1.000 & 0.763 & 0.865 \\
tissue\_glue & 63 & 52 & 10 & 11 & 30823 & 0.839 & 0.825 & 0.832 \\
bipolar\_forceps & 49 & 4 & 0 & 45 & 30847 & 1.000 & 0.082 & 0.151 \\
\bottomrule
\end{tabular}
\end{table}

After fine-tuning, the model achieves macro-averaged ROC-AUC of 0.966 and macro AUPRC of 0.691. Table~\ref{tab:pitvis_roc_auc} shows per-tool ROC-AUC and AUPRC.

\begin{table}[H]
\centering
\caption{Per-tool ROC-AUC and AUPRC for Gemma 3 27B fine-tuned on PitVis-2023. Sorted by ROC-AUC in descending order.}
\label{tab:pitvis_roc_auc}
\small
\begin{tabular}{lrr}
\toprule
Tool & ROC-AUC & AUPRC \\
\midrule
bipolar\_forceps & 0.999 & 0.756 \\
cottle & 0.994 & 0.662 \\
ring\_curette & 0.988 & 0.965 \\
surgical\_drill & 0.991 & 0.917 \\
micro\_doppler & 0.991 & 0.903 \\
dural\_scissors & 0.984 & 0.336 \\
tissue\_glue & 0.998 & 0.829 \\
suction & 0.974 & 0.942 \\
freer\_elevator & 0.968 & 0.599 \\
spatula\_dissector & 0.965 & 0.493 \\
haemostatic\_foam & 0.963 & 0.811 \\
irrigation\_syringe & 0.964 & 0.707 \\
stealth\_pointer & 0.963 & 0.676 \\
cup\_forceps & 0.941 & 0.345 \\
kerrisons & 0.939 & 0.846 \\
nasal\_cutting\_forceps & 0.924 & 0.544 \\
pituitary\_rongeurs & 0.921 & 0.626 \\
retractable\_knife & 0.917 & 0.483 \\
\midrule
Macro Average & 0.966 & 0.691 \\
\bottomrule
\end{tabular}
\end{table}

\begin{table}[H]
\centering
\caption{Per-tool classification metrics for \textbf{YOLOv12-m} on PitVis-2023. Sorted by ground truth count ($N$) in descending order.}
\label{tab:pitvis_yolo_metrics}
\footnotesize
\begin{tabular}{lrrrrrrrr}
\toprule
Tool & $N$ & TP & FP & FN & TN & Precision & Recall & F1 \\
\midrule
suction & 11971 & 11630 & 2250 & 341 & 16675 & 0.838 & 0.972 & 0.900 \\
ring\_curette & 4314 & 3364 & 88 & 950 & 26494 & 0.975 & 0.780 & 0.866 \\
kerrisons & 3567 & 2748 & 387 & 819 & 26942 & 0.877 & 0.770 & 0.820 \\
pituitary\_rongeurs & 909 & 457 & 230 & 452 & 29757 & 0.665 & 0.503 & 0.573 \\
spatula\_dissector & 412 & 163 & 183 & 249 & 30301 & 0.471 & 0.396 & 0.430 \\
nasal\_cutting\_forceps & 374 & 228 & 174 & 146 & 30348 & 0.567 & 0.610 & 0.588 \\
stealth\_pointer & 353 & 179 & 36 & 174 & 30507 & 0.833 & 0.507 & 0.630 \\
irrigation\_syringe & 286 & 171 & 183 & 115 & 30427 & 0.483 & 0.598 & 0.534 \\
micro\_doppler & 251 & 204 & 7 & 47 & 30638 & 0.967 & 0.813 & 0.883 \\
cup\_forceps & 241 & 171 & 775 & 70 & 29880 & 0.181 & 0.710 & 0.288 \\
freer\_elevator & 226 & 154 & 79 & 72 & 30591 & 0.661 & 0.681 & 0.671 \\
haemostatic\_foam & 179 & 130 & 14 & 49 & 30703 & 0.903 & 0.726 & 0.805 \\
retractable\_knife & 136 & 66 & 29 & 70 & 30731 & 0.695 & 0.485 & 0.571 \\
cottle & 130 & 112 & 83 & 18 & 30683 & 0.574 & 0.862 & 0.689 \\
dural\_scissors & 129 & 38 & 26 & 91 & 30741 & 0.594 & 0.295 & 0.394 \\
surgical\_drill & 80 & 43 & 5 & 37 & 30811 & 0.896 & 0.538 & 0.672 \\
tissue\_glue & 63 & 60 & 22 & 3 & 30811 & 0.732 & 0.952 & 0.828 \\
bipolar\_forceps & 49 & 16 & 3 & 33 & 30844 & 0.842 & 0.327 & 0.471 \\
\bottomrule
\end{tabular}
\end{table}

\begin{table}[H]
\centering
\caption{Per-tool ROC-AUC for YOLOv12-m on PitVis-2023 (using maximum detection confidence per class as the continuous score). Sorted by ROC-AUC in descending order.}
\label{tab:pitvis_yolo_roc_auc}
\small
\begin{tabular}{lr}
\toprule
Tool & ROC-AUC \\
\midrule
tissue\_glue & 0.984 \\
suction & 0.979 \\
cottle & 0.945 \\
kerrisons & 0.943 \\
ring\_curette & 0.938 \\
micro\_doppler & 0.920 \\
cup\_forceps & 0.908 \\
stealth\_pointer & 0.898 \\
irrigation\_syringe & 0.892 \\
haemostatic\_foam & 0.880 \\
freer\_elevator & 0.852 \\
pituitary\_rongeurs & 0.816 \\
nasal\_cutting\_forceps & 0.806 \\
retractable\_knife & 0.794 \\
surgical\_drill & 0.775 \\
spatula\_dissector & 0.706 \\
bipolar\_forceps & 0.663 \\
dural\_scissors & 0.647 \\
\midrule
Macro Average & 0.853 \\
\bottomrule
\end{tabular}
\end{table}

\section{Per-Tool Metrics for SurgVU Evaluation}
\label{sec:appendix_surgvu}

Tables~\ref{tab:surgvu_zeroshot_metrics}--\ref{tab:surgvu_yolo_roc_auc} show per-tool classification metrics on the SurgVU validation set ($n=18{,}919$ frames) for zero-shot Gemma 3 27B, fine-tuned Gemma 3 27B (LoRA + classification head), and YOLOv12-m. Tools with zero ground truth instances in the validation set (bipolar dissector, potts scissors, tenaculum forceps) are omitted from per-tool tables but contribute to macro averages where applicable.

\begin{table}[H]
\centering
\caption{Per-tool classification metrics for Gemma 3 27B \textbf{zero-shot} on SurgVU. Sorted by ground truth count ($N$) in descending order.}
\label{tab:surgvu_zeroshot_metrics}
\footnotesize
\begin{tabular}{lrrrrrrrr}
\toprule
Tool & $N$ & TP & FP & FN & TN & Precision & Recall & F1 \\
\midrule
needle driver & 7946 & 3540 & 3022 & 4406 & 7951 & 0.540 & 0.446 & 0.488 \\
cadiere forceps & 7928 & 53 & 22 & 7875 & 10969 & 0.707 & 0.007 & 0.013 \\
bipolar forceps & 6980 & 2624 & 5623 & 4356 & 6316 & 0.318 & 0.376 & 0.345 \\
monopolar curved scissors & 6906 & 229 & 447 & 6677 & 11566 & 0.339 & 0.033 & 0.060 \\
prograsp forceps & 3341 & 1128 & 3869 & 2213 & 11709 & 0.226 & 0.338 & 0.271 \\
grasping retractor & 3096 & 46 & 232 & 3050 & 15591 & 0.166 & 0.015 & 0.027 \\
force bipolar & 2091 & 699 & 5049 & 1392 & 11779 & 0.122 & 0.334 & 0.178 \\
permanent cautery hook/spatula & 1275 & 27 & 119 & 1248 & 17525 & 0.185 & 0.021 & 0.038 \\
clip applier & 967 & 45 & 680 & 922 & 17272 & 0.062 & 0.047 & 0.053 \\
vessel sealer & 984 & 224 & 3135 & 760 & 14800 & 0.067 & 0.228 & 0.103 \\
stapler & 444 & 114 & 1079 & 330 & 17396 & 0.096 & 0.257 & 0.139 \\
tip-up fenestrated grasper & 352 & 11 & 508 & 341 & 18059 & 0.021 & 0.031 & 0.025 \\
suction irrigator & 62 & 16 & 3999 & 46 & 14858 & 0.004 & 0.258 & 0.008 \\
synchroseal & 10 & 0 & 1 & 10 & 18908 & 0.000 & 0.000 & 0.000 \\
\bottomrule
\end{tabular}
\end{table}

In the zero-shot setting, no tool exceeds F1 of 0.5; the model over-predicts bipolar forceps (5{,}623 FP), prograsp forceps (3{,}869 FP), force bipolar (5{,}049 FP), and vessel sealer (3{,}135 FP), resulting in low precision across the board. Even cadiere forceps---present in 42\% of validation frames---is detected only 0.7\% of the time.

\begin{table}[H]
\centering
\caption{Per-tool classification metrics for Gemma 3 27B \textbf{fine-tuned} (LoRA + classification head) on SurgVU. Sorted by ground truth count ($N$) in descending order.}
\label{tab:surgvu_finetuned_metrics}
\footnotesize
\begin{tabular}{lrrrrrrrr}
\toprule
Tool & $N$ & TP & FP & FN & TN & Precision & Recall & F1 \\
\midrule
needle driver & 7946 & 5840 & 1162 & 2106 & 9811 & 0.834 & 0.735 & 0.781 \\
cadiere forceps & 7928 & 6842 & 1626 & 1086 & 9365 & 0.808 & 0.863 & 0.835 \\
bipolar forceps & 6980 & 6031 & 933 & 949 & 11006 & 0.866 & 0.864 & 0.865 \\
monopolar curved scissors & 6906 & 4945 & 465 & 1961 & 11548 & 0.914 & 0.716 & 0.803 \\
prograsp forceps & 3341 & 1681 & 738 & 1660 & 14840 & 0.695 & 0.503 & 0.584 \\
grasping retractor & 3096 & 2077 & 816 & 1019 & 15007 & 0.718 & 0.671 & 0.694 \\
force bipolar & 2091 & 320 & 115 & 1771 & 16713 & 0.736 & 0.153 & 0.253 \\
permanent cautery hook/spatula & 1275 & 323 & 100 & 952 & 17544 & 0.764 & 0.253 & 0.380 \\
vessel sealer & 984 & 705 & 288 & 279 & 17647 & 0.710 & 0.716 & 0.713 \\
clip applier & 967 & 89 & 44 & 878 & 17908 & 0.669 & 0.092 & 0.162 \\
stapler & 444 & 289 & 35 & 155 & 18440 & 0.892 & 0.651 & 0.753 \\
tip-up fenestrated grasper & 352 & 0 & 0 & 352 & 18567 & 0.000 & 0.000 & 0.000 \\
suction irrigator & 62 & 0 & 0 & 62 & 18857 & 0.000 & 0.000 & 0.000 \\
synchroseal & 10 & 0 & 49 & 10 & 18860 & 0.000 & 0.000 & 0.000 \\
\bottomrule
\end{tabular}
\end{table}

After fine-tuning, the model achieves macro-averaged ROC-AUC of 0.740 and macro AUPRC of 0.502 across the 14 tool classes present in the validation set. Table~\ref{tab:surgvu_roc_auc} shows per-tool ROC-AUC and AUPRC.

\begin{table}[H]
\centering
\caption{Per-tool ROC-AUC and AUPRC for Gemma 3 27B fine-tuned on SurgVU. Sorted by ROC-AUC in descending order.}
\label{tab:surgvu_roc_auc}
\small
\begin{tabular}{lrr}
\toprule
Tool & ROC-AUC & AUPRC \\
\midrule
stapler & 0.955 & 0.781 \\
vessel sealer & 0.932 & 0.671 \\
bipolar forceps & 0.928 & 0.885 \\
cadiere forceps & 0.912 & 0.876 \\
grasping retractor & 0.897 & 0.660 \\
monopolar curved scissors & 0.881 & 0.851 \\
needle driver & 0.859 & 0.867 \\
prograsp forceps & 0.738 & 0.606 \\
force bipolar & 0.735 & 0.352 \\
permanent cautery hook/spatula & 0.665 & 0.286 \\
clip applier & 0.609 & 0.151 \\
suction irrigator & 0.605 & 0.040 \\
tip-up fenestrated grasper & 0.363 & 0.009 \\
synchroseal & 0.282 & 0.000 \\
\midrule
Macro Average & 0.740 & 0.502 \\
\bottomrule
\end{tabular}
\end{table}

\begin{table}[H]
\centering
\caption{Per-tool classification metrics for \textbf{YOLOv12-m} on SurgVU. Sorted by ground truth count ($N$) in descending order.}
\label{tab:surgvu_yolo_metrics}
\footnotesize
\begin{tabular}{lrrrrrrrr}
\toprule
Tool & $N$ & TP & FP & FN & TN & Precision & Recall & F1 \\
\midrule
needle driver & 7946 & 5928 & 1087 & 2018 & 9886 & 0.845 & 0.746 & 0.792 \\
cadiere forceps & 7928 & 7397 & 1884 & 531 & 9107 & 0.797 & 0.933 & 0.860 \\
bipolar forceps & 6980 & 6205 & 1390 & 775 & 10549 & 0.817 & 0.889 & 0.851 \\
monopolar curved scissors & 6906 & 5104 & 696 & 1802 & 11317 & 0.880 & 0.739 & 0.803 \\
prograsp forceps & 3341 & 1804 & 810 & 1537 & 14768 & 0.690 & 0.540 & 0.606 \\
grasping retractor & 3096 & 2226 & 742 & 870 & 15081 & 0.750 & 0.719 & 0.734 \\
force bipolar & 2091 & 397 & 137 & 1694 & 16691 & 0.743 & 0.190 & 0.302 \\
permanent cautery hook/spatula & 1275 & 293 & 255 & 982 & 17389 & 0.535 & 0.230 & 0.321 \\
vessel sealer & 984 & 703 & 644 & 281 & 17291 & 0.522 & 0.714 & 0.603 \\
clip applier & 967 & 109 & 120 & 858 & 17832 & 0.476 & 0.113 & 0.182 \\
stapler & 444 & 321 & 88 & 123 & 18387 & 0.785 & 0.723 & 0.753 \\
tip-up fenestrated grasper & 352 & 0 & 0 & 352 & 18567 & 0.000 & 0.000 & 0.000 \\
suction irrigator & 62 & 0 & 0 & 62 & 18857 & 0.000 & 0.000 & 0.000 \\
synchroseal & 10 & 0 & 458 & 10 & 18451 & 0.000 & 0.000 & 0.000 \\
\bottomrule
\end{tabular}
\end{table}

\begin{table}[H]
\centering
\caption{Per-tool ROC-AUC for YOLOv12-m on SurgVU (using maximum detection confidence per class as the continuous score). Sorted by ROC-AUC in descending order.}
\label{tab:surgvu_yolo_roc_auc}
\small
\begin{tabular}{lr}
\toprule
Tool & ROC-AUC \\
\midrule
cadiere forceps & 0.937 \\
bipolar forceps & 0.934 \\
stapler & 0.888 \\
vessel sealer & 0.875 \\
monopolar curved scissors & 0.873 \\
needle driver & 0.867 \\
grasping retractor & 0.840 \\
prograsp forceps & 0.763 \\
permanent cautery hook/spatula & 0.611 \\
clip applier & 0.588 \\
force bipolar & 0.582 \\
suction irrigator & 0.500 \\
tip-up fenestrated grasper & 0.500 \\
synchroseal & 0.477 \\
\midrule
Macro Average & 0.731 \\
\bottomrule
\end{tabular}
\end{table}

\input{sections/appendix_new_tasks}

\section{Robustness Check: CNN without Bounding Box Supervision}
\label{sec:appendix_resnet}

Our comparison between YOLOv12-m and VLM-based classifiers evaluates tool presence only: both models are scored on whether the predicted tool set matches the ground truth set, ignoring spatial localization. However, the two approaches differ in their training signal. YOLO is trained with bounding box supervision, while VLMs receive only set-level labels during fine-tuning. YOLO's localization objective may confer an indirect advantage for presence detection by forcing the model to ground each prediction spatially, reducing hallucinated detections. Conversely, VLMs must learn tool presence from a weaker supervisory signal. This asymmetry could mean that YOLO's advantage partly reflects the richer information content of bounding box annotations rather than a fundamental architectural superiority for the presence detection task.

To test this, we train a ResNet-50 (23.6M parameters; \citealt{he2016deep}) for multi-label tool classification using \textit{only} set-level labels---the same supervisory signal available to VLMs---with no bounding box information. The model uses ImageNet-pretrained weights, a dropout layer ($p=0.5$) followed by a 31-class linear head, and is trained with binary cross-entropy loss (label smoothing $\epsilon=0.1$). We use differential learning rates (backbone: $10^{-4}$, head: $10^{-3}$), AdamW optimizer (weight decay $10^{-2}$), cosine annealing schedule, and aggressive data augmentation (random resized crops, color jitter, random erasing, rotation). Training uses 8$\times$L40S GPUs for 50 epochs with a total batch size of 512.

\begin{figure}[H]
\centering
\includegraphics[width=\textwidth]{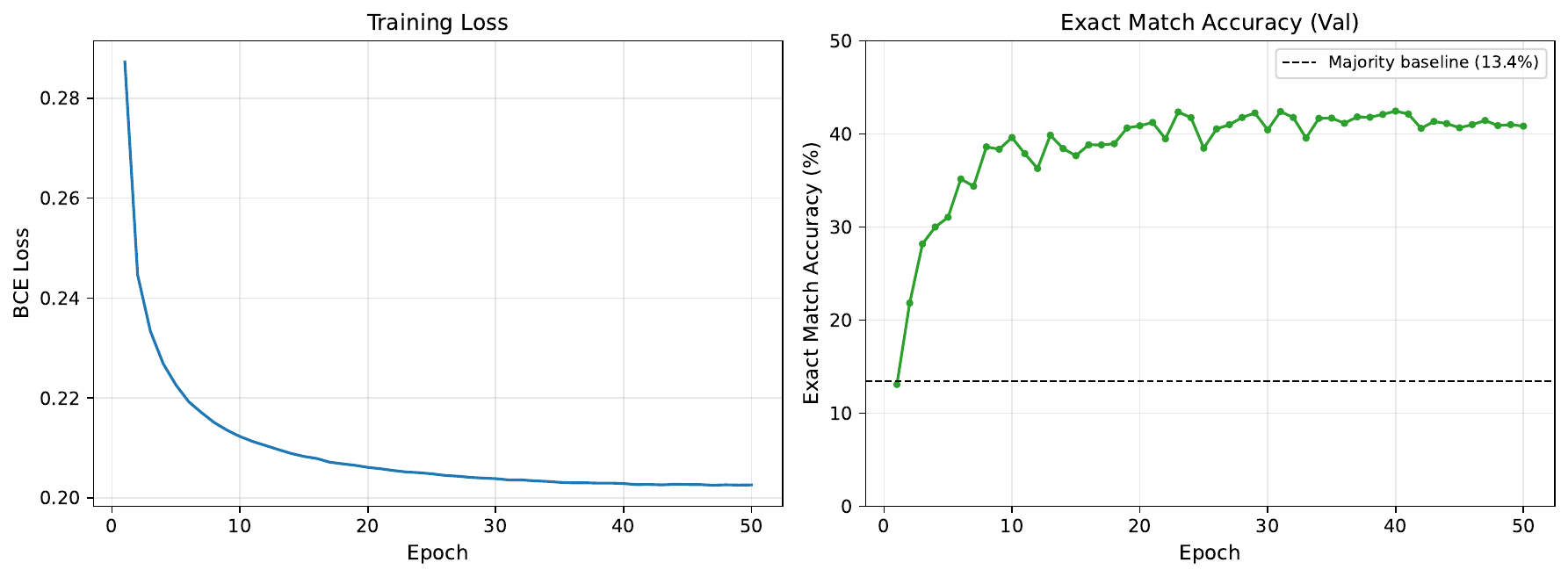}
\caption{Training dynamics for ResNet-50 multi-label classification without bounding box supervision. Left: Training loss (binary cross-entropy with label smoothing). Right: Exact match accuracy on the validation set evaluated at each epoch. The dashed line indicates the majority class baseline (13.4\%).}
\label{fig:s31_resnet_training}
\end{figure}

Figure~\ref{fig:s31_resnet_training} shows training dynamics. The model achieves 39.6\% exact match accuracy (95\% CI: 38.9\%--40.3\%) on the validation set ($n=20{,}016$), with 52.6\% Jaccard similarity, 70.3\% top-1 accuracy, and 0.673 micro F1.

\clearpage

This result exceeds all zero-shot VLMs and matches the 3-epoch LoRA rank sweep at $r=1024$ (39.6\%, Section~\ref{subsec:lora_rank_sweep}), but falls below the best fine-tuned VLM (Gemma~3 27B with LoRA classification head trained for 10 epochs: 51.08\%, Section~\ref{subsec:lora_classification_head}), despite using roughly 1{,}000$\times$ fewer parameters and receiving the same set-level supervision. The ResNet-50's performance also falls below YOLOv12-m (54.7\%), suggesting that bounding box supervision does confer some advantage for presence detection. The fact that a 23.6M-parameter CNN trained with set-level labels alone outperforms all zero-shot VLMs---including models with up to 235B parameters---underscores the difficulty of surgical tool detection as a zero-shot task.

\section{Effect of Sampling Temperature on Zero-Shot Accuracy}
\label{sec:appendix_temperature}

\begin{figure}[H]
\centering
\includegraphics[width=0.75\textwidth]{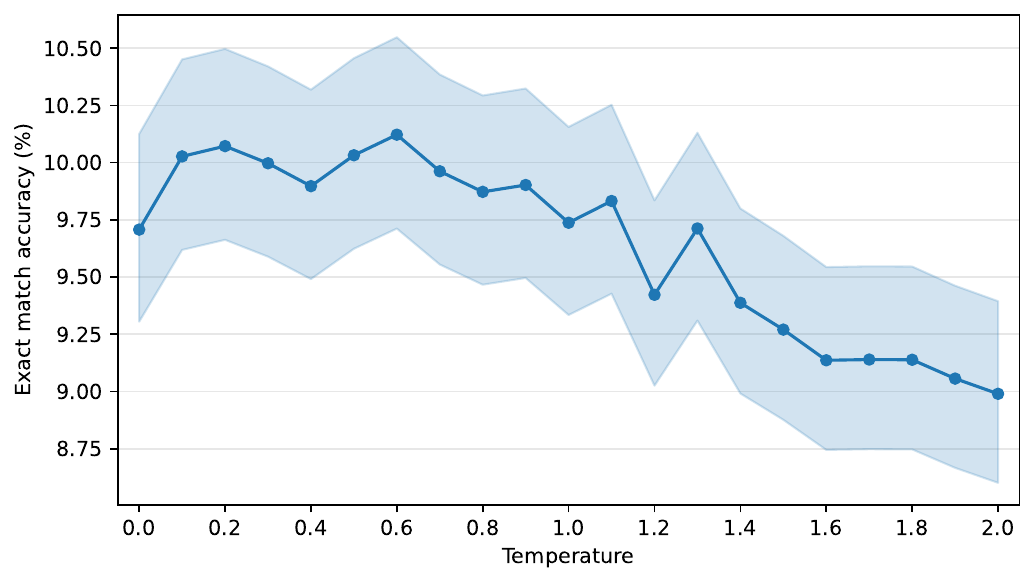}
\caption{Zero-shot exact match accuracy of Gemma~3 27B-it on the SDSC-EEA validation set ($n=20{,}016$) as a function of sampling temperature. Shaded region: 95\% Wilson binomial CI. Accuracy is largely insensitive to temperature in the 0--0.7 range (${\sim}10\%$), with a gradual decline at higher temperatures. At $T \geq 1.5$, output validation failures begin to appear (up to 38 at $T=2.0$), indicating that high temperatures degrade the model's ability to produce valid JSON. All CIs overlap, confirming that temperature has no statistically significant effect on zero-shot performance.}
\label{fig:accuracy_vs_temperature}
\end{figure}

To investigate whether sampling temperature affects zero-shot tool detection performance, we sweep the generation temperature of Gemma~3 27B-it from 0 (greedy decoding) to 2.0 in increments of 0.1, evaluating exact match accuracy on the full validation set ($n=20{,}016$) at each setting. Figure~\ref{fig:accuracy_vs_temperature} shows the results with 95\% Wilson binomial confidence intervals.

\clearpage

\input{sections/appendix_zero_shot_tables}

\section{Qwen3-VL-235B-A22B-Thinking's thinking}
\label{sec:appendix_qwen_thinking}

\begin{qwenanswer}
</think>
So, let's look at the surgical video frame. There are two tools visible. Let's check the list. The tools here seem to be a suction and maybe a grasper or something else. Wait, the left tool might be suction, and the right one... Let's see. The valid names: Suction is one. Then, the other tool--- maybe Grasper? Wait, in the image, the two instruments. Let's check each.

First, Suction: yes, one of the tools is likely suction. Then, the other tool--- Grasper? Let's confirm. The list has Grasper. So detected tools would be Suction and Grasper. Wait, let's make sure. The image shows two surgical instruments. Let's check the options. Suction is a valid name. Grasper is also there. So the detected tools are Suction and Grasper.
</think>

{"detected_tools": ["Suction", "Grasper"]} 
\end{qwenanswer}

\begin{qwenanswer}
</think>
So, let's look at the image. First, identify the tools. There's a white gauze-like material, which might be a Cotton Patty. Then there are two instruments: one looks like a suction tool (Suction), and another might be a Grasper. Wait, let's check the list.

Cotton Patty is in the list. Then, the metallic instrument with a tube--- Suction. Also, the other instrument that's holding or manipulating--- Grasper. Let's confirm.

In the image, the white pad is Cotton Patty. The tube-like instrument is Suction. The other tool with the jaws is Grasper. Are there any others? Let's check the valid names.

So detected tools: Cotton Patty, Suction, Grasper.
</think>

{"detected_tools": ["Cotton Patty", "Suction", "Grasper"]}
\end{qwenanswer}

\begin{qwenanswer}
</think>
So, let's look at the surgical video frame. There are two tools visible. First, the one on the left---what is that? Wait, the options include Suction, maybe? Wait, no. Wait, the tool on the left might be a Suction Coagulator? Or maybe Suction. Wait, let's check the list. The tools present: the left tool looks like a suction device, maybe Suction. The right tool---wait, the right one has a metallic tip, maybe a Suction Coagulator? Wait, no. Wait, let's think. The options include Suction. Let's see. In the image, there are two instruments. Let's check each.

First, the left instrument: it's a metallic tube, possibly a Suction tool. The right instrument: maybe a Suction Coagulator? Wait, no. Wait, the valid tools: Suction is a valid name. Let's check. The list has "Suction" as a tool. So if there's a suction device, that's one. Are there others?

Wait, the image shows two surgical instruments. Let's identify each. The left one: maybe Suction. The right one: maybe Suction Coagulator? Wait, no. Wait, the Suction Coagulator is a specific tool. Alternatively, maybe the right one is a Suction microdebrider? No, the microdebrider is different. Wait, let's check the options again.

Wait, the tools in the list: Suction is a valid entry. So if there's a suction device, that's one. Are there any other tools? Let's see. The image has two instruments. Let's check each.

First instrument (left): looks like a suction tube. So "Suction".

Second instrument (right): the tip is rounded, maybe a Suction Coagulator? Wait, Suction Coagulator is a tool that combines suction and coagulation. But maybe in this case, the right tool is a Suction Coagulator. Wait, but let's check the options. The valid names include "Suction Coagulator" and "Suction".

Wait, but maybe both are suction-related. Wait, no. Let's think again. The left tool: maybe it's a Suction. The right tool: perhaps a Suction Coagulator. But I need to be precise.

Alternatively, maybe the right tool is a Suction microdebrider? No, the microdebrider has a different tip. Wait, the image shows two instruments. Let's check the list again.

Wait, the correct approach is to match the visible tools to the exact names. Let's see:

- Suction: yes, if there's a suction device.

Looking at the image, there are two instruments. Let's check each.

First, the left instrument: it's a metallic tube, likely a suction device. So "Suction".

Second, the right instrument: the tip is a rounded metallic part, maybe a Suction Coagulator. Wait, but Suction Coagulator is a specific tool. Alternatively, maybe it's a different type. Wait, the list includes "Suction" as a separate entry. So if both are suction-related, but the exact names.

Wait, the problem says to list each tool only once even if multiple. But in this case, are there two different tools or same?

Wait, maybe the left is Suction and the right is Suction Coagulator. Let's check.

Suction Coagulator: a tool that suctions and coagulates. The right tool might be that. The left one is just Suction.

So detected tools would be ["Suction", "Suction Coagulator"].

Wait, but let's confirm. Let's look at the image again. The left instrument is a larger tube, maybe a suction, and the right one is a different type. Alternatively, maybe both are suction, but the list has "Suction" as a single entry. Wait, no---if they are different tools, but the names must be exact.

Wait, the valid tool names include "Suction" and "Suction Coagulator". So if one is Suction and the other is Suction Coagulator, then both are listed.

Alternatively, maybe the right tool is a Suction microdebrider? No, the microdebrider has a rotating tip, which isn't visible here.

Wait, another approach: check common endoscopic tools. In sinus surgery, for example, suction and coagulator are common. The image shows two instruments. Let's assume left is Suction, right is Suction Coagulator.

So the detected tools would be Suction and Suction Coagulator.
</think>

{"detected_tools": ["Suction", "Suction Coagulator"]}
\end{qwenanswer}

\subsection{Model Responses to Questions about Neurosurgery}

\begin{questionbox}{1}
What roles do suction, cotton patties, and hemostatic agents play during pituitary tumor resection?
\end{questionbox}

\begin{answerbox}
Okay, let's break down the roles of suction, cotton patties, and hemostatic agents during pituitary tumor resection. These are *critical* components of the surgical toolkit, and understanding their interplay is key to a successful and safe operation.  I'll cover each, then discuss how they work together.  This is a complex surgery, so I'll aim for a detailed but understandable explanation.  **Please remember I am an AI and this is for informational purposes only. It is not medical advice. Consult with a qualified healthcare professional for any health concerns or before making any decisions related to your health or treatment.**

**1. Suction**

* **Primary Role: Maintaining a Clear Operative Field & Removing Fluid/Debris.**  This is arguably the *most* important element throughout the entire procedure.  The surgeon needs to *see* what they're doing, and the surgical space within the sella (the bony cavity where the pituitary gland sits) is incredibly small and prone to bleeding.
* **Types of Suction Used:**
* **Irrigation Suction:**  This is the workhorse.  It simultaneously delivers irrigating fluid (usually normal saline) to wash away blood, debris, and tumor fragments *while* simultaneously suctioning it all away.  The fluid keeps the field clean and allows the surgeon to visualize the delicate structures.  Different tip sizes and shapes are used depending on the specific area being worked on.  Gentle irrigation is crucial to avoid damaging brain tissue.
* **Ultrasonic Aspirator (CUSA):**  Often used for softer tumors (like pituitary adenomas). CUSA uses high-frequency vibrations to break up the tumor tissue into small fragments, which are then suctioned away.  It's more selective than traditional suction, helping to preserve normal pituitary tissue and surrounding structures.  It's particularly useful for tumors that are adherent to critical structures.
* **Bipolar Suction:**  Some suction devices combine suction with bipolar electrocautery. This allows for simultaneous coagulation of small bleeding vessels while suctioning away the blood.
* **Why it's crucial in pituitary surgery:**
* **Limited Space:** The sella is tiny. Blood quickly obscures the view.
* **Proximity to Critical Structures:** The pituitary gland is nestled near the optic nerves, the cavernous sinus (a major venous structure), and the brainstem.  Clear visualization is essential to avoid damage.
* **Fragile Tissue:** The pituitary gland and surrounding tissues are delicate.

**2. Cottonoid Patties (Cotton Patties/Neuro-Patties)**

* **Primary Role: Hemostasis (Stopping Bleeding) & Tissue Protection.**  These aren't your typical cotton balls. Neuro-Patties are specifically designed for neurosurgery. They are made of highly absorbent, tightly woven cotton.
* **How they work:**
* **Direct Pressure:**  The primary mechanism.  Gently packing the patties into the surgical bed applies direct pressure to bleeding points, helping to stop the hemorrhage.
* **Absorption:** They soak up blood, further contributing to a dry field.
* **Temporary Space Filling:**  They can temporarily fill small defects in the bone or dura (the membrane covering the brain) created during tumor removal.
* **Tissue Protection:** They can act as a barrier to protect delicate structures from inadvertent injury during manipulation.
* **Specific Use in Pituitary Surgery:**
* **Sella Floor Repair:** After the tumor is removed, there's often a defect in the sella floor.  A patty soaked in hemostatic agent (see below) is often placed here to promote healing and prevent cerebrospinal fluid (CSF) leak.
* **Packing Around Vessels:**  Used to gently pack around the cavernous sinus or other vessels to control bleeding.
* **Dural Closure Assistance:**  Can help to support the dural repair.
* **Important Considerations:**
* **Careful Removal:** Patties *must* be carefully removed after the surgery to avoid leaving any remnants behind, which could cause infection or inflammation.  They are often marked with radiopaque threads to aid in identification on post-operative imaging.
* **Not a Long-Term Solution:**  Patties are for temporary hemostasis and space filling. They don't promote permanent tissue repair on their own.

**3. Hemostatic Agents**

* **Primary Role:  Enhancing Hemostasis (Stopping Bleeding).** These are substances applied to bleeding surfaces to accelerate clot formation. They are often used *in conjunction* with suction and cotton patties.
* **Types of Hemostatic Agents Commonly Used in Pituitary Surgery:**
* **Gelatin-Based (e.g., Gelfoam):**  A porous sponge that provides a matrix for clot formation. It's relatively inexpensive and effective for controlling minor bleeding.
* **Oxidized Regenerated Cellulose (e.g., Surgicel):**  Another absorbent material that promotes clotting. It has some antibacterial properties.
* **Thrombin:** A potent clotting factor.  Often used in combination with collagen or other carriers.  Can be very effective for controlling more significant bleeding.  (There are bovine and human forms; human thrombin is preferred to minimize immune reactions).
* **Fibrin Sealants:**  Contain fibrinogen and thrombin, mimicking the natural clotting cascade.  They create a strong, adhesive clot.  Often used for dural closure.
* **Polysaccharide-Based (e.g., Arista AH):** Derived from plant sources, these agents promote rapid clot formation and are biocompatible.
* **How they're used in pituitary surgery:**
* **Soaking Cotton Patties:**  Patties are frequently soaked in a hemostatic agent before being placed in the surgical bed. This provides both direct pressure and a boost to clotting.
* **Direct Application:**  Hemostatic agents can be applied directly to bleeding vessels or surfaces.
* **Dural Sealants:** Fibrin sealants are often used to reinforce the dural closure, reducing the risk of CSF leak.
* **Important Considerations:**
* **Appropriate Agent Selection:** The choice of hemostatic agent depends on the type and severity of bleeding, the location, and the surgeon's preference.
* **Potential Complications:**  While generally safe, some hemostatic agents can cause inflammation or allergic reactions.

**How They Work Together: A Surgical Scenario**

Imagine the surgeon is removing a pituitary adenoma:

1. **Initial Exposure:** The surgeon gains access to the sella through the nose (transsphenoidal approach).
2. **Tumor Dissection:**  Using microsurgical instruments, the surgeon carefully dissects the tumor from surrounding structures.  **Continuous irrigation suction** is used to keep the field clear of blood and debris.
3. **Bleeding Control:**  As vessels are encountered, the surgeon may use **bipolar cautery** for small vessels. For larger bleeding points, they might apply a **hemostatic agent** (like thrombin) directly or pack the area with a **cotton patty soaked in hemostatic agent**.
4. **Tumor Removal (CUSA):** If the tumor is soft, a **CUSA** might be used to debulk it, with simultaneous suction to remove the fragmented tissue.
5. **Sella Floor Repair:** Once the tumor is removed, any defects in the sella floor are addressed. A **cotton patty saturated with a hemostatic agent and potentially a dural sealant** is carefully placed to promote healing and prevent CSF leak.
6. **Dural Closure:** The dura is closed, often with sutures and reinforced with a **fibrin sealant**.
7. **Final Inspection:**  The surgeon meticulously inspects the field with **suction** to ensure complete hemostasis and remove any remaining debris or packing material.

**In conclusion:** Suction, cotton patties, and hemostatic agents are not independent tools but rather a coordinated system used to achieve a safe and effective pituitary tumor resection.  They work synergistically to provide visualization, control bleeding, and promote healing in this delicate and challenging surgery.

**Disclaimer:**  This information is for general knowledge and informational purposes only, and does not constitute medical advice. It is essential to consult with a qualified healthcare professional for any health concerns or before making any decisions related to your health or treatment.
\end{answerbox}

\section{Gemma 3 27B Responses to Neurosurgery Questions}
\label{sec:appendix_gemma_responses}

This appendix presents the complete responses from Gemma 3 27B to five neurosurgery questions. The model was prompted with each question and generated the responses shown below.

\subsection{Model Configuration}

\begin{table}[H]
\centering
\begin{tabular}{ll}
\toprule
Parameter & Value \\
\midrule
Model & Gemma 3 27B Instruct \\
Model ID & \texttt{google/gemma-3-27b-it} \\
Precision & bfloat16 \\
Max New Tokens & 8,192 \\
Temperature & 0.0\\
\bottomrule
\end{tabular}
\caption{Gemma 3 27B generation configuration.}
\label{tab:gemma_config}
\end{table}

\vspace{1em}

\begin{questionbox}{2}
Describe the standard steps of an endoscopic endonasal transsphenoidal approach to a pituitary adenoma.
\end{questionbox}

\begin{answerbox}
## Endoscopic Endonasal Transsphenoidal Approach to Pituitary Adenoma: A Step-by-Step Description

The endoscopic endonasal transsphenoidal approach (EETS) is now the gold standard for removing most pituitary adenomas. Here's a detailed breakdown of the standard steps involved.  It's a complex procedure, and this is a general overview; specifics can vary based on tumor size, location, and surgeon preference.

**I. Preoperative Preparation:**

* **Imaging:** High-resolution MRI with and without contrast is *essential* for planning. CT scans may be used to assess sphenoid sinus anatomy.  Angiography may be considered for very large or invasive tumors to assess relationship to major vessels.
* **Hormonal Evaluation:**  Complete endocrine workup to assess pituitary function and identify hormone excess.
* **Visual Field Testing:**  To document pre-operative visual function, especially if the tumor is compressing the optic chiasm.
* **Patient Counseling:**  Detailed discussion of risks, benefits, alternatives, and expected recovery.
* **Medical Optimization:**  Management of any co-morbidities and potential cessation of blood thinners.

**II. Intraoperative Steps:**

**1. Anesthesia & Positioning:**

* **General Anesthesia:**  The patient is placed under general anesthesia.
* **Positioning:** Typically a supine position with the head slightly elevated and fixed in a three-point head holder.  This allows for optimal visualization and access.  Sometimes a slight reverse Trendelenburg position is used.

**2. Nasal Preparation & Endoscopic Access:**

* **Nasal Decongestion:**  Topical decongestants (like oxymetazoline) are used to shrink nasal mucosa.
* **Endoscopic Setup:** A rigid endoscope (typically 0- or 30-degree) is inserted through one nostril.  The surgeon operates by looking at a monitor displaying the endoscopic view.
* **Septal Mucosa Incision:** A small incision is made in the nasal septum (the cartilage dividing the nostrils). This is often unilateral (one side) but can be bilateral if needed for wider access.
* **Sphenoid Sinus Entry:** The posterior nasal septum is removed to expose the sphenoid sinus.  The sphenoid sinus is then widely opened.  This often involves removing any septations or disease within the sinus.  Image guidance (see below) is crucial here.

**3. Sellar Floor Exposure & Opening:**

* **Sellar Floor Identification:** The surgeon identifies the bony sphenoid planum (sellar floor), which is the bone separating the sphenoid sinus from the sella turcica (the bony cavity housing the pituitary gland).
* **Sellar Floor Removal:** Using a combination of micro-instruments (drills, curettes, and endoscopes), the surgeon carefully removes the bony sphenoid planum to create an opening into the sella turcica.  This is done under constant endoscopic visualization and often with the assistance of image guidance.  The dura (the membrane covering the brain) is usually visualized at this point.
* **Dural Opening:** A small incision is made in the dura mater to access the sella.  The size of the dural opening is tailored to the size of the tumor.

**4. Tumor Resection:**

* **Tumor Debulking:**  The surgeon begins to remove the tumor.  For softer tumors, this can be done with suction and micro-instruments.  For harder or fibrous tumors, a combination of techniques is used.
* **Internal Decompression:**  The tumor is often debulked from within, starting with the central portion.
* **Capsule Dissection:**  The tumor capsule is carefully dissected from the surrounding structures the pituitary gland, optic chiasm, cavernous sinuses, and sphenoid sinus walls.  This is the most critical and delicate part of the surgery.
* **Complete Resection (Goal):** The goal is to achieve gross total resection (GTR) of the tumor, meaning all visible tumor is removed.  However, this isn't always possible or safe, especially for invasive tumors.
* **Suction Irrigation:**  Continuous irrigation and suction are used throughout the resection to maintain a clear surgical field.

**5. Reconstruction & Closure:**

* **Dural Repair:**  The dural opening is meticulously repaired.  This is often done with a combination of:
* **Fat Graft:**  Fat harvested from the abdomen is used to provide a watertight seal.
* **Dural Substitutes:**  Collagen matrices or other synthetic materials may be used in conjunction with fat.
* **Fibrin Glue:**  Fibrin glue is often applied to reinforce the dural repair.
* **Sphenoid Sinus Reconstruction (Optional):**  In some cases, the sphenoid sinus is partially reconstructed with a flap or graft to prevent cerebrospinal fluid (CSF) leaks.
* **Nasal Packing:**  Nasal packing (e.g., with absorbable material or silicone splints) is placed to support the nasal mucosa and promote healing.

**III.  Important Adjuncts:**

* **Image Guidance (Neuronavigation):**  This is *crucial*.  It uses pre-operative MRI or CT scans to create a 3D map of the patient's anatomy, allowing the surgeon to precisely locate the tumor and surrounding structures during surgery.
* **Intraoperative Neuromonitoring (IONM):**  This may be used for larger or more complex tumors, especially those near the optic nerves.  It involves monitoring the function of the optic nerves and brainstem during surgery to detect any potential injury.
* **Fluorescence-Guided Surgery:**  Some surgeons use fluorescent dyes (like 5-ALA) to help differentiate tumor tissue from normal pituitary tissue, improving the extent of resection.

**IV. Postoperative Care:**

* **ICU Monitoring:**  Patients are typically monitored in the intensive care unit (ICU) for 1-2 days.
* **Hormone Replacement:**  Postoperative hormone deficiencies are common and require hormone replacement therapy (e.g., cortisol, thyroid hormone, growth hormone).
* **CSF Leak Monitoring:**  Close monitoring for CSF leaks is essential.
* **Follow-up MRI:**  A follow-up MRI is performed within a few weeks to assess the extent of resection and rule out any complications.
* **Endocrine Follow-up:**  Regular endocrine follow-up is necessary to monitor hormone levels and adjust hormone replacement therapy as needed.

**Disclaimer:** This is a general overview and should not be considered medical advice.  The specific steps and techniques used in an EETS procedure can vary depending on the individual patient and the surgeon's experience.  Always consult with a qualified neurosurgeon for any medical concerns.
\end{answerbox}

\vspace{1em}

\begin{questionbox}{3}
Compare microscopic versus endoscopic approaches for pituitary surgery.
\end{questionbox}

\begin{answerbox}
## Microscopic vs. Endoscopic Pituitary Surgery: A Detailed Comparison

Pituitary surgery aims to remove pituitary tumors (adenomas) that cause hormonal imbalances or compress surrounding structures like the optic nerves.  Both microscopic and endoscopic approaches are common, but they differ significantly in technique, advantages, and disadvantages. Here's a detailed comparison:

**1. Surgical Approach & Technique:**

* **Microscopic Surgery (Transsphenoidal):**
* **Access:** Traditionally, this involves a surgical incision inside the nose (transsphenoidal approach) to reach the sphenoid sinus, a cavity behind the nose.  A speculum is used to retract tissues, and a high-powered surgical microscope is essential for visualization.
* **Visualization:** Relies heavily on the surgeon's skill and the magnification provided by the microscope.  The surgeon directly views the pituitary gland and tumor through the microscope.
* **Instrumentation:** Uses specialized micro-instruments manipulated directly by the surgeon under microscopic guidance.
* **Sinus Involvement:**  Often requires significant opening of the sphenoid sinus to gain adequate access.
* **Endoscopic Surgery (Transsphenoidal):**
* **Access:** Also uses a transsphenoidal approach, but instead of a microscope and speculum, a thin, flexible endoscope (a tube with a camera and light source) is inserted through the nostrils.
* **Visualization:** Provides a wide-angle, high-definition view on a monitor.  The surgeon views the surgical field on the monitor, manipulating instruments through channels in the endoscope.
* **Instrumentation:** Uses specialized endoscopic instruments that are smaller and more maneuverable than microscopic instruments.
* **Sinus Involvement:**  Generally requires less sphenoid sinus opening than microscopic surgery, potentially minimizing nasal complications.

**2. Advantages & Disadvantages:**

| Feature | Microscopic Surgery | Endoscopic Surgery |
|---|---|---|
| **Visualization** | Excellent depth perception, tactile feedback.  Surgeon directly views the field. | Wide-angle view, high definition.  Can visualize around corners.  Less depth perception. |
| **Surgical Precision** | Highly precise, especially for complex tumors.  Surgeon has direct control and feel. |  Precision improving rapidly with advancements in endoscope technology and surgeon experience. |
| **Tumor Size** | Well-suited for a wide range of tumor sizes, including larger and more complex tumors. |  Historically better for smaller, less complex tumors.  Increasingly used for larger tumors with experienced surgeons. |
| **Sinus Complications** | Higher risk of nasal congestion, sinusitis, cerebrospinal fluid (CSF) leak, and septal perforation due to wider sphenoid sinus opening. | Lower risk of these complications due to less sinus disruption. |
| **Nasal Morbidity** |  More significant nasal packing often required post-operatively, leading to discomfort. |  Less or no nasal packing often needed, leading to faster recovery and less discomfort. |
| **Recovery Time** |  Generally longer recovery due to more extensive sinus manipulation and nasal packing. |  Generally faster recovery with less pain and quicker return to normal activities. |
| **Learning Curve** |  Steeper learning curve for surgeons initially, but well-established technique. |  Steeper learning curve for surgeons transitioning from microscopic to endoscopic techniques. Requires specialized training. |
| **Cost** |  Can be slightly less expensive initially (equipment costs are established). |  Can be more expensive due to the cost of endoscopes and specialized instruments. |
| **Surgeon Experience** |  Widely available with many experienced surgeons. |  Availability of highly experienced endoscopic pituitary surgeons is growing but may be limited in some areas. |
| **Anterior Pituitary Preservation** |  Generally good, but can be challenging with larger tumors. |  Potentially better preservation of normal anterior pituitary function due to less manipulation. |

**3. Specific Tumor Types & Considerations:**

* **Microadenomas (<10mm):**  Both approaches are effective. Endoscopic is often preferred due to faster recovery and lower morbidity.
* **Macroadenomas (>10mm):**  Microscopic surgery has traditionally been the gold standard, especially for complex macroadenomas with significant extension. However, experienced endoscopic surgeons are increasingly successful with these tumors.
* **Aggressive Tumors (e.g., invasive adenomas):** Microscopic surgery may be favored for more extensive resection and better control.
* **Recurrent Tumors:**  Endoscopic approach can be useful for accessing residual tumor tissue after previous microscopic surgery.
* **Craniopharyngiomas:**  Both approaches can be used, but endoscopic may be preferred for smaller lesions and to minimize brain retraction.
* **Cholesterol Granulomas:** Endoscopic approach is often preferred for these cysts due to its ability to effectively decompress the cyst and preserve pituitary function.

**4.  Current Trends & Future Directions:**

* **Increasing adoption of endoscopic surgery:**  Endoscopic techniques are becoming increasingly popular and are now considered the preferred approach by many surgeons, particularly for smaller and less complex tumors.
* **Hybrid approaches:**  Some surgeons utilize a combination of microscopic and endoscopic techniques to leverage the advantages of both.
* **Robotic assistance:**  Robotic-assisted endoscopic pituitary surgery is emerging as a potential future direction, offering even greater precision and dexterity.
* **Improved endoscope technology:**  Advancements in endoscope resolution, illumination, and instrument design are continually improving the capabilities of endoscopic surgery.

**In conclusion:**

Both microscopic and endoscopic approaches are viable options for pituitary surgery. The best approach depends on several factors, including the **size and type of tumor, the surgeon's experience, and the patient's overall health.** 

**It's crucial to discuss the risks and benefits of each approach with a neurosurgeon specializing in pituitary surgery to determine the most appropriate treatment plan.**

**Disclaimer:** I am an AI chatbot and cannot provide medical advice. This information is for general knowledge and informational purposes only, and does not constitute medical advice. It is essential to consult with a qualified healthcare professional for any health concerns or before making any decisions related to your health or treatment.
\end{answerbox}

\vspace{1em}

\begin{questionbox}{4}
Compare the use of a Rhoton dissector with suction or curettes during pituitary adenoma resection. When is each preferred?
\end{questionbox}

\begin{answerbox}
## Rhoton Dissector vs. Suction/Curettes in Pituitary Adenoma Resection: A Comparison

Both the Rhoton dissector and suction/curettes are crucial tools in transsphenoidal pituitary adenoma resection, but they serve different purposes and are used in different phases of the surgery. Here's a detailed comparison, including when each is preferred:

**1. Rhoton Dissector**

* **What it is:** A thin, blunt instrument with a rounded tip, typically made of titanium or a similar material. It comes in various sizes and angles.  It's designed for *dissection*---gently separating the tumor from surrounding structures.
* **Mechanism of Action:**  Works by creating a plane between the tumor and the pituitary gland, cavernous sinus, optic chiasm, and dura. It relies on blunt dissection, minimizing trauma.
* **Advantages:**
* **Precise Dissection:** Excellent for delicate dissection around critical neurovascular structures (optic chiasm, internal carotid artery).
* **Minimizes Trauma:**  Blunt dissection reduces the risk of bleeding and damage to surrounding tissues compared to sharp instruments or aggressive curettage.
* **Visualization:**  Helps to define the tumor capsule and identify the optimal plane for removal.
* **Soft/Cystic Tumors:** Particularly useful for softer, cystic adenomas where suction alone may not be effective.
* **Capsule Dissection:**  Essential for debulking and defining the tumor capsule, especially for larger or invasive adenomas.
* **Disadvantages:**
* **Time-Consuming:**  Blunt dissection is slower than curettage.
* **Requires Skill:**  Effective use requires significant surgical experience and a thorough understanding of anatomy.
* **Limited for Hard Tumors:** Less effective for very firm, fibrous adenomas.
* **Can't Remove Fluid/Debris:** Doesn't actively remove tumor material; it *exposes* it for removal by other means.

**2. Suction & Curettes**

* **What they are:**
* **Suction:**  A device that uses negative pressure to remove fluid, blood, and soft tissue fragments.  Often uses a Frazier tip or similar.
* **Curettes:**  Loop-shaped instruments (various sizes and shapes - e.g., Ball, Luer, Janzen) used to scrape and remove soft tissue.  Can be rigid or flexible.
* **Mechanism of Action:**
* **Suction:** Removes fluid and soft debris, improving visualization.
* **Curettes:**  Mechanically remove tumor tissue by scraping.
* **Advantages:**
* **Efficient Removal:**  Rapidly removes large volumes of soft tumor tissue.
* **Good for Soft/Friable Tumors:**  Effective for adenomas that are easily fragmented.
* **Hemostasis:** Suction can help control bleeding by removing blood from the surgical field.
* **Debulking:**  Quickly reduces tumor size, facilitating further dissection.
* **Disadvantages:**
* **Risk of Trauma:**  Aggressive curettage can damage the pituitary gland, optic chiasm, or cavernous sinus.
* **Poor Visualization:**  Can create a bloody field, obscuring visualization.
* **Not Ideal for Encapsulated Tumors:**  May not effectively remove tumors with a well-defined capsule without first being dissected.
* **Risk of CSF Leak:**  Aggressive curettage of the sella can increase the risk of cerebrospinal fluid leak.

**When is each preferred? A Phased Approach & Specific Scenarios**

The use of these tools is often *sequential* and *complementary* during a pituitary adenoma resection.  Here's a breakdown:

**Phase 1: Initial Entry & Tumor Exposure**

* **Rhoton Dissector:**  Primarily used for the initial entry into the sella, opening the sphenoid sinus, and identifying the dura.  It's also used to initially define the tumor margins and begin dissecting the tumor from the pituitary gland.
* **Suction:** Used to remove any blood or debris encountered during the initial entry and to maintain a clear surgical field.

**Phase 2: Tumor Debulking & Dissection**

* **Rhoton Dissector:**  Crucial for dissecting the tumor from the optic chiasm, cavernous sinus, and pituitary gland.  This is especially important for:
* **Optically Important Tumors:**  Tumors that are close to or compressing the optic chiasm.  Gentle dissection with the Rhoton is paramount.
* **Invasive Adenomas:**  Tumors that extend beyond the sella.
* **Larger Adenomas:**  To define the tumor capsule and create a plane for removal.
* **Suction & Curettes:** Used *after* initial dissection with the Rhoton.  
* **Soft, Friable Tumors:** Curettes are used to remove the bulk of the tumor tissue that has been exposed by the Rhoton dissector. Suction simultaneously removes debris and maintains visualization.
* **Cystic Components:** Suction is excellent for removing cystic fluid.

**Phase 3: Final Removal & Pituitary Reconstruction**

* **Rhoton Dissector:** Used for meticulous removal of any remaining tumor fragments, especially in areas near critical structures.
* **Suction & Curettes:**  Used to remove any residual tumor tissue and to carefully curette the sella floor to ensure complete removal.  However, curettage is used more cautiously in this phase to minimize the risk of pituitary damage or CSF leak.

**Specific Scenarios & Preferences:**

| Scenario | Preferred Tool(s) | Rationale |
|---|---|---|
| **Small, Soft Prolactinoma** | Suction & Curettes (primarily) |  Often easily fragmented and removed with minimal dissection. |
| **Large, Invasive Prolactinoma** | Rhoton Dissector + Suction/Curettes | Requires careful dissection from surrounding structures before debulking. |
| **Non-Functioning Adenoma with Chiasm Compression** | Rhoton Dissector (primary) + Suction/Curettes |  Optic chiasm preservation is paramount; gentle dissection is essential. |
| **Firm, Fibrous Adenoma** | Rhoton Dissector + Micro-instruments |  Requires more meticulous dissection; curettage may be less effective. |
| **Recurrent Adenoma** | Rhoton Dissector + Suction/Curettes |  Often involves scar tissue, requiring careful dissection. |
| **Cystic Adenoma** | Suction (primary) + Rhoton Dissector |  Suction effectively removes the cyst fluid, and the Rhoton helps define the capsule. |

**Important Considerations:**

* **Endoscopic vs. Microscopic Approach:**  The specific techniques and tool preferences can vary slightly depending on whether the surgery is performed using an endoscope or a microscope. Endoscopes often allow for more angled visualization and may favor certain Rhoton dissector shapes.
* **Surgeon's Experience:**  The surgeon's skill and preference play a significant role in the choice of instruments and techniques.
* **Intraoperative Monitoring:**  Neurophysiological monitoring (e.g., visual evoked potentials) can guide the surgeon's dissection and minimize the risk of damage to critical structures.

**Disclaimer:**  I am an AI chatbot and cannot provide medical advice. This information is for general knowledge and informational purposes only, and does not constitute medical advice. It is essential to consult with a qualified healthcare professional for any health concerns or before making any decisions related to your health or treatment.
\end{answerbox}

\vspace{1em}

\begin{questionbox}{5}
What anatomical landmarks are used to identify the sella turcica during transsphenoidal pituitary surgery, and why are they critical for avoiding complications?
\end{questionbox}

\begin{answerbox}
## Anatomical Landmarks for Sella Turcica Identification in Transsphenoidal Pituitary Surgery & Why They Matter

Transsphenoidal pituitary surgery is a delicate procedure requiring precise anatomical knowledge. Identifying the sella turcica (Turkish saddle) and its surrounding structures is *critical* for successful surgery and minimizing complications. Here's a breakdown of the key landmarks and their importance:

**I. External Landmarks (Pre-operative Imaging & Intraoperative Guidance)**

These are identified primarily through pre-operative imaging (CT and MRI) and then confirmed intraoperatively using fluoroscopy, neuronavigation, and direct visualization.

* **Dorsum Sellae:** This is the posterior wall of the sella, formed by the posterior clinoid processes. It's the "back" of the saddle.  *Importance:*  Defining the dorsum sellae helps establish the posterior limit of the surgical field.  Overly aggressive removal of the dorsum can lead to cerebrospinal fluid (CSF) leak and damage to the pituitary stalk.
* **Tuberculum Sellae:** The anterior projection of the sella, formed by the anterior clinoid processes. It's the "front" of the saddle. *Importance:*  Helps define the anterior boundary.  Fracturing the tuberculum sellae can cause bleeding and potentially damage the optic nerves/chiasm.
* **Clivus:** The bony base of the skull anterior to the pons. The sella sits *on* the clivus. *Importance:*  The clivus provides a key reference point for the overall orientation and depth of the surgical approach.  Understanding its contours is vital to avoid injury to the brainstem.
* **Sphenoid Sinus:** The air-filled cavity *below* the sella. The surgeon accesses the sella by opening the sphenoid sinus. *Importance:*  The size and pneumatization (air-filling) of the sphenoid sinus vary significantly between individuals.  Knowing its anatomy is crucial for safe and effective sinus entry.  Variations like concha (bony shelves) within the sinus need to be identified to avoid damage.
* **Carotid Groove:**  Located laterally and inferior to the sella, housing the internal carotid arteries. *Importance:*  This is arguably the *most critical* landmark. The internal carotid arteries are immediately adjacent to the sella and are at high risk of injury during surgery.  Identifying the carotid groove and its relationship to the sella is paramount.
* **Optic Canal & Optic Strut:**  The bony canal through which the optic nerves pass, located superolaterally to the sella. The optic strut is the bony support between the optic canal and the sella. *Importance:*  The optic nerves and chiasm lie *above* the sella and are vulnerable to compression or injury.  The optic canals and strut define the superior limits of the surgical field and help protect the visual pathways.
* **Superior Sella Rim:** The bony edge of the sella superiorly. *Importance:*  Helps define the superior limit of the sella and the proximity of the optic structures.
* **Foramen Ovale:** Located posterolaterally, containing the accessory meningeal artery. *Importance:*  While not directly within the sella, it's a nearby structure that can be encountered during surgery and a potential source of bleeding.

**II. Intraoperative Landmarks (Once the Sphenoid Sinus is Entered)**

Once the surgeon has entered the sphenoid sinus, these landmarks become visible:

* **Sellar Floor:** The bony floor of the sella, which the surgeon must remove to access the pituitary gland. *Importance:*  Careful removal of the floor is essential to avoid dural tears and CSF leaks.
* **Dural Ring:** The circular opening in the dura mater covering the sella. *Importance:*  The dural ring is incised to gain access to the subarachnoid space and the pituitary gland.  The incision must be carefully planned to minimize tension and prevent CSF leaks.
* **Pituitary Gland:** The target of the surgery, usually appearing as a soft, pinkish-gray mass within the sella. *Importance:*  Visualizing the gland and any associated tumor is crucial for accurate resection.
* **Pituitary Stalk:** The connection between the pituitary gland and the hypothalamus. *Importance:*  Preserving the pituitary stalk is often desirable to maintain residual pituitary function.
* **Cavernous Sinus:** Located laterally to the sella, containing the internal carotid artery, abducens nerve (CN VI), and oculomotor nerve (CN III). *Importance:*  The cavernous sinus is a critical landmark for avoiding injury to the internal carotid artery and cranial nerves.

**Why are these landmarks critical for avoiding complications?**

Complications in transsphenoidal pituitary surgery can be devastating.  Precise anatomical knowledge and identification of these landmarks are essential to prevent:

* **Cerebrospinal Fluid (CSF) Leak:**  Damage to the dura mater or inadequate dural repair can lead to CSF leakage, potentially causing meningitis.
* **Internal Carotid Artery Injury:**  The internal carotid artery is in close proximity to the sella. Injury can cause stroke or massive hemorrhage.
* **Optic Nerve/Chiasm Damage:**  Compression or injury to the optic nerves/chiasm can result in vision loss (blindness).
* **Hypopituitarism:**  Damage to the pituitary gland or pituitary stalk can lead to hormone deficiencies.
* **Cranial Nerve Palsies:**  Injury to cranial nerves within the cavernous sinus (CN III, CN VI) can cause double vision or other neurological deficits.
* **Bleeding:**  Damage to blood vessels during surgery can lead to significant bleeding.

**Tools to Aid Landmark Identification:**

* **Preoperative Imaging (CT & MRI):**  Provides a detailed roadmap of the sella and surrounding structures.
* **Fluoroscopy:**  Real-time X-ray imaging used intraoperatively to confirm the position of instruments and identify bony landmarks.
* **Neuronavigation:**  A computer-assisted surgical planning and guidance system that uses pre-operative imaging to create a 3D map of the surgical field.  This allows the surgeon to precisely locate anatomical structures during surgery.
* **Endoscopic Visualization:**  Provides a magnified view of the surgical field, allowing for better identification of anatomical landmarks.

In conclusion, successful and safe transsphenoidal pituitary surgery relies heavily on a thorough understanding of the sella turcica's anatomy and meticulous identification of its surrounding landmarks.  The use of advanced imaging and surgical guidance technologies further enhances the surgeon's ability to navigate this complex anatomical region and minimize the risk of complications.
\end{answerbox}

%% file: sections/appendix_new_tasks.tex
\section{Beyond Detection: Four Additional Annotation Tasks}
\label{sec:appendix_newtasks}

This appendix provides the full protocol for the four bounding-box-free annotation tasks introduced in Section~\ref{subsec:newtasks_methods}, whose results are reported in Section~\ref{subsec:newtasks}. The four benchmarks cover the two families of surgical-video annotation that object detection does not address---temporal workflow understanding and intent analysis---plus anatomy recognition, and none involves bounding boxes.

\subsection{Datasets, tasks, and splits}

\textbf{PitVQA} (workflow recognition; \citealp{he2024pitvqa}). Endoscopic pituitary surgery (endonasal transsphenoidal approach), 25 videos. Each frame is labeled with one of 3 surgical phases (\texttt{nasal\_sphenoid}, \texttt{sellar}, \texttt{closure}) and one of 14 surgical steps (e.g., \texttt{nasal\_corridor\_creation}, \texttt{tumour\_excision}, \texttt{haemostasis}). We follow the official split of 20 training and 5 validation videos (84{,}332 and 24{,}767 frames, respectively) and score phase and step jointly: a prediction is an exact match only when both are correct, and micro-F1 is computed over the union of the phase and step label sets. The underlying videos come from the same public challenge as PitVis-2023 (Section~\ref{subsec:pitvis_methods}), but the labels and task differ.

\textbf{SAR-RARP50} (gesture recognition; \citealp{psychogyios2023sarrarp50}). Robot-assisted radical prostatectomy, suturing segment, 50 operations with 10\,Hz action annotations over 8 gesture classes (\texttt{Other}, \texttt{Picking Up The Needle}, \texttt{Positioning The Needle Tip}, \texttt{Pushing The Needle Through The Tissue}, \texttt{Pulling The Needle Out Of The Tissue}, \texttt{Tying A Knot}, \texttt{Cutting The Suture}, \texttt{Returning Or Dropping The Needle}). We export the temporal midpoint frame of every contiguous gesture segment and split by operation (seed 42): 35 operations / 1{,}563 frames for training and 15 operations / 636 frames for validation.

\textbf{CholecT50 action recognition} \citep{cholec}. Laparoscopic cholecystectomy, 50 videos. Each frame of CholecT50 is annotated with action triplets $\langle$instrument, verb, target$\rangle$; we take the verb component as a multi-label intent-analysis target over 9 instrument--tissue actions (\texttt{grasp}, \texttt{retract}, \texttt{dissect}, \texttt{coagulate}, \texttt{clip}, \texttt{cut}, \texttt{aspirate}, \texttt{irrigate}, \texttt{pack}) plus an explicit \texttt{idle} class assigned to frames with no active action, so every frame carries at least one label. We reuse the video-level split of the tool-detection protocol (seed 42; Section~\ref{subsec:cholect50_methods}): 40 videos / 80{,}940 frames for training and 10 videos / 19{,}923 frames for validation, so no video crosses splits and the task shares images, but not labels or output space, with the tool-detection benchmark.

\textbf{DSAD} (anatomy presence; \citealp{carstens2023dresden}). The Dresden Surgical Anatomy Dataset: robot-assisted rectal resection, multi-label presence of 12 abdominal anatomical structures (abdominal wall, colon, inferior mesenteric artery, intestinal veins, liver, pancreas, small intestine, spleen, stomach, ureter, uterus, vesicular glands). We use the official train/validation/test splits of the public release (7{,}889 / 1{,}978 / 3{,}328 frames) and score on the validation split.

\subsection{Models and adaptation protocols}

All evaluations are frame-level: each model receives a single RGB frame plus the task's label ontology, so no model class benefits from temporal context. Sampling temperature is 1.0 with a 1{,}024-token generation cap for all zero-shot VLM evaluations, matching the tool-detection protocol; outputs that fail schema validation are counted as incorrect predictions. On SAR-RARP50 and CholecT50 action recognition, the Qwen3.8 27B evaluation disables the model's thinking mode to keep generations within the token budget; scoring is identical.

\textbf{Zero-shot open-weight VLMs.} Gemma~3 27B-it (the strongest open-weight zero-shot model in the tool-detection evaluation) and Qwen3.8 27B, prompted with the templates in Appendix~\ref{sec:appendix_newtasks_prompts}.

\textbf{Frontier VLMs.} Claude Fable~5, Claude Opus~5, Claude Sonnet~5, Gemini~3.1 Pro, Gemini~3.7 Flash, GPT-5.6 Luna, GPT-5.6 Sol, and GPT-5.6 Terra, called through their public APIs with the identical prompts and temperature. On the three larger validation splits (PitVQA, CholecT50 action recognition, and DSAD) frontier models are scored on a random seed-42 subsample of 1{,}000 frames; on SAR-RARP50 they are scored on the full validation split.

\textbf{Surgical foundation model + linear probe.} LemonFM \citep{che2026lemonlargeendoscopicmonocular}, a ConvNeXt-Large backbone pretrained at scale on endoscopic video. The backbone is frozen; a single linear layer is trained per task with binary cross-entropy on the training split. Decision rules follow each task's structure: independent 0.5 thresholds for the multi-label DSAD and CholecT50 action-recognition tasks, argmax within the phase group and within the step group for PitVQA, and a single argmax for SAR-RARP50.

\textbf{LoRA fine-tuned VLM.} Gemma~3 27B fine-tuned on each task's training split with LoRA ($r=16$, learning rate $10^{-4}$, batch size 1 per device, effective batch size 16; 32 for DSAD and CholecT50 action recognition) using the same autoregressive JSON output head as the tool-detection fine-tuning (Section~\ref{subsec:lora_methods}); the evaluation prompt is identical to the zero-shot prompt. The number of epochs is scaled inversely with training-set size: 3 for DSAD, 2 for PitVQA (16{,}000 sampled frames, each contributing multiple question--answer pairs), 2 for CholecT50 action recognition (80{,}940 frames), and 3 for SAR-RARP50.

We additionally train the linear classification-head variant from the tool-detection protocol on each task: mean pooling over the final hidden states feeds task-appropriate linear heads (12-way and 10-way sigmoid heads for the multi-label DSAD and CholecT50 action-recognition tasks, trained with binary cross-entropy; a single softmax head for SAR-RARP50 trained with cross-entropy; and two softmax heads, one for the 3 phases and one for the 14 steps, for PitVQA). LoRA $r = 128$, $\alpha = 256$, dropout 0.05, adapting the attention projections of both the language model and the vision tower. Training: 10 epochs, learning rate $5 \times 10^{-6}$, effective batch size 16, bfloat16, PyTorch DDP on H100 GPUs.

\textbf{Supervised specialists.} ResNet-50 \citep{he2016deep} initialized from ImageNet-1K weights for all four tasks; no detection components are involved. For the multi-label DSAD and CholecT50 action-recognition tasks and the two-head PitVQA task: learning rate $10^{-4}$, up to 25, 25, and 4 epochs, respectively, with a seed-42 frame-level monitor holdout (10\%, 10\%, and 5\% of training frames) for early stopping. For the single-label SAR-RARP50 task: AdamW, learning rate $10^{-4}$, weight decay $10^{-4}$, cosine annealing, batch size 32, up to 100 epochs with early stopping (patience 15) on a seed-42 video-level monitor split (10\% of training videos). No validation frames are used during any training.

\textbf{Majority class baselines.} Computed from training-split label frequencies and scored on the validation split: \texttt{abdominal\_wall} for DSAD (1.42\% exact match, 29.65\% micro-F1), \texttt{phase:sellar} + \texttt{step:tumour\_excision} for PitVQA (29.73\%, 39.37\%), \texttt{dissect} for CholecT50 action recognition (18.46\%, 39.82\%), and \texttt{Positioning The Needle Tip} for SAR-RARP50 (25.63\% for both metrics).

\subsection{Prompt templates for the four additional tasks}
\label{sec:appendix_newtasks_prompts}

\begin{promptbox}
Identify all anatomical structures visible in this laparoscopic surgical video frame.

Valid structure names:
- abdominal_wall
- colon
- inferior_mesenteric_artery
- intestinal_veins
- liver
- pancreas
- small_intestine
- spleen
- stomach
- ureter
- uterus
- vesicular_glands

Return your answer as a JSON object with this exact format:
{"visible_structures": ["structure_name_1", "structure_name_2"]}

Use ONLY the exact structure names from the list above.
If a structure appears multiple times, list it only once.
If none of the listed structures are visible, return: {"visible_structures": []}

Return ONLY valid JSON, nothing else.
\end{promptbox}

\begin{promptbox}
This frame is from an endoscopic pituitary surgery (endonasal transsphenoidal approach). Identify the current surgical phase and the current surgical step.

Valid phases:
- closure
- nasal_sphenoid
- sellar

Valid steps:
- anterior_sphenoidotomy
- debris_clearance
- dural_sealant
- durotomy
- fat_graft_placement
- gasket_seal_construct
- haemostasis
- nasal_corridor_creation
- nasal_packing
- sellotomy
- septum_displacement
- sphenoid_sinus_clearance
- synthetic_graft_placement
- tumour_excision

Return your answer as a JSON object with this exact format:
{"phase": "<one phase from the list>", "step": "<one step from the list>"}

Use ONLY the exact names from the lists above.
Return ONLY valid JSON, nothing else.
\end{promptbox}

\begin{promptbox}
Identify all surgical actions currently being performed in this laparoscopic cholecystectomy video frame.

Valid action names:
- grasp
- retract
- dissect
- coagulate
- clip
- cut
- aspirate
- irrigate
- pack
- idle

Action definitions:
- grasp: closing an instrument's jaws on tissue or an object to hold it.
- retract: pulling or holding tissue aside to expose the operative field.
- dissect: separating tissue planes to expose anatomy.
- coagulate: applying energy for hemostasis.
- clip: applying a surgical clip to a duct or vessel.
- cut: dividing tissue or a structure with scissors.
- aspirate: suctioning fluid from the field.
- irrigate: flushing the field with fluid.
- pack: placing the specimen into a retrieval bag.
- idle: no listed action is being performed (instruments idle or absent).

Return your answer as a JSON object with this exact format:
{"current_actions": ["action_1", "action_2"]}

Use ONLY the exact action names from the list above.
List each action at most once. Multiple actions can occur simultaneously.
If no action is being performed, return: {"current_actions": ["idle"]}

Return ONLY valid JSON, nothing else.
\end{promptbox}

\begin{promptbox}
This frame is from a robot-assisted radical prostatectomy (RARP) suturing phase. Identify the current surgical action gesture being performed.

Valid actions:
- Other
- Picking Up The Needle
- Positioning The Needle Tip
- Pushing The Needle Through The Tissue
- Pulling The Needle Out Of The Tissue
- Tying A Knot
- Cutting The Suture
- Returning Or Dropping The Needle

Return your answer as a JSON object with this exact format:
{"action": "<one action from the list>"}

Use ONLY the exact action names from the list above.
Return ONLY valid JSON, nothing else.
\end{promptbox}

\subsection{Full results on the four additional tasks}

Tables~\ref{tab:pitvqa_new_ci}--\ref{tab:dsad_new_ci} report exact-match accuracy and micro-averaged F1 with 95\% bootstrap confidence intervals ($B = 1{,}000$) for every model class on each of the four datasets, sorted by micro-F1.

\begin{table}[H]
\centering
\caption{Surgical workflow (phase + step) recognition on the PitVQA validation split ($n=24{,}767$ frames from 5 held-out videos, 3 phases + 14 steps) with 95\% bootstrap confidence intervals ($B=1{,}000$). Exact match requires both the phase and the step to be correct. Output validation failures are counted as incorrect predictions. Best in bold. $^{\ast}$Frontier models were evaluated on a random seed-42 subsample of the validation split ($n=1{,}000$ frames).}
\label{tab:pitvqa_new_ci}
\footnotesize
\begin{tabular}{lrllll}
\toprule
Model & $n$ & EM \% & 95\% CI & Micro-F1 \% & 95\% CI \\
\midrule
ResNet-50 (supervised) & 24{,}767 & \textbf{66.42} & \textbf{65.87--67.03} & \textbf{77.05} & \textbf{76.63--77.49} \\
LemonFM (ConvNeXt-L) + linear probe & 24{,}767 & 63.27 & 62.66--63.83 & 76.09 & 75.67--76.49 \\
Gemma 3 27B + LoRA cls.\ head & 24{,}767 & 64.62 & 64.04--65.18 & 74.63 & 74.15--75.10 \\
Gemma 3 27B + LoRA (JSON) & 24{,}767 & 50.93 & 50.30--51.54 & 64.44 & 63.96--64.90 \\
GPT-5.6 Sol$^{\ast}$ & 1{,}000 & 32.10 & 29.30--35.20 & 50.30 & 47.80--53.00 \\
Claude Opus 5$^{\ast}$ & 1{,}000 & 30.60 & 27.70--33.40 & 46.60 & 44.10--49.20 \\
GPT-5.6 Terra$^{\ast}$ & 1{,}000 & 30.40 & 27.60--33.20 & 46.10 & 43.45--48.75 \\
GPT-5.6 Luna$^{\ast}$ & 1{,}000 & 29.90 & 27.10--32.70 & 44.90 & 42.30--47.45 \\
Claude Fable 5$^{\ast}$ & 1{,}000 & 18.10 & 15.90--20.40 & 43.60 & 41.30--45.75 \\
Gemma 3 27B-it (zero-shot) & 24{,}767 & 25.06 & 24.51--25.59 & 40.09 & 39.58--40.63 \\
Gemini 3.8 Flash$^{\ast}$ & 1{,}000 & 24.60 & 22.00--27.40 & 39.90 & 37.30--42.30 \\
Gemini 3.1 Pro$^{\ast}$ & 1{,}000 & 22.00 & 19.30--24.60 & 39.55 & 37.05--42.05 \\
Claude Sonnet 5$^{\ast}$ & 1{,}000 & 23.20 & 20.80--25.80 & 38.90 & 36.40--41.25 \\
Gemini 3.7 Flash$^{\ast}$ & 1{,}000 & 23.70 & 21.20--26.40 & 37.97 & 35.57--40.60 \\
Claude Fable 5.1$^{\ast}$ & 1{,}000 & 14.60 & 12.50--17.00 & 37.90 & 35.80--40.10 \\
Qwen3.8 27B (zero-shot) & 24{,}767 & 12.32 & 11.94--12.74 & 33.31 & 32.88--33.77 \\
\midrule
Majority class baseline & --- & 29.73 & --- & 39.37 & --- \\
\bottomrule
\end{tabular}
\end{table}

\begin{table}[H]
\centering
\caption{Surgical gesture recognition on the SAR-RARP50 validation split ($n=636$ segment-midpoint frames from 15 held-out operations, 8 gesture classes) with 95\% bootstrap confidence intervals ($B=1{,}000$). Single-label task: exact match and micro-F1 coincide up to parse failures. Best in bold.}
\label{tab:sarrarp50_new_ci}
\footnotesize
\begin{tabular}{lrllll}
\toprule
Model & $n$ & EM \% & 95\% CI & Micro-F1 \% & 95\% CI \\
\midrule
ResNet-50 (supervised) & 636 & \textbf{52.99} & \textbf{49.21--57.08} & \textbf{52.99} & \textbf{49.21--57.08} \\
LemonFM (ConvNeXt-L) + linear probe & 636 & 42.92 & 39.31--47.01 & 42.92 & 39.31--47.01 \\
Gemma 3 27B + LoRA cls.\ head & 636 & 35.22 & 31.76--39.15 & 35.22 & 31.76--39.15 \\
Claude Fable 5.1 & 636 & 30.00 & 26.70--33.50 & 30.00 & 26.70--33.50 \\
Claude Fable 5 & 636 & 27.20 & 23.90--30.66 & 27.20 & 23.90--30.66 \\
Claude Opus 5 & 636 & 24.84 & 21.38--28.46 & 24.84 & 21.38--28.46 \\
Gemini 3.1 Pro & 636 & 24.84 & 21.54--28.30 & 24.84 & 21.54--28.30 \\
Gemma 3 27B-it (zero-shot) & 636 & 24.21 & 20.91--27.52 & 24.21 & 20.91--27.52 \\
GPT-5.6 Sol & 636 & 23.90 & 20.60--27.36 & 23.90 & 20.60--27.36 \\
Qwen3.8 27B (zero-shot) & 636 & 23.27 & 20.28--26.57 & 23.29 & 20.30--26.57 \\
Gemma 3 27B + LoRA (JSON) & 636 & 22.48 & 19.34--25.79 & 22.52 & 19.40--25.79 \\
GPT-5.6 Luna & 636 & 22.33 & 19.34--25.47 & 22.33 & 19.34--25.47 \\
GPT-5.6 Terra & 636 & 21.86 & 18.87--25.31 & 21.86 & 18.87--25.31 \\
Gemini 3.8 Flash & 636 & 21.50 & 18.40--24.70 & 21.50 & 18.40--24.70 \\
Gemini 3.7 Flash & 636 & 21.23 & 18.24--24.53 & 21.24 & 18.24--24.55 \\
Claude Sonnet 5 & 636 & 18.55 & 15.25--21.86 & 18.55 & 15.25--21.86 \\
\midrule
Majority class baseline & --- & 25.63 & --- & 25.63 & --- \\
\bottomrule
\end{tabular}
\end{table}

\begin{table}[H]
\centering
\caption{Surgical action recognition on the CholecT50 validation split ($n=19{,}923$ frames from 10 held-out videos, 9 actions + idle, multi-label) with 95\% bootstrap confidence intervals ($B=1{,}000$). Labels are the verb component of the CholecT50 action-triplet annotations. Output validation failures are counted as incorrect predictions. Best in bold. $^{\ast}$Frontier models were evaluated on a random seed-42 subsample of the validation split ($n=1{,}000$ frames).}
\label{tab:cholect50verbs_new_ci}
\footnotesize
\begin{tabular}{lrllll}
\toprule
Model & $n$ & EM \% & 95\% CI & Micro-F1 \% & 95\% CI \\
\midrule
Gemma 3 27B + LoRA cls.\ head & 19{,}923 & \textbf{60.65} & \textbf{60.01--61.31} & \textbf{78.80} & \textbf{78.41--79.18} \\
Gemma 3 27B + LoRA (JSON) & 19{,}923 & 60.00 & 59.29--60.71 & 76.59 & 76.13--77.02 \\
ResNet-50 (supervised) & 19{,}923 & 52.94 & 52.26--53.61 & 75.19 & 74.75--75.60 \\
LemonFM (ConvNeXt-L) + linear probe & 19{,}923 & 49.76 & 49.09--50.42 & 72.94 & 72.48--73.36 \\
Gemini 3.8 Flash$^{\ast}$ & 1{,}000 & 39.80 & 36.80--42.70 & 70.10 & 68.30--72.00 \\
Gemini 3.7 Flash$^{\ast}$ & 1{,}000 & 32.20 & 29.50--35.10 & 68.81 & 67.10--70.50 \\
Gemini 3.1 Pro$^{\ast}$ & 1{,}000 & 37.80 & 34.60--40.80 & 68.54 & 66.48--70.33 \\
Claude Sonnet 5$^{\ast}$ & 1{,}000 & 45.40 & 42.30--48.50 & 66.47 & 64.19--68.81 \\
Claude Opus 5$^{\ast}$ & 1{,}000 & 40.20 & 37.00--43.20 & 62.41 & 60.19--64.73 \\
GPT-5.6 Terra$^{\ast}$ & 1{,}000 & 34.70 & 31.70--37.80 & 61.20 & 58.87--63.47 \\
GPT-5.6 Sol$^{\ast}$ & 1{,}000 & 30.40 & 27.70--33.60 & 59.68 & 57.63--61.91 \\
Claude Fable 5.1$^{\ast}$ & 1{,}000 & 31.30 & 28.50--34.20 & 58.30 & 56.10--60.60 \\
Claude Fable 5$^{\ast}$ & 1{,}000 & 35.70 & 32.80--38.70 & 57.16 & 54.62--59.55 \\
GPT-5.6 Luna$^{\ast}$ & 1{,}000 & 29.60 & 26.90--32.40 & 54.07 & 51.73--56.37 \\
Qwen3.8 27B (zero-shot) & 19{,}923 & 31.34 & 30.66--31.96 & 53.35 & 52.83--53.89 \\
Gemma 3 27B-it (zero-shot) & 19{,}923 & 1.54 & 1.37--1.71 & 24.93 & 24.54--25.32 \\
\midrule
Majority class baseline & --- & 18.46 & --- & 39.82 & --- \\
\bottomrule
\end{tabular}
\end{table}

\begin{table}[H]
\centering
\caption{Anatomy presence recognition on the DSAD validation split ($n=1{,}978$ frames, 12 anatomical structures, multi-label) with 95\% bootstrap confidence intervals ($B=1{,}000$). Output validation failures are counted as incorrect predictions. Best in bold. $^{\ast}$Frontier models were evaluated on a random seed-42 subsample of the validation split ($n=1{,}000$ frames).}
\label{tab:dsad_new_ci}
\footnotesize
\begin{tabular}{lrllll}
\toprule
Model & $n$ & EM \% & 95\% CI & Micro-F1 \% & 95\% CI \\
\midrule
ResNet-50 (supervised) & 1{,}978 & \textbf{30.59} & \textbf{28.51--32.56} & \textbf{71.63} & \textbf{70.51--72.79} \\
Gemma 3 27B + LoRA cls.\ head & 1{,}978 & 27.86 & 25.88--29.83 & 65.24 & 63.88--66.50 \\
LemonFM (ConvNeXt-L) + linear probe & 1{,}978 & 17.29 & 15.72--19.06 & 57.63 & 56.22--59.13 \\
Gemini 3.8 Flash$^{\ast}$ & 1{,}000 & 14.10 & 12.00--16.20 & 54.40 & 52.40--56.50 \\
Gemini 3.7 Flash$^{\ast}$ & 1{,}000 & 12.10 & 10.20--14.00 & 53.02 & 50.98--55.07 \\
Claude Fable 5.1$^{\ast}$ & 1{,}000 & 11.00 & 9.00--13.10 & 52.30 & 50.30--54.30 \\
Claude Fable 5$^{\ast}$ & 1{,}000 & 9.80 & 7.80--11.80 & 49.49 & 47.35--51.49 \\
Claude Opus 5$^{\ast}$ & 1{,}000 & 2.60 & 1.70--3.60 & 45.85 & 44.33--47.43 \\
Gemini 3.1 Pro$^{\ast}$ & 1{,}000 & 7.80 & 6.20--9.50 & 44.26 & 42.20--46.44 \\
GPT-5.6 Sol$^{\ast}$ & 1{,}000 & 5.20 & 3.80--6.60 & 43.75 & 41.85--45.57 \\
Gemma 3 27B + LoRA (JSON) & 1{,}978 & 10.82 & 9.50--12.18 & 41.59 & 40.06--43.08 \\
Gemma 3 27B-it (zero-shot) & 1{,}978 & 1.62 & 1.11--2.17 & 41.31 & 40.11--42.45 \\
GPT-5.6 Terra$^{\ast}$ & 1{,}000 & 5.10 & 3.90--6.50 & 40.53 & 38.63--42.45 \\
GPT-5.6 Luna$^{\ast}$ & 1{,}000 & 3.30 & 2.20--4.50 & 36.58 & 34.71--38.58 \\
Claude Sonnet 5$^{\ast}$ & 1{,}000 & 6.10 & 4.70--7.60 & 35.74 & 33.75--37.70 \\
Qwen3.8 27B (zero-shot) & 1{,}978 & 2.28 & 1.62--2.93 & 30.97 & 29.74--32.29 \\
\midrule
Majority class baseline & --- & 1.42 & --- & 29.65 & --- \\
\bottomrule
\end{tabular}
\end{table}

%% file: sections/appendix_zero_shot_tables.tex
\section{Per-Tool Metrics for Zero-Shot VLM Evaluation}
\label{sec:appendix_zero_shot_metrics}

The following tables present per-tool classification metrics for each zero-shot vision-language model evaluated on our surgical tool detection benchmark ($n=67{,}716$ frames for most models). Only tools with at least one true positive, false positive, or false negative are shown. Tools are sorted by F1 score in descending order.

\begin{table}[H]
\centering
\caption{Per-tool metrics for Qwen3-VL-235B-A22B-Thinking (235B) zero-shot evaluation. P = precision, R = recall.}
\label{tab:zero_shot_qwen3_vl_235b_a22b_thinking}
\scriptsize
\begin{tabular}{lrrrrrr}
\toprule
Tool & TP & FP & FN & P & R & F1 \\
\midrule
Suction & 28745 & 12845 & 10868 & 0.691 & 0.726 & 0.708 \\
Cotton Patty & 2329 & 387 & 8498 & 0.858 & 0.215 & 0.344 \\
Sonopet pineapple tip & 367 & 356 & 1624 & 0.508 & 0.184 & 0.270 \\
Grasper & 2216 & 8314 & 5106 & 0.210 & 0.303 & 0.248 \\
Drill & 467 & 260 & 2930 & 0.642 & 0.137 & 0.226 \\
Rongeur & 593 & 4131 & 3577 & 0.126 & 0.142 & 0.133 \\
Straight Forceps & 292 & 3286 & 1403 & 0.082 & 0.172 & 0.111 \\
Bipolar Forceps & 110 & 848 & 2087 & 0.115 & 0.050 & 0.070 \\
Rhoton Dissector & 193 & 883 & 5520 & 0.179 & 0.034 & 0.057 \\
Suction Coagulator & 88 & 1003 & 2298 & 0.081 & 0.037 & 0.051 \\
Floseal Gel & 13 & 115 & 514 & 0.102 & 0.025 & 0.040 \\
Surgicel & 36 & 512 & 3551 & 0.066 & 0.010 & 0.017 \\
Backbiting rongeur & 11 & 256 & 1174 & 0.041 & 0.009 & 0.015 \\
Scissor & 30 & 225 & 3895 & 0.118 & 0.008 & 0.014 \\
Floseal Applicator & 2 & 115 & 214 & 0.017 & 0.009 & 0.012 \\
Suction microdebrider & 16 & 2728 & 575 & 0.006 & 0.027 & 0.010 \\
Aspirating dissector & 12 & 407 & 2369 & 0.029 & 0.005 & 0.009 \\
Irrigation & 2 & 175 & 384 & 0.011 & 0.005 & 0.007 \\
Cottle Elevator & 3 & 158 & 852 & 0.019 & 0.004 & 0.006 \\
Surgical Knife & 7 & 17 & 2780 & 0.292 & 0.003 & 0.005 \\
Through cutting forceps & 1 & 108 & 796 & 0.009 & 0.001 & 0.002 \\
Curette & 1 & 4 & 5538 & 0.200 & 0.000 & 0.000 \\
Collagen Matrix & 0 & 206 & 280 & 0.000 & 0.000 & 0.000 \\
Doppler & 0 & 32 & 1482 & 0.000 & 0.000 & 0.000 \\
Local Anesthesia Needle & 0 & 22 & 139 & 0.000 & 0.000 & 0.000 \\
Monopolar Electrocautery & 0 & 18 & 190 & 0.000 & 0.000 & 0.000 \\
Needle & 0 & 117 & 31 & 0.000 & 0.000 & 0.000 \\
Straight Curette & 0 & 2 & 102 & 0.000 & 0.000 & 0.000 \\
Tisseel Applicator & 0 & 8 & 23 & 0.000 & 0.000 & 0.000 \\
Tissue shaver & 0 & 9 & 991 & 0.000 & 0.000 & 0.000 \\
unknown & 0 & 0 & 1195 & 0.000 & 0.000 & 0.000 \\
\bottomrule
\end{tabular}
\end{table}

\begin{table}[H]
\centering
\caption{Per-tool metrics for Qwen3-VL-32B-Instruct (32B) zero-shot evaluation. P = precision, R = recall.}
\label{tab:zero_shot_qwen3_vl_32b_instruct}
\scriptsize
\begin{tabular}{lrrrrrr}
\toprule
Tool & TP & FP & FN & P & R & F1 \\
\midrule
Suction & 33117 & 12849 & 7248 & 0.721 & 0.820 & 0.767 \\
Grasper & 1376 & 2617 & 5943 & 0.345 & 0.188 & 0.243 \\
Cotton Patty & 1077 & 144 & 9887 & 0.882 & 0.098 & 0.177 \\
Rhoton Dissector & 1298 & 9156 & 4414 & 0.124 & 0.227 & 0.161 \\
Bipolar Forceps & 387 & 2891 & 1810 & 0.118 & 0.176 & 0.141 \\
Rongeur & 502 & 2101 & 4016 & 0.193 & 0.111 & 0.141 \\
Floseal Gel & 85 & 1103 & 457 & 0.071 & 0.157 & 0.098 \\
Curette & 351 & 1939 & 5186 & 0.153 & 0.063 & 0.090 \\
Surgical Knife & 241 & 3422 & 2544 & 0.066 & 0.086 & 0.075 \\
Backbiting rongeur & 63 & 607 & 1120 & 0.094 & 0.053 & 0.068 \\
Surgicel & 140 & 1384 & 3446 & 0.092 & 0.039 & 0.055 \\
Aspirating dissector & 146 & 3753 & 2260 & 0.037 & 0.061 & 0.046 \\
Sonopet pineapple tip & 44 & 37 & 1947 & 0.543 & 0.022 & 0.043 \\
Floseal Applicator & 14 & 685 & 225 & 0.020 & 0.059 & 0.030 \\
Scissor & 61 & 266 & 3862 & 0.186 & 0.015 & 0.029 \\
Suction Coagulator & 93 & 4345 & 2495 & 0.021 & 0.036 & 0.026 \\
Drill & 37 & 190 & 3716 & 0.163 & 0.010 & 0.019 \\
Straight Forceps & 18 & 564 & 1756 & 0.031 & 0.010 & 0.015 \\
Monopolar Electrocautery & 5 & 1173 & 185 & 0.004 & 0.026 & 0.007 \\
Through cutting forceps & 2 & 32 & 794 & 0.059 & 0.003 & 0.005 \\
Irrigation & 1 & 121 & 385 & 0.008 & 0.003 & 0.004 \\
Collagen Matrix & 0 & 886 & 280 & 0.000 & 0.000 & 0.000 \\
Cottle Elevator & 0 & 0 & 855 & 0.000 & 0.000 & 0.000 \\
Doppler & 0 & 0 & 1591 & 0.000 & 0.000 & 0.000 \\
Local Anesthesia Needle & 0 & 5 & 139 & 0.000 & 0.000 & 0.000 \\
Needle & 0 & 107 & 31 & 0.000 & 0.000 & 0.000 \\
Straight Curette & 0 & 322 & 118 & 0.000 & 0.000 & 0.000 \\
Suction microdebrider & 0 & 1359 & 590 & 0.000 & 0.000 & 0.000 \\
Tisseel Applicator & 0 & 198 & 23 & 0.000 & 0.000 & 0.000 \\
Tissue shaver & 0 & 24 & 991 & 0.000 & 0.000 & 0.000 \\
unknown & 0 & 0 & 1195 & 0.000 & 0.000 & 0.000 \\
\bottomrule
\end{tabular}
\end{table}

\begin{table}[H]
\centering
\caption{Per-tool metrics for Qwen3-VL-8B-Instruct (8B) zero-shot evaluation. P = precision, R = recall.}
\label{tab:zero_shot_qwen3_vl_8b_instruct}
\scriptsize
\begin{tabular}{lrrrrrr}
\toprule
Tool & TP & FP & FN & P & R & F1 \\
\midrule
Suction & 15594 & 4603 & 24783 & 0.772 & 0.386 & 0.515 \\
Grasper & 1587 & 3204 & 5734 & 0.331 & 0.217 & 0.262 \\
Surgical Knife & 228 & 886 & 2558 & 0.205 & 0.082 & 0.117 \\
Cotton Patty & 649 & 35 & 10316 & 0.949 & 0.059 & 0.111 \\
Rhoton Dissector & 522 & 4399 & 5189 & 0.106 & 0.091 & 0.098 \\
Aspirating dissector & 842 & 19376 & 1565 & 0.042 & 0.350 & 0.074 \\
Drill & 116 & 115 & 3639 & 0.502 & 0.031 & 0.058 \\
Rongeur & 135 & 1045 & 4385 & 0.114 & 0.030 & 0.047 \\
Floseal Applicator & 7 & 263 & 232 & 0.026 & 0.029 & 0.028 \\
Irrigation & 14 & 660 & 372 & 0.021 & 0.036 & 0.026 \\
Scissor & 46 & 241 & 3878 & 0.160 & 0.012 & 0.022 \\
Surgicel & 34 & 335 & 3552 & 0.092 & 0.009 & 0.017 \\
Floseal Gel & 5 & 68 & 537 & 0.069 & 0.009 & 0.016 \\
Suction Coagulator & 22 & 1047 & 2566 & 0.021 & 0.009 & 0.012 \\
Bipolar Forceps & 10 & 171 & 2187 & 0.055 & 0.005 & 0.008 \\
Tissue shaver & 5 & 340 & 986 & 0.015 & 0.005 & 0.007 \\
Curette & 21 & 132 & 5516 & 0.137 & 0.004 & 0.007 \\
Monopolar Electrocautery & 1 & 361 & 189 & 0.003 & 0.005 & 0.004 \\
Straight Forceps & 3 & 12 & 1772 & 0.200 & 0.002 & 0.003 \\
Backbiting rongeur & 1 & 1 & 1184 & 0.500 & 0.001 & 0.002 \\
Suction microdebrider & 5 & 7259 & 586 & 0.001 & 0.009 & 0.001 \\
Collagen Matrix & 0 & 54 & 280 & 0.000 & 0.000 & 0.000 \\
Cottle Elevator & 0 & 23 & 855 & 0.000 & 0.000 & 0.000 \\
Doppler & 0 & 0 & 1592 & 0.000 & 0.000 & 0.000 \\
Local Anesthesia Needle & 0 & 7 & 139 & 0.000 & 0.000 & 0.000 \\
Needle & 0 & 21 & 31 & 0.000 & 0.000 & 0.000 \\
Sonopet pineapple tip & 0 & 20 & 1991 & 0.000 & 0.000 & 0.000 \\
Straight Curette & 0 & 34 & 118 & 0.000 & 0.000 & 0.000 \\
Through cutting forceps & 0 & 3 & 797 & 0.000 & 0.000 & 0.000 \\
Tisseel Applicator & 0 & 6 & 23 & 0.000 & 0.000 & 0.000 \\
unknown & 0 & 2 & 1195 & 0.000 & 0.000 & 0.000 \\
\bottomrule
\end{tabular}
\end{table}

\begin{table}[H]
\centering
\caption{Per-tool metrics for Qwen3-VL-4B-Instruct (4B) zero-shot evaluation. P = precision, R = recall.}
\label{tab:zero_shot_qwen3_vl_4b_instruct}
\scriptsize
\begin{tabular}{lrrrrrr}
\toprule
Tool & TP & FP & FN & P & R & F1 \\
\midrule
Suction & 26189 & 10548 & 14026 & 0.713 & 0.651 & 0.681 \\
Drill & 1126 & 402 & 2615 & 0.737 & 0.301 & 0.427 \\
Grasper & 4922 & 24241 & 2368 & 0.169 & 0.675 & 0.270 \\
Cotton Patty & 1297 & 120 & 9615 & 0.915 & 0.119 & 0.210 \\
Scissor & 2016 & 18097 & 1886 & 0.100 & 0.517 & 0.168 \\
Surgical Knife & 251 & 3478 & 2527 & 0.067 & 0.090 & 0.077 \\
Curette & 290 & 3419 & 5224 & 0.078 & 0.053 & 0.063 \\
Aspirating dissector & 90 & 566 & 2309 & 0.137 & 0.037 & 0.059 \\
Suction Coagulator & 271 & 9345 & 2311 & 0.028 & 0.105 & 0.044 \\
Irrigation & 7 & 368 & 378 & 0.019 & 0.018 & 0.018 \\
Local Anesthesia Needle & 1 & 52 & 138 & 0.019 & 0.007 & 0.010 \\
Through cutting forceps & 5 & 167 & 789 & 0.029 & 0.006 & 0.010 \\
Straight Curette & 1 & 127 & 114 & 0.008 & 0.009 & 0.008 \\
Surgicel & 14 & 172 & 3563 & 0.075 & 0.004 & 0.007 \\
Floseal Gel & 2 & 57 & 539 & 0.034 & 0.004 & 0.007 \\
Monopolar Electrocautery & 1 & 110 & 189 & 0.009 & 0.005 & 0.007 \\
Floseal Applicator & 1 & 68 & 238 & 0.015 & 0.004 & 0.006 \\
unknown & 5 & 355 & 1187 & 0.014 & 0.004 & 0.006 \\
Rongeur & 8 & 230 & 4487 & 0.034 & 0.002 & 0.003 \\
Backbiting rongeur & 2 & 30 & 1179 & 0.062 & 0.002 & 0.003 \\
Straight Forceps & 3 & 132 & 1766 & 0.022 & 0.002 & 0.003 \\
Rhoton Dissector & 3 & 24 & 5681 & 0.111 & 0.001 & 0.001 \\
Sonopet pineapple tip & 1 & 23 & 1986 & 0.042 & 0.001 & 0.001 \\
Bipolar Forceps & 0 & 25 & 2194 & 0.000 & 0.000 & 0.000 \\
Collagen Matrix & 0 & 172 & 280 & 0.000 & 0.000 & 0.000 \\
Cottle Elevator & 0 & 84 & 855 & 0.000 & 0.000 & 0.000 \\
Doppler & 0 & 32 & 1587 & 0.000 & 0.000 & 0.000 \\
Needle & 0 & 59 & 30 & 0.000 & 0.000 & 0.000 \\
Suction microdebrider & 0 & 214 & 591 & 0.000 & 0.000 & 0.000 \\
Tisseel Applicator & 0 & 71 & 23 & 0.000 & 0.000 & 0.000 \\
Tissue shaver & 0 & 224 & 990 & 0.000 & 0.000 & 0.000 \\
\bottomrule
\end{tabular}
\end{table}

\begin{table}[H]
\centering
\caption{Per-tool metrics for Qwen3-VL-2B-Instruct (2B) zero-shot evaluation. P = precision, R = recall.}
\label{tab:zero_shot_qwen3_vl_2b_instruct}
\scriptsize
\begin{tabular}{lrrrrrr}
\toprule
Tool & TP & FP & FN & P & R & F1 \\
\midrule
Suction & 24482 & 14568 & 3503 & 0.627 & 0.875 & 0.730 \\
Grasper & 4424 & 30813 & 958 & 0.126 & 0.822 & 0.218 \\
Drill & 581 & 2309 & 2010 & 0.201 & 0.224 & 0.212 \\
Rongeur & 1472 & 19535 & 1906 & 0.070 & 0.436 & 0.121 \\
Cotton Patty & 603 & 2503 & 7108 & 0.194 & 0.078 & 0.112 \\
Scissor & 1236 & 18066 & 1624 & 0.064 & 0.432 & 0.112 \\
Curette & 710 & 9347 & 3243 & 0.071 & 0.180 & 0.101 \\
Rhoton Dissector & 481 & 5552 & 3599 & 0.080 & 0.118 & 0.095 \\
Surgical Knife & 2110 & 43265 & 4 & 0.046 & 0.998 & 0.089 \\
Suction Coagulator & 732 & 18203 & 1053 & 0.039 & 0.410 & 0.071 \\
Straight Forceps & 476 & 16156 & 792 & 0.029 & 0.375 & 0.053 \\
Bipolar Forceps & 100 & 2770 & 1407 & 0.035 & 0.066 & 0.046 \\
Aspirating dissector & 79 & 1986 & 1433 & 0.038 & 0.052 & 0.044 \\
Doppler & 47 & 1623 & 991 & 0.028 & 0.045 & 0.035 \\
Tissue shaver & 269 & 15834 & 307 & 0.017 & 0.467 & 0.032 \\
Sonopet pineapple tip & 19 & 351 & 1152 & 0.051 & 0.016 & 0.025 \\
Cottle Elevator & 240 & 19232 & 334 & 0.012 & 0.418 & 0.024 \\
Through cutting forceps & 156 & 13162 & 427 & 0.012 & 0.268 & 0.022 \\
unknown & 16 & 690 & 716 & 0.023 & 0.022 & 0.022 \\
Surgicel & 28 & 464 & 2436 & 0.057 & 0.011 & 0.019 \\
Backbiting rongeur & 19 & 1345 & 911 & 0.014 & 0.020 & 0.017 \\
Floseal Gel & 58 & 7239 & 304 & 0.008 & 0.160 & 0.015 \\
Suction microdebrider & 214 & 29996 & 72 & 0.007 & 0.748 & 0.014 \\
Irrigation & 35 & 5748 & 247 & 0.006 & 0.124 & 0.011 \\
Monopolar Electrocautery & 29 & 8210 & 99 & 0.004 & 0.227 & 0.007 \\
Collagen Matrix & 5 & 1746 & 213 & 0.003 & 0.023 & 0.005 \\
Floseal Applicator & 18 & 7161 & 148 & 0.003 & 0.108 & 0.005 \\
Local Anesthesia Needle & 14 & 6382 & 93 & 0.002 & 0.131 & 0.004 \\
Straight Curette & 8 & 5957 & 69 & 0.001 & 0.104 & 0.003 \\
Needle & 2 & 1901 & 22 & 0.001 & 0.083 & 0.002 \\
Tisseel Applicator & 1 & 14970 & 12 & 0.000 & 0.077 & 0.000 \\
\bottomrule
\end{tabular}
\end{table}

\begin{table}[H]
\centering
\caption{Per-tool metrics for Qwen2.5-VL-72B-Instruct (72B) zero-shot evaluation. P = precision, R = recall.}
\label{tab:zero_shot_qwen2_5_vl_72b_instruct}
\scriptsize
\begin{tabular}{lrrrrrr}
\toprule
Tool & TP & FP & FN & P & R & F1 \\
\midrule
Suction & 10304 & 3016 & 29750 & 0.774 & 0.257 & 0.386 \\
Sonopet pineapple tip & 174 & 272 & 1806 & 0.390 & 0.088 & 0.143 \\
Grasper & 472 & 1330 & 6784 & 0.262 & 0.065 & 0.104 \\
Cotton Patty & 391 & 285 & 10473 & 0.578 & 0.036 & 0.068 \\
Rongeur & 156 & 1070 & 4308 & 0.127 & 0.035 & 0.055 \\
Scissor & 107 & 585 & 3779 & 0.155 & 0.028 & 0.047 \\
Curette & 118 & 902 & 5378 & 0.116 & 0.021 & 0.036 \\
unknown & 92 & 3878 & 1098 & 0.023 & 0.077 & 0.036 \\
Surgical Knife & 61 & 622 & 2687 & 0.089 & 0.022 & 0.036 \\
Surgicel & 68 & 529 & 3497 & 0.114 & 0.019 & 0.033 \\
Rhoton Dissector & 106 & 1226 & 5571 & 0.080 & 0.019 & 0.030 \\
Straight Forceps & 36 & 1250 & 1716 & 0.028 & 0.021 & 0.024 \\
Backbiting rongeur & 14 & 275 & 1153 & 0.048 & 0.012 & 0.019 \\
Drill & 36 & 93 & 3694 & 0.279 & 0.010 & 0.019 \\
Bipolar Forceps & 20 & 329 & 2160 & 0.057 & 0.009 & 0.016 \\
Irrigation & 4 & 406 & 377 & 0.010 & 0.011 & 0.010 \\
Aspirating dissector & 8 & 419 & 2385 & 0.019 & 0.003 & 0.006 \\
Cottle Elevator & 2 & 254 & 850 & 0.008 & 0.002 & 0.004 \\
Floseal Gel & 1 & 63 & 533 & 0.016 & 0.002 & 0.003 \\
Through cutting forceps & 1 & 66 & 789 & 0.015 & 0.001 & 0.002 \\
Collagen Matrix & 1 & 658 & 279 & 0.002 & 0.004 & 0.002 \\
Monopolar Electrocautery & 1 & 1189 & 187 & 0.001 & 0.005 & 0.002 \\
Suction Coagulator & 1 & 94 & 2567 & 0.011 & 0.000 & 0.001 \\
Doppler & 0 & 17 & 1576 & 0.000 & 0.000 & 0.000 \\
Floseal Applicator & 0 & 142 & 236 & 0.000 & 0.000 & 0.000 \\
Local Anesthesia Needle & 0 & 7 & 138 & 0.000 & 0.000 & 0.000 \\
Needle & 0 & 6 & 30 & 0.000 & 0.000 & 0.000 \\
Straight Curette & 0 & 229 & 117 & 0.000 & 0.000 & 0.000 \\
Suction microdebrider & 0 & 64 & 591 & 0.000 & 0.000 & 0.000 \\
Tisseel Applicator & 0 & 38 & 22 & 0.000 & 0.000 & 0.000 \\
Tissue shaver & 0 & 139 & 985 & 0.000 & 0.000 & 0.000 \\
\bottomrule
\end{tabular}
\end{table}

\begin{table}[H]
\centering
\caption{Per-tool metrics for Qwen2.5-VL-32B-Instruct (32B) zero-shot evaluation. P = precision, R = recall.}
\label{tab:zero_shot_qwen2_5_vl_32b_instruct}
\scriptsize
\begin{tabular}{lrrrrrr}
\toprule
Tool & TP & FP & FN & P & R & F1 \\
\midrule
Suction & 10990 & 4861 & 27100 & 0.693 & 0.288 & 0.407 \\
Cotton Patty & 929 & 518 & 9475 & 0.642 & 0.089 & 0.157 \\
Scissor & 269 & 2041 & 3436 & 0.117 & 0.073 & 0.089 \\
Bipolar Forceps & 236 & 3079 & 1867 & 0.071 & 0.112 & 0.087 \\
Sonopet pineapple tip & 77 & 50 & 1787 & 0.606 & 0.041 & 0.077 \\
Curette & 352 & 3950 & 4847 & 0.082 & 0.068 & 0.074 \\
Surgical Knife & 157 & 1770 & 2514 & 0.082 & 0.059 & 0.068 \\
Rhoton Dissector & 260 & 2364 & 5125 & 0.099 & 0.048 & 0.065 \\
Rongeur & 208 & 1972 & 4069 & 0.095 & 0.049 & 0.064 \\
Grasper & 230 & 948 & 6702 & 0.195 & 0.033 & 0.057 \\
Drill & 56 & 102 & 3486 & 0.354 & 0.016 & 0.030 \\
Straight Forceps & 29 & 864 & 1637 & 0.033 & 0.017 & 0.023 \\
Through cutting forceps & 10 & 307 & 746 & 0.032 & 0.013 & 0.019 \\
Aspirating dissector & 24 & 1249 & 2258 & 0.019 & 0.011 & 0.013 \\
Floseal Gel & 4 & 76 & 512 & 0.050 & 0.008 & 0.013 \\
Irrigation & 24 & 3707 & 332 & 0.006 & 0.067 & 0.012 \\
Backbiting rongeur & 6 & 115 & 1110 & 0.050 & 0.005 & 0.010 \\
Floseal Applicator & 2 & 230 & 225 & 0.009 & 0.009 & 0.009 \\
Monopolar Electrocautery & 9 & 3185 & 172 & 0.003 & 0.050 & 0.005 \\
Local Anesthesia Needle & 1 & 294 & 135 & 0.003 & 0.007 & 0.005 \\
Suction Coagulator & 5 & 223 & 2431 & 0.022 & 0.002 & 0.004 \\
Tissue shaver & 2 & 257 & 912 & 0.008 & 0.002 & 0.003 \\
unknown & 2 & 92 & 1109 & 0.021 & 0.002 & 0.003 \\
Doppler & 1 & 73 & 1514 & 0.013 & 0.001 & 0.001 \\
Surgicel & 2 & 22 & 3371 & 0.083 & 0.001 & 0.001 \\
Collagen Matrix & 0 & 166 & 263 & 0.000 & 0.000 & 0.000 \\
Cottle Elevator & 0 & 89 & 812 & 0.000 & 0.000 & 0.000 \\
Needle & 0 & 307 & 27 & 0.000 & 0.000 & 0.000 \\
Straight Curette & 0 & 607 & 103 & 0.000 & 0.000 & 0.000 \\
Suction microdebrider & 0 & 55 & 556 & 0.000 & 0.000 & 0.000 \\
Tisseel Applicator & 0 & 55 & 21 & 0.000 & 0.000 & 0.000 \\
\bottomrule
\end{tabular}
\end{table}

\begin{table}[H]
\centering
\caption{Per-tool metrics for Qwen2.5-VL-7B-Instruct (7B) zero-shot evaluation. P = precision, R = recall.}
\label{tab:zero_shot_qwen2_5_vl_7b_instruct}
\scriptsize
\begin{tabular}{lrrrrrr}
\toprule
Tool & TP & FP & FN & P & R & F1 \\
\midrule
Suction & 2207 & 1031 & 32342 & 0.682 & 0.064 & 0.117 \\
Grasper & 480 & 1506 & 5921 & 0.242 & 0.075 & 0.115 \\
Surgical Knife & 799 & 11805 & 1707 & 0.063 & 0.319 & 0.106 \\
Curette & 399 & 2817 & 4341 & 0.124 & 0.084 & 0.100 \\
Scissor & 185 & 1476 & 3252 & 0.111 & 0.054 & 0.073 \\
Rhoton Dissector & 192 & 1678 & 4718 & 0.103 & 0.039 & 0.057 \\
Suction Coagulator & 113 & 1792 & 2083 & 0.059 & 0.051 & 0.055 \\
Backbiting rongeur & 43 & 632 & 982 & 0.064 & 0.042 & 0.051 \\
Rongeur & 112 & 890 & 3864 & 0.112 & 0.028 & 0.045 \\
Bipolar Forceps & 59 & 998 & 1827 & 0.056 & 0.031 & 0.040 \\
Drill & 56 & 334 & 3193 & 0.144 & 0.017 & 0.031 \\
unknown & 19 & 462 & 956 & 0.040 & 0.019 & 0.026 \\
Aspirating dissector & 55 & 2346 & 1777 & 0.023 & 0.030 & 0.026 \\
Straight Forceps & 26 & 789 & 1465 & 0.032 & 0.017 & 0.022 \\
Doppler & 15 & 399 & 1404 & 0.036 & 0.011 & 0.016 \\
Tissue shaver & 11 & 1071 & 779 & 0.010 & 0.014 & 0.012 \\
Cottle Elevator & 10 & 1200 & 680 & 0.008 & 0.015 & 0.011 \\
Cotton Patty & 40 & 60 & 9386 & 0.400 & 0.004 & 0.008 \\
Sonopet pineapple tip & 8 & 251 & 1634 & 0.031 & 0.005 & 0.008 \\
Local Anesthesia Needle & 1 & 156 & 119 & 0.006 & 0.008 & 0.007 \\
Floseal Applicator & 3 & 652 & 199 & 0.005 & 0.015 & 0.007 \\
Floseal Gel & 2 & 110 & 457 & 0.018 & 0.004 & 0.007 \\
Through cutting forceps & 3 & 220 & 674 & 0.013 & 0.004 & 0.007 \\
Surgicel & 9 & 211 & 3020 & 0.041 & 0.003 & 0.005 \\
Monopolar Electrocautery & 4 & 1302 & 159 & 0.003 & 0.025 & 0.005 \\
Irrigation & 1 & 271 & 338 & 0.004 & 0.003 & 0.003 \\
Collagen Matrix & 0 & 110 & 245 & 0.000 & 0.000 & 0.000 \\
Needle & 0 & 743 & 26 & 0.000 & 0.000 & 0.000 \\
Straight Curette & 0 & 762 & 102 & 0.000 & 0.000 & 0.000 \\
Suction microdebrider & 0 & 109 & 435 & 0.000 & 0.000 & 0.000 \\
Tisseel Applicator & 0 & 99 & 20 & 0.000 & 0.000 & 0.000 \\
\bottomrule
\end{tabular}
\end{table}

\begin{table}[H]
\centering
\caption{Per-tool metrics for Qwen2.5-VL-3B-Instruct (3B) zero-shot evaluation. P = precision, R = recall.}
\label{tab:zero_shot_qwen2_5_vl_3b_instruct}
\scriptsize
\begin{tabular}{lrrrrrr}
\toprule
Tool & TP & FP & FN & P & R & F1 \\
\midrule
Drill & 573 & 1106 & 2447 & 0.341 & 0.190 & 0.244 \\
Surgical Knife & 434 & 5197 & 1854 & 0.077 & 0.190 & 0.110 \\
Curette & 498 & 4650 & 3916 & 0.097 & 0.113 & 0.104 \\
Suction & 1650 & 660 & 30778 & 0.714 & 0.051 & 0.095 \\
Rongeur & 276 & 2926 & 3392 & 0.086 & 0.075 & 0.080 \\
Straight Forceps & 175 & 4054 & 1253 & 0.041 & 0.122 & 0.062 \\
Grasper & 157 & 783 & 5797 & 0.167 & 0.026 & 0.045 \\
Bipolar Forceps & 55 & 686 & 1775 & 0.074 & 0.030 & 0.043 \\
Local Anesthesia Needle & 3 & 86 & 99 & 0.034 & 0.029 & 0.031 \\
Rhoton Dissector & 51 & 421 & 4556 & 0.108 & 0.011 & 0.020 \\
Through cutting forceps & 29 & 2458 & 587 & 0.012 & 0.047 & 0.019 \\
Aspirating dissector & 23 & 877 & 1711 & 0.026 & 0.013 & 0.018 \\
Scissor & 30 & 239 & 3182 & 0.112 & 0.009 & 0.017 \\
Cotton Patty & 62 & 139 & 8760 & 0.308 & 0.007 & 0.014 \\
Surgicel & 19 & 214 & 2894 & 0.082 & 0.006 & 0.012 \\
Collagen Matrix & 3 & 308 & 233 & 0.010 & 0.013 & 0.011 \\
Irrigation & 4 & 466 & 308 & 0.009 & 0.013 & 0.010 \\
Floseal Gel & 3 & 179 & 429 & 0.017 & 0.007 & 0.010 \\
Sonopet pineapple tip & 6 & 167 & 1559 & 0.035 & 0.004 & 0.007 \\
Suction Coagulator & 6 & 162 & 1972 & 0.036 & 0.003 & 0.006 \\
Backbiting rongeur & 4 & 544 & 923 & 0.007 & 0.004 & 0.005 \\
Tissue shaver & 3 & 429 & 744 & 0.007 & 0.004 & 0.005 \\
Monopolar Electrocautery & 3 & 1070 & 133 & 0.003 & 0.022 & 0.005 \\
Cottle Elevator & 2 & 216 & 656 & 0.009 & 0.003 & 0.005 \\
Straight Curette & 7 & 3403 & 86 & 0.002 & 0.075 & 0.004 \\
Floseal Applicator & 1 & 541 & 189 & 0.002 & 0.005 & 0.003 \\
Doppler & 2 & 236 & 1277 & 0.008 & 0.002 & 0.003 \\
Suction microdebrider & 1 & 498 & 424 & 0.002 & 0.002 & 0.002 \\
Needle & 0 & 42 & 25 & 0.000 & 0.000 & 0.000 \\
Tisseel Applicator & 0 & 22 & 18 & 0.000 & 0.000 & 0.000 \\
unknown & 0 & 67 & 916 & 0.000 & 0.000 & 0.000 \\
\bottomrule
\end{tabular}
\end{table}

\begin{table}[H]
\centering
\caption{Per-tool metrics for Qwen2-VL-72B-Instruct (72B) zero-shot evaluation. P = precision, R = recall.}
\label{tab:zero_shot_qwen2_vl_72b_instruct}
\scriptsize
\begin{tabular}{lrrrrrr}
\toprule
Tool & TP & FP & FN & P & R & F1 \\
\midrule
Suction & 9085 & 4437 & 29320 & 0.672 & 0.237 & 0.350 \\
Drill & 458 & 907 & 3064 & 0.336 & 0.130 & 0.187 \\
Grasper & 1069 & 4589 & 5896 & 0.189 & 0.153 & 0.169 \\
Curette & 401 & 3817 & 4881 & 0.095 & 0.076 & 0.084 \\
Bipolar Forceps & 191 & 2473 & 1893 & 0.072 & 0.092 & 0.081 \\
Scissor & 219 & 1650 & 3522 & 0.117 & 0.059 & 0.078 \\
Rhoton Dissector & 327 & 3356 & 5129 & 0.089 & 0.060 & 0.072 \\
Sonopet pineapple tip & 107 & 1039 & 1760 & 0.093 & 0.057 & 0.071 \\
Rongeur & 208 & 2034 & 4098 & 0.093 & 0.048 & 0.064 \\
Surgical Knife & 91 & 1493 & 2540 & 0.057 & 0.035 & 0.043 \\
Straight Forceps & 64 & 1818 & 1632 & 0.034 & 0.038 & 0.036 \\
Backbiting rongeur & 52 & 2098 & 1078 & 0.024 & 0.046 & 0.032 \\
Surgicel & 59 & 334 & 3333 & 0.150 & 0.017 & 0.031 \\
unknown & 35 & 1182 & 1117 & 0.029 & 0.030 & 0.029 \\
Aspirating dissector & 75 & 2785 & 2245 & 0.026 & 0.032 & 0.029 \\
Cotton Patty & 146 & 70 & 10294 & 0.676 & 0.014 & 0.027 \\
Suction Coagulator & 52 & 1361 & 2447 & 0.037 & 0.021 & 0.027 \\
Floseal Gel & 10 & 509 & 494 & 0.019 & 0.020 & 0.020 \\
Irrigation & 19 & 2964 & 345 & 0.006 & 0.052 & 0.011 \\
Local Anesthesia Needle & 1 & 74 & 133 & 0.013 & 0.007 & 0.010 \\
Doppler & 6 & 135 & 1518 & 0.043 & 0.004 & 0.007 \\
Tissue shaver & 5 & 719 & 946 & 0.007 & 0.005 & 0.006 \\
Cottle Elevator & 3 & 281 & 807 & 0.011 & 0.004 & 0.005 \\
Through cutting forceps & 3 & 369 & 758 & 0.008 & 0.004 & 0.005 \\
Suction microdebrider & 7 & 2096 & 566 & 0.003 & 0.012 & 0.005 \\
Collagen Matrix & 3 & 979 & 266 & 0.003 & 0.011 & 0.005 \\
Monopolar Electrocautery & 7 & 2923 & 177 & 0.002 & 0.038 & 0.004 \\
Floseal Applicator & 1 & 436 & 225 & 0.002 & 0.004 & 0.003 \\
Straight Curette & 3 & 1897 & 111 & 0.002 & 0.026 & 0.003 \\
Needle & 0 & 9 & 31 & 0.000 & 0.000 & 0.000 \\
Tisseel Applicator & 0 & 83 & 21 & 0.000 & 0.000 & 0.000 \\
\bottomrule
\end{tabular}
\end{table}

\begin{table}[H]
\centering
\caption{Per-tool metrics for Qwen2-VL-7B-Instruct (7B) zero-shot evaluation. P = precision, R = recall.}
\label{tab:zero_shot_qwen2_vl_7b_instruct}
\scriptsize
\begin{tabular}{lrrrrrr}
\toprule
Tool & TP & FP & FN & P & R & F1 \\
\midrule
Suction & 14191 & 7274 & 15552 & 0.661 & 0.477 & 0.554 \\
Drill & 1691 & 5743 & 1337 & 0.228 & 0.558 & 0.323 \\
Curette & 948 & 8601 & 3154 & 0.099 & 0.231 & 0.139 \\
Grasper & 680 & 4174 & 4663 & 0.140 & 0.127 & 0.133 \\
Scissor & 537 & 5600 & 2306 & 0.087 & 0.189 & 0.120 \\
Surgical Knife & 449 & 5921 & 1540 & 0.070 & 0.226 & 0.107 \\
Rongeur & 219 & 2482 & 3100 & 0.081 & 0.066 & 0.073 \\
Suction Coagulator & 338 & 7800 & 1673 & 0.042 & 0.168 & 0.067 \\
Cotton Patty & 276 & 513 & 7527 & 0.350 & 0.035 & 0.064 \\
Straight Forceps & 226 & 5955 & 1063 & 0.037 & 0.175 & 0.060 \\
Bipolar Forceps & 48 & 1001 & 1529 & 0.046 & 0.030 & 0.037 \\
Aspirating dissector & 61 & 1912 & 1657 & 0.031 & 0.035 & 0.033 \\
Doppler & 39 & 1172 & 1125 & 0.032 & 0.034 & 0.033 \\
Rhoton Dissector & 61 & 544 & 4122 & 0.101 & 0.015 & 0.025 \\
Backbiting rongeur & 21 & 876 & 836 & 0.023 & 0.025 & 0.024 \\
Tissue shaver & 38 & 2435 & 686 & 0.015 & 0.052 & 0.024 \\
Through cutting forceps & 29 & 2265 & 543 & 0.013 & 0.051 & 0.020 \\
Cottle Elevator & 55 & 5065 & 572 & 0.011 & 0.088 & 0.019 \\
Floseal Gel & 10 & 870 & 381 & 0.011 & 0.026 & 0.016 \\
Surgicel & 22 & 290 & 2598 & 0.070 & 0.008 & 0.015 \\
Irrigation & 93 & 12128 & 186 & 0.008 & 0.333 & 0.015 \\
Sonopet pineapple tip & 12 & 288 & 1452 & 0.040 & 0.008 & 0.014 \\
Suction microdebrider & 19 & 3950 & 424 & 0.005 & 0.043 & 0.009 \\
Monopolar Electrocautery & 18 & 4194 & 120 & 0.004 & 0.130 & 0.008 \\
Floseal Applicator & 7 & 1680 & 168 & 0.004 & 0.040 & 0.007 \\
Local Anesthesia Needle & 4 & 1353 & 105 & 0.003 & 0.037 & 0.005 \\
Straight Curette & 6 & 3032 & 83 & 0.002 & 0.067 & 0.004 \\
Collagen Matrix & 0 & 91 & 214 & 0.000 & 0.000 & 0.000 \\
Needle & 0 & 837 & 20 & 0.000 & 0.000 & 0.000 \\
Tisseel Applicator & 0 & 700 & 15 & 0.000 & 0.000 & 0.000 \\
unknown & 0 & 134 & 890 & 0.000 & 0.000 & 0.000 \\
\bottomrule
\end{tabular}
\end{table}

\begin{table}[H]
\centering
\caption{Per-tool metrics for Qwen2-VL-2B-Instruct (2B) zero-shot evaluation. P = precision, R = recall.}
\label{tab:zero_shot_qwen2_vl_2b_instruct}
\scriptsize
\begin{tabular}{lrrrrrr}
\toprule
Tool & TP & FP & FN & P & R & F1 \\
\midrule
Grasper & 748 & 5279 & 3420 & 0.124 & 0.179 & 0.147 \\
Curette & 1110 & 11909 & 2159 & 0.085 & 0.340 & 0.136 \\
Rhoton Dissector & 593 & 6046 & 2669 & 0.089 & 0.182 & 0.120 \\
Drill & 238 & 1733 & 2024 & 0.121 & 0.105 & 0.112 \\
Aspirating dissector & 283 & 5676 & 1093 & 0.048 & 0.206 & 0.077 \\
Surgical Knife & 343 & 7496 & 1282 & 0.044 & 0.211 & 0.072 \\
Surgicel & 139 & 2026 & 1948 & 0.064 & 0.067 & 0.065 \\
Rongeur & 126 & 1728 & 2517 & 0.068 & 0.048 & 0.056 \\
Bipolar Forceps & 114 & 2978 & 1119 & 0.037 & 0.092 & 0.053 \\
Suction & 633 & 414 & 22857 & 0.605 & 0.027 & 0.052 \\
Doppler & 133 & 4515 & 864 & 0.029 & 0.133 & 0.047 \\
Scissor & 49 & 715 & 2126 & 0.064 & 0.022 & 0.033 \\
Backbiting rongeur & 28 & 1005 & 720 & 0.027 & 0.037 & 0.031 \\
Straight Forceps & 21 & 519 & 1030 & 0.039 & 0.020 & 0.026 \\
Suction Coagulator & 27 & 507 & 1490 & 0.051 & 0.018 & 0.026 \\
Sonopet pineapple tip & 22 & 468 & 1174 & 0.045 & 0.018 & 0.026 \\
Tissue shaver & 25 & 1485 & 511 & 0.017 & 0.047 & 0.024 \\
Cottle Elevator & 28 & 1921 & 462 & 0.014 & 0.057 & 0.023 \\
Floseal Gel & 25 & 2934 & 309 & 0.008 & 0.075 & 0.015 \\
Through cutting forceps & 12 & 1274 & 486 & 0.009 & 0.024 & 0.013 \\
Collagen Matrix & 31 & 4494 & 140 & 0.007 & 0.181 & 0.013 \\
Suction microdebrider & 8 & 878 & 351 & 0.009 & 0.022 & 0.013 \\
Irrigation & 14 & 1991 & 223 & 0.007 & 0.059 & 0.013 \\
Floseal Applicator & 20 & 4768 & 129 & 0.004 & 0.134 & 0.008 \\
Straight Curette & 5 & 1170 & 69 & 0.004 & 0.068 & 0.008 \\
Monopolar Electrocautery & 5 & 1324 & 101 & 0.004 & 0.047 & 0.007 \\
Cotton Patty & 16 & 71 & 6175 & 0.184 & 0.003 & 0.005 \\
Tisseel Applicator & 1 & 732 & 14 & 0.001 & 0.067 & 0.003 \\
Local Anesthesia Needle & 1 & 790 & 97 & 0.001 & 0.010 & 0.002 \\
Needle & 0 & 371 & 16 & 0.000 & 0.000 & 0.000 \\
unknown & 0 & 33 & 723 & 0.000 & 0.000 & 0.000 \\
\bottomrule
\end{tabular}
\end{table}

\begin{table}[H]
\centering
\caption{Per-tool metrics for Gemma 3 27B-it (27B) zero-shot evaluation. P = precision, R = recall.}
\label{tab:zero_shot_gemma_3_27b_it}
\scriptsize
\begin{tabular}{lrrrrrr}
\toprule
Tool & TP & FP & FN & P & R & F1 \\
\midrule
Suction & 39555 & 26381 & 826 & 0.600 & 0.980 & 0.744 \\
Rongeur & 1424 & 14204 & 3095 & 0.091 & 0.315 & 0.141 \\
Drill & 154 & 32 & 3600 & 0.828 & 0.041 & 0.078 \\
Bipolar Forceps & 277 & 5299 & 1920 & 0.050 & 0.126 & 0.071 \\
Grasper & 329 & 1596 & 6993 & 0.171 & 0.045 & 0.071 \\
Cotton Patty & 232 & 69 & 10737 & 0.771 & 0.021 & 0.041 \\
Through cutting forceps & 216 & 14551 & 581 & 0.015 & 0.271 & 0.028 \\
Scissor & 42 & 177 & 3882 & 0.192 & 0.011 & 0.020 \\
Straight Forceps & 25 & 666 & 1750 & 0.036 & 0.014 & 0.020 \\
Surgical Knife & 24 & 182 & 2763 & 0.117 & 0.009 & 0.016 \\
Sonopet pineapple tip & 21 & 816 & 1970 & 0.025 & 0.011 & 0.015 \\
Surgicel & 25 & 169 & 3562 & 0.129 & 0.007 & 0.013 \\
Curette & 35 & 129 & 5504 & 0.213 & 0.006 & 0.012 \\
Irrigation & 49 & 8101 & 337 & 0.006 & 0.127 & 0.011 \\
Monopolar Electrocautery & 23 & 5683 & 167 & 0.004 & 0.121 & 0.008 \\
Suction microdebrider & 3 & 461 & 588 & 0.006 & 0.005 & 0.006 \\
Cottle Elevator & 1 & 60 & 854 & 0.016 & 0.001 & 0.002 \\
Straight Curette & 1 & 962 & 117 & 0.001 & 0.009 & 0.002 \\
Rhoton Dissector & 5 & 27 & 5708 & 0.156 & 0.001 & 0.002 \\
Aspirating dissector & 0 & 10 & 2407 & 0.000 & 0.000 & 0.000 \\
Backbiting rongeur & 0 & 0 & 1185 & 0.000 & 0.000 & 0.000 \\
Collagen Matrix & 0 & 538 & 280 & 0.000 & 0.000 & 0.000 \\
Doppler & 0 & 0 & 1592 & 0.000 & 0.000 & 0.000 \\
Floseal Applicator & 0 & 4 & 239 & 0.000 & 0.000 & 0.000 \\
Floseal Gel & 0 & 12 & 542 & 0.000 & 0.000 & 0.000 \\
Local Anesthesia Needle & 0 & 5 & 139 & 0.000 & 0.000 & 0.000 \\
Needle & 0 & 257 & 31 & 0.000 & 0.000 & 0.000 \\
Suction Coagulator & 0 & 54 & 2588 & 0.000 & 0.000 & 0.000 \\
Tisseel Applicator & 0 & 0 & 23 & 0.000 & 0.000 & 0.000 \\
Tissue shaver & 0 & 47 & 991 & 0.000 & 0.000 & 0.000 \\
unknown & 0 & 2 & 1195 & 0.000 & 0.000 & 0.000 \\
\bottomrule
\end{tabular}
\end{table}

\begin{table}[H]
\centering
\caption{Per-tool metrics for Gemma 3 12B-it (12B) zero-shot evaluation. P = precision, R = recall.}
\label{tab:zero_shot_gemma_3_12b_it}
\scriptsize
\begin{tabular}{lrrrrrr}
\toprule
Tool & TP & FP & FN & P & R & F1 \\
\midrule
Suction & 26679 & 15734 & 13698 & 0.629 & 0.661 & 0.644 \\
Grasper & 4431 & 23958 & 2889 & 0.156 & 0.605 & 0.248 \\
Suction Coagulator & 1180 & 15044 & 1407 & 0.073 & 0.456 & 0.126 \\
Rhoton Dissector & 825 & 6977 & 4887 & 0.106 & 0.144 & 0.122 \\
Scissor & 327 & 2297 & 3598 & 0.125 & 0.083 & 0.100 \\
Sonopet pineapple tip & 101 & 611 & 1889 & 0.142 & 0.051 & 0.075 \\
Local Anesthesia Needle & 9 & 236 & 130 & 0.037 & 0.065 & 0.047 \\
Drill & 85 & 27 & 3670 & 0.759 & 0.023 & 0.044 \\
Rongeur & 158 & 2530 & 4361 & 0.059 & 0.035 & 0.044 \\
Bipolar Forceps & 65 & 1562 & 2132 & 0.040 & 0.030 & 0.034 \\
Cotton Patty & 181 & 250 & 10786 & 0.420 & 0.017 & 0.032 \\
Surgical Knife & 71 & 2207 & 2716 & 0.031 & 0.025 & 0.028 \\
Surgicel & 38 & 119 & 3548 & 0.242 & 0.011 & 0.020 \\
Monopolar Electrocautery & 5 & 311 & 185 & 0.016 & 0.026 & 0.020 \\
Irrigation & 149 & 24718 & 237 & 0.006 & 0.386 & 0.012 \\
Straight Forceps & 10 & 552 & 1765 & 0.018 & 0.006 & 0.009 \\
Suction microdebrider & 3 & 366 & 588 & 0.008 & 0.005 & 0.006 \\
Through cutting forceps & 2 & 47 & 794 & 0.041 & 0.003 & 0.005 \\
Tissue shaver & 3 & 282 & 988 & 0.011 & 0.003 & 0.005 \\
Aspirating dissector & 7 & 667 & 2400 & 0.010 & 0.003 & 0.004 \\
Curette & 9 & 32 & 5530 & 0.220 & 0.002 & 0.003 \\
Backbiting rongeur & 0 & 0 & 1185 & 0.000 & 0.000 & 0.000 \\
Collagen Matrix & 0 & 954 & 280 & 0.000 & 0.000 & 0.000 \\
Cottle Elevator & 0 & 76 & 855 & 0.000 & 0.000 & 0.000 \\
Doppler & 0 & 26 & 1592 & 0.000 & 0.000 & 0.000 \\
Floseal Applicator & 0 & 133 & 238 & 0.000 & 0.000 & 0.000 \\
Floseal Gel & 0 & 43 & 541 & 0.000 & 0.000 & 0.000 \\
Needle & 0 & 582 & 31 & 0.000 & 0.000 & 0.000 \\
Straight Curette & 0 & 1 & 118 & 0.000 & 0.000 & 0.000 \\
Tisseel Applicator & 0 & 0 & 23 & 0.000 & 0.000 & 0.000 \\
unknown & 0 & 3 & 1195 & 0.000 & 0.000 & 0.000 \\
\bottomrule
\end{tabular}
\end{table}

\begin{table}[H]
\centering
\caption{Per-tool metrics for Gemma 3 4B-it (4B) zero-shot evaluation. P = precision, R = recall.}
\label{tab:zero_shot_gemma_3_4b_it}
\scriptsize
\begin{tabular}{lrrrrrr}
\toprule
Tool & TP & FP & FN & P & R & F1 \\
\midrule
Suction & 39017 & 26280 & 1298 & 0.598 & 0.968 & 0.739 \\
Grasper & 3699 & 24871 & 3598 & 0.130 & 0.507 & 0.206 \\
Rongeur & 1714 & 23095 & 2802 & 0.069 & 0.380 & 0.117 \\
Surgical Knife & 2314 & 43203 & 469 & 0.051 & 0.832 & 0.096 \\
Bipolar Forceps & 597 & 13902 & 1596 & 0.041 & 0.272 & 0.071 \\
Suction Coagulator & 136 & 1328 & 2449 & 0.093 & 0.053 & 0.067 \\
Straight Forceps & 1299 & 46446 & 468 & 0.027 & 0.735 & 0.052 \\
Surgicel & 96 & 489 & 3485 & 0.164 & 0.027 & 0.046 \\
Monopolar Electrocautery & 5 & 131 & 185 & 0.037 & 0.026 & 0.031 \\
Curette & 102 & 1512 & 5430 & 0.063 & 0.018 & 0.029 \\
Cottle Elevator & 200 & 15114 & 655 & 0.013 & 0.234 & 0.025 \\
Aspirating dissector & 90 & 5718 & 2314 & 0.015 & 0.037 & 0.022 \\
Local Anesthesia Needle & 1 & 5 & 138 & 0.167 & 0.007 & 0.014 \\
Irrigation & 100 & 15849 & 286 & 0.006 & 0.259 & 0.012 \\
Drill & 21 & 7 & 3731 & 0.750 & 0.006 & 0.011 \\
Rhoton Dissector & 32 & 415 & 5669 & 0.072 & 0.006 & 0.010 \\
Scissor & 14 & 123 & 3906 & 0.102 & 0.004 & 0.007 \\
Doppler & 6 & 448 & 1582 & 0.013 & 0.004 & 0.006 \\
Cotton Patty & 27 & 8 & 10920 & 0.771 & 0.003 & 0.005 \\
Straight Curette & 4 & 2463 & 114 & 0.002 & 0.034 & 0.003 \\
Backbiting rongeur & 1 & 28 & 1184 & 0.035 & 0.001 & 0.002 \\
Needle & 7 & 9710 & 24 & 0.001 & 0.226 & 0.001 \\
Collagen Matrix & 0 & 3 & 280 & 0.000 & 0.000 & 0.000 \\
Floseal Applicator & 0 & 0 & 239 & 0.000 & 0.000 & 0.000 \\
Floseal Gel & 0 & 0 & 541 & 0.000 & 0.000 & 0.000 \\
Sonopet pineapple tip & 0 & 0 & 1983 & 0.000 & 0.000 & 0.000 \\
Suction microdebrider & 0 & 38 & 590 & 0.000 & 0.000 & 0.000 \\
Through cutting forceps & 0 & 32 & 795 & 0.000 & 0.000 & 0.000 \\
Tisseel Applicator & 0 & 0 & 23 & 0.000 & 0.000 & 0.000 \\
Tissue shaver & 0 & 108 & 985 & 0.000 & 0.000 & 0.000 \\
unknown & 0 & 0 & 1192 & 0.000 & 0.000 & 0.000 \\
\bottomrule
\end{tabular}
\end{table}

\begin{table}[H]
\centering
\caption{Per-tool metrics for MedGemma 3 27B-it (27B) zero-shot evaluation. P = precision, R = recall.}
\label{tab:zero_shot_medgemma_3_27b_it}
\scriptsize
\begin{tabular}{lrrrrrr}
\toprule
Tool & TP & FP & FN & P & R & F1 \\
\midrule
Suction & 38413 & 24419 & 1970 & 0.611 & 0.951 & 0.744 \\
Rongeur & 2042 & 13802 & 2478 & 0.129 & 0.452 & 0.201 \\
Surgical Knife & 851 & 9387 & 1936 & 0.083 & 0.305 & 0.131 \\
Curette & 286 & 2206 & 5253 & 0.115 & 0.052 & 0.071 \\
Sonopet pineapple tip & 72 & 110 & 1919 & 0.396 & 0.036 & 0.066 \\
Cotton Patty & 344 & 29 & 10625 & 0.922 & 0.031 & 0.061 \\
Grasper & 190 & 295 & 7132 & 0.392 & 0.026 & 0.049 \\
Suction Coagulator & 218 & 12464 & 2370 & 0.017 & 0.084 & 0.029 \\
Straight Forceps & 36 & 870 & 1739 & 0.040 & 0.020 & 0.027 \\
unknown & 147 & 10757 & 1048 & 0.013 & 0.123 & 0.024 \\
Monopolar Electrocautery & 13 & 1375 & 177 & 0.009 & 0.068 & 0.017 \\
Irrigation & 52 & 9955 & 334 & 0.005 & 0.135 & 0.010 \\
Suction microdebrider & 33 & 6177 & 558 & 0.005 & 0.056 & 0.010 \\
Scissor & 15 & 44 & 3910 & 0.254 & 0.004 & 0.007 \\
Straight Curette & 4 & 3246 & 114 & 0.001 & 0.034 & 0.002 \\
Tissue shaver & 1 & 30 & 990 & 0.032 & 0.001 & 0.002 \\
Doppler & 1 & 14 & 1591 & 0.067 & 0.001 & 0.001 \\
Bipolar Forceps & 1 & 23 & 2196 & 0.042 & 0.001 & 0.001 \\
Aspirating dissector & 1 & 26 & 2406 & 0.037 & 0.000 & 0.001 \\
Rhoton Dissector & 2 & 19 & 5711 & 0.095 & 0.000 & 0.001 \\
Surgicel & 1 & 28 & 3586 & 0.035 & 0.000 & 0.001 \\
Backbiting rongeur & 0 & 10 & 1185 & 0.000 & 0.000 & 0.000 \\
Collagen Matrix & 0 & 636 & 280 & 0.000 & 0.000 & 0.000 \\
Cottle Elevator & 0 & 12 & 855 & 0.000 & 0.000 & 0.000 \\
Drill & 0 & 16 & 3755 & 0.000 & 0.000 & 0.000 \\
Floseal Applicator & 0 & 15 & 239 & 0.000 & 0.000 & 0.000 \\
Floseal Gel & 0 & 15 & 542 & 0.000 & 0.000 & 0.000 \\
Local Anesthesia Needle & 0 & 16 & 139 & 0.000 & 0.000 & 0.000 \\
Needle & 0 & 17 & 31 & 0.000 & 0.000 & 0.000 \\
Through cutting forceps & 0 & 44 & 797 & 0.000 & 0.000 & 0.000 \\
Tisseel Applicator & 0 & 15 & 23 & 0.000 & 0.000 & 0.000 \\
\bottomrule
\end{tabular}
\end{table}

\begin{table}[H]
\centering
\caption{Per-tool metrics for Llama-3.2-90B-Vision (90B) zero-shot evaluation. P = precision, R = recall.}
\label{tab:zero_shot_llama_3_2_90b_vision}
\scriptsize
\begin{tabular}{lrrrrrr}
\toprule
Tool & TP & FP & FN & P & R & F1 \\
\midrule
Suction & 24755 & 6689 & 15626 & 0.787 & 0.613 & 0.689 \\
Drill & 1264 & 1005 & 2489 & 0.557 & 0.337 & 0.420 \\
Cotton Patty & 2569 & 2671 & 8400 & 0.490 & 0.234 & 0.317 \\
Rongeur & 1062 & 5120 & 3458 & 0.172 & 0.235 & 0.199 \\
Curette & 657 & 2324 & 4882 & 0.220 & 0.119 & 0.154 \\
Grasper & 669 & 1046 & 6653 & 0.390 & 0.091 & 0.148 \\
Rhoton Dissector & 705 & 7066 & 5008 & 0.091 & 0.123 & 0.105 \\
Surgical Knife & 167 & 1094 & 2620 & 0.132 & 0.060 & 0.083 \\
Bipolar Forceps & 153 & 2900 & 2044 & 0.050 & 0.070 & 0.058 \\
Surgicel & 97 & 1064 & 3490 & 0.084 & 0.027 & 0.041 \\
Backbiting rongeur & 53 & 1699 & 1132 & 0.030 & 0.045 & 0.036 \\
Floseal Gel & 24 & 927 & 517 & 0.025 & 0.044 & 0.032 \\
Doppler & 27 & 225 & 1565 & 0.107 & 0.017 & 0.029 \\
Scissor & 64 & 396 & 3861 & 0.139 & 0.016 & 0.029 \\
Irrigation & 12 & 588 & 374 & 0.020 & 0.031 & 0.024 \\
Monopolar Electrocautery & 67 & 5254 & 123 & 0.013 & 0.353 & 0.024 \\
Through cutting forceps & 18 & 846 & 779 & 0.021 & 0.023 & 0.022 \\
Straight Forceps & 27 & 1143 & 1748 & 0.023 & 0.015 & 0.018 \\
Floseal Applicator & 6 & 932 & 233 & 0.006 & 0.025 & 0.010 \\
Sonopet pineapple tip & 10 & 207 & 1981 & 0.046 & 0.005 & 0.009 \\
Suction Coagulator & 39 & 6827 & 2550 & 0.006 & 0.015 & 0.008 \\
Collagen Matrix & 3 & 553 & 277 & 0.005 & 0.011 & 0.007 \\
unknown & 45 & 11803 & 1150 & 0.004 & 0.038 & 0.007 \\
Cottle Elevator & 2 & 225 & 853 & 0.009 & 0.002 & 0.004 \\
Suction microdebrider & 7 & 5387 & 584 & 0.001 & 0.012 & 0.002 \\
Tissue shaver & 1 & 594 & 990 & 0.002 & 0.001 & 0.001 \\
Aspirating dissector & 0 & 478 & 2407 & 0.000 & 0.000 & 0.000 \\
Local Anesthesia Needle & 0 & 295 & 138 & 0.000 & 0.000 & 0.000 \\
Needle & 0 & 291 & 31 & 0.000 & 0.000 & 0.000 \\
Straight Curette & 0 & 467 & 118 & 0.000 & 0.000 & 0.000 \\
Tisseel Applicator & 0 & 789 & 23 & 0.000 & 0.000 & 0.000 \\
\bottomrule
\end{tabular}
\end{table}

\begin{table}[H]
\centering
\caption{Per-tool metrics for Llama-3.2-11B-Vision (11B) zero-shot evaluation. P = precision, R = recall.}
\label{tab:zero_shot_llama_3_2_11b_vision}
\scriptsize
\begin{tabular}{lrrrrrr}
\toprule
Tool & TP & FP & FN & P & R & F1 \\
\midrule
Suction & 505 & 392 & 1058 & 0.563 & 0.323 & 0.411 \\
Curette & 23 & 190 & 174 & 0.108 & 0.117 & 0.112 \\
Rongeur & 15 & 163 & 162 & 0.084 & 0.085 & 0.085 \\
Scissor & 8 & 87 & 125 & 0.084 & 0.060 & 0.070 \\
Bipolar Forceps & 8 & 132 & 87 & 0.057 & 0.084 & 0.068 \\
Grasper & 13 & 115 & 253 & 0.102 & 0.049 & 0.066 \\
Drill & 5 & 9 & 165 & 0.357 & 0.029 & 0.054 \\
Aspirating dissector & 13 & 364 & 104 & 0.035 & 0.111 & 0.053 \\
Surgical Knife & 8 & 309 & 89 & 0.025 & 0.083 & 0.039 \\
Rhoton Dissector & 5 & 19 & 238 & 0.208 & 0.021 & 0.037 \\
Tissue shaver & 10 & 467 & 61 & 0.021 & 0.141 & 0.036 \\
Suction microdebrider & 2 & 168 & 40 & 0.012 & 0.048 & 0.019 \\
Suction Coagulator & 1 & 53 & 110 & 0.018 & 0.009 & 0.012 \\
Cotton Patty & 1 & 4 & 414 & 0.200 & 0.002 & 0.005 \\
Backbiting rongeur & 0 & 20 & 57 & 0.000 & 0.000 & 0.000 \\
Collagen Matrix & 0 & 3 & 17 & 0.000 & 0.000 & 0.000 \\
Cottle Elevator & 0 & 16 & 26 & 0.000 & 0.000 & 0.000 \\
Doppler & 0 & 19 & 64 & 0.000 & 0.000 & 0.000 \\
Floseal Applicator & 0 & 2 & 6 & 0.000 & 0.000 & 0.000 \\
Floseal Gel & 0 & 0 & 16 & 0.000 & 0.000 & 0.000 \\
Irrigation & 0 & 99 & 17 & 0.000 & 0.000 & 0.000 \\
Local Anesthesia Needle & 0 & 26 & 11 & 0.000 & 0.000 & 0.000 \\
Monopolar Electrocautery & 0 & 38 & 11 & 0.000 & 0.000 & 0.000 \\
Needle & 0 & 11 & 5 & 0.000 & 0.000 & 0.000 \\
Sonopet pineapple tip & 0 & 0 & 78 & 0.000 & 0.000 & 0.000 \\
Straight Curette & 0 & 13 & 6 & 0.000 & 0.000 & 0.000 \\
Straight Forceps & 0 & 17 & 78 & 0.000 & 0.000 & 0.000 \\
Surgicel & 0 & 85 & 116 & 0.000 & 0.000 & 0.000 \\
Through cutting forceps & 0 & 24 & 48 & 0.000 & 0.000 & 0.000 \\
Tisseel Applicator & 0 & 78 & 2 & 0.000 & 0.000 & 0.000 \\
unknown & 0 & 2 & 50 & 0.000 & 0.000 & 0.000 \\
\bottomrule
\end{tabular}
\end{table}

\begin{table}[H]
\centering
\caption{Per-tool metrics for LLaVA-1.5-13B (13B) zero-shot evaluation. P = precision, R = recall.}
\label{tab:zero_shot_llava_1_5_13b}
\scriptsize
\begin{tabular}{lrrrrrr}
\toprule
Tool & TP & FP & FN & P & R & F1 \\
\midrule
Suction & 22423 & 14880 & 17960 & 0.601 & 0.555 & 0.577 \\
Cotton Patty & 6339 & 27699 & 4630 & 0.186 & 0.578 & 0.282 \\
Grasper & 6850 & 54394 & 472 & 0.112 & 0.935 & 0.200 \\
Rhoton Dissector & 5405 & 58418 & 308 & 0.085 & 0.946 & 0.155 \\
Curette & 5429 & 59769 & 110 & 0.083 & 0.980 & 0.153 \\
Rongeur & 3798 & 53913 & 722 & 0.066 & 0.840 & 0.122 \\
Scissor & 3876 & 61604 & 49 & 0.059 & 0.988 & 0.112 \\
Surgicel & 2338 & 39493 & 1249 & 0.056 & 0.652 & 0.103 \\
Drill & 1661 & 27772 & 2094 & 0.056 & 0.442 & 0.100 \\
Suction Coagulator & 1309 & 29208 & 1279 & 0.043 & 0.506 & 0.079 \\
Surgical Knife & 2406 & 56776 & 381 & 0.041 & 0.863 & 0.078 \\
Aspirating dissector & 2353 & 62594 & 54 & 0.036 & 0.978 & 0.070 \\
Bipolar Forceps & 2150 & 62184 & 47 & 0.033 & 0.979 & 0.065 \\
Sonopet pineapple tip & 905 & 28587 & 1086 & 0.031 & 0.455 & 0.058 \\
Backbiting rongeur & 110 & 2896 & 1075 & 0.037 & 0.093 & 0.052 \\
Straight Forceps & 821 & 28844 & 954 & 0.028 & 0.463 & 0.052 \\
Doppler & 1010 & 49108 & 582 & 0.020 & 0.634 & 0.039 \\
unknown & 163 & 7614 & 1032 & 0.021 & 0.136 & 0.036 \\
Tissue shaver & 924 & 61184 & 67 & 0.015 & 0.932 & 0.029 \\
Cottle Elevator & 355 & 29201 & 500 & 0.012 & 0.415 & 0.023 \\
Floseal Gel & 261 & 28761 & 281 & 0.009 & 0.481 & 0.018 \\
Suction microdebrider & 252 & 29226 & 339 & 0.009 & 0.426 & 0.017 \\
Irrigation & 374 & 63968 & 12 & 0.006 & 0.969 & 0.012 \\
Floseal Applicator & 91 & 28909 & 148 & 0.003 & 0.381 & 0.006 \\
Collagen Matrix & 73 & 25090 & 207 & 0.003 & 0.261 & 0.006 \\
Monopolar Electrocautery & 185 & 64062 & 5 & 0.003 & 0.974 & 0.006 \\
Local Anesthesia Needle & 123 & 49177 & 16 & 0.003 & 0.885 & 0.005 \\
Straight Curette & 53 & 29563 & 65 & 0.002 & 0.449 & 0.004 \\
Needle & 18 & 32276 & 13 & 0.001 & 0.581 & 0.001 \\
Tisseel Applicator & 17 & 46181 & 6 & 0.000 & 0.739 & 0.001 \\
Through cutting forceps & 0 & 137 & 797 & 0.000 & 0.000 & 0.000 \\
\bottomrule
\end{tabular}
\end{table}

\begin{table}[H]
\centering
\caption{Per-tool metrics for Gemma 4 31B-it (31B) zero-shot evaluation. P = precision, R = recall.}
\label{tab:zero_shot_gemma_4_31b_it}
\scriptsize
\begin{tabular}{lrrrrrr}
\toprule
Tool & TP & FP & FN & P & R & F1 \\
\midrule
Suction & 30023 & 14133 & 10360 & 0.680 & 0.744 & 0.710 \\
Cotton Patty & 1504 & 211 & 9465 & 0.877 & 0.137 & 0.237 \\
Drill & 408 & 421 & 3347 & 0.492 & 0.109 & 0.178 \\
Bipolar Forceps & 691 & 5315 & 1506 & 0.115 & 0.315 & 0.169 \\
Straight Forceps & 339 & 2565 & 1436 & 0.117 & 0.191 & 0.145 \\
Grasper & 500 & 327 & 6822 & 0.605 & 0.068 & 0.123 \\
Curette & 341 & 103 & 5198 & 0.768 & 0.062 & 0.114 \\
Surgicel & 146 & 180 & 3441 & 0.448 & 0.041 & 0.075 \\
Sonopet pineapple tip & 74 & 22 & 1917 & 0.771 & 0.037 & 0.071 \\
Rhoton Dissector & 80 & 260 & 5633 & 0.235 & 0.014 & 0.026 \\
Suction Coagulator & 19 & 280 & 2569 & 0.064 & 0.007 & 0.013 \\
Irrigation & 2 & 34 & 384 & 0.056 & 0.005 & 0.009 \\
Rongeur & 13 & 302 & 4507 & 0.041 & 0.003 & 0.005 \\
Through cutting forceps & 2 & 14 & 795 & 0.125 & 0.003 & 0.005 \\
unknown & 2 & 51 & 1193 & 0.038 & 0.002 & 0.003 \\
Scissor & 2 & 1 & 3923 & 0.667 & 0.001 & 0.001 \\
Aspirating dissector & 0 & 0 & 2407 & 0.000 & 0.000 & 0.000 \\
Backbiting rongeur & 0 & 2 & 1185 & 0.000 & 0.000 & 0.000 \\
Collagen Matrix & 0 & 476 & 280 & 0.000 & 0.000 & 0.000 \\
Cottle Elevator & 0 & 12 & 855 & 0.000 & 0.000 & 0.000 \\
Doppler & 0 & 0 & 1592 & 0.000 & 0.000 & 0.000 \\
Floseal Applicator & 0 & 3 & 239 & 0.000 & 0.000 & 0.000 \\
Floseal Gel & 0 & 4 & 542 & 0.000 & 0.000 & 0.000 \\
Local Anesthesia Needle & 0 & 0 & 139 & 0.000 & 0.000 & 0.000 \\
Monopolar Electrocautery & 0 & 5 & 190 & 0.000 & 0.000 & 0.000 \\
Needle & 0 & 11 & 31 & 0.000 & 0.000 & 0.000 \\
Straight Curette & 0 & 5 & 118 & 0.000 & 0.000 & 0.000 \\
Suction microdebrider & 0 & 186 & 591 & 0.000 & 0.000 & 0.000 \\
Surgical Knife & 0 & 1 & 2787 & 0.000 & 0.000 & 0.000 \\
Tisseel Applicator & 0 & 0 & 23 & 0.000 & 0.000 & 0.000 \\
Tissue shaver & 0 & 79 & 991 & 0.000 & 0.000 & 0.000 \\
\bottomrule
\end{tabular}
\end{table}